# HIERARCHICAL MODELING

# OF MULTIDIMENSIONAL DATA

# IN REGULARLY DECOMPOSED SPACES

## TOME 3 : APPLICATIONS IN IMAGE ANALYSIS

## (1982 – 2006)

**- 2016 -**

**Olivier Guye**

# Table of Contents

# Table of Figures

# Introduction

This third tome presents the developments realized by a work group interested in the applications of image analysis and in the progressive introduction of multidimensional hierarchical modeling techniques. These activities have started in the field of industrial imaging, then have been focused on satellite imaging and cartography, and after by carrying on face recognition and finally ending by settling the principles of self-descriptive video coding.

This activity group was implied during a bit more than twenty years in the following fields inside ADERSA:

- real-time and parallel computing architectures,
- data capture and registration,
- image analysis,
- statistical data analysis,
- decision making,
- visual rendering.

The main part of these activities have been focused on the following areas :

- industrial applications of real-time image analysis,
- advanced techniques in image analysis and problem solving.

The application domains of artificial vision are metrology, localization in a work space, sorting and object identification, inspection and quality check, surveillance and environment control.

The applications of data analysis and problem solving are located in an area which is at the conjunction of image analysis, statistical data analysis and operations research. The problems to be solved are represented in a geometrical manner in a space with as many dimensions as there are variables in the problem. After problem modeling, solving algorithms are used in order to find some compliant solutions.

 

# 1. Overall Presentation

This last document presents the progressive introduction of hierarchical modeling techniques in image analysis

The first chapter deals with the early works made in the field of the industrial applications of pattern recognition. It is a project done before the beginning of the works realized in the field of multidimensional hierarchical modeling but which is already included the main steps of a system for visual pattern recognition. It has been developed at the early beginning of the development of industrial applications in robotics and when it was looking for extending the perception capabilities of robots and industrial automatisms. This project has got the support of Pr Claude Laurgeau, at this moment responsible of department « Research in Robotics and Computer-Integrated Manufacturing» at the French Computing Science Agency. The results of these works and their industrial impacts have been presented during the MODULAD days organized by the CNET and the INRIA in Brittany.

The second chapter deals with the application of these techniques to satellite imagery. Nearly at the moment when our attention was focused on multidimensional hierarchical modeling for designing the next systems in modeling, simulating and decision making, it has appeared necessary to develop tools enabling to exploit the new images captured by higher resolution earth resource observation satellites in the aim to build and to update cartographic data bases in a more effective manner than at this while. It is why we have been asked in a first time for developing image analysis functions enabling to automatically extract digital information from views captured by these satellites, then for designing the means enabling to manage such information in a huge manner. We have been led to develop a cataloging system of satellite data indexed by its response (radiometric), its shape, its localization and its orientation and to experiment it on parallel architecture processing and storage systems.

Then we have been interested in coding and describing data in still and moving pictures by focusing our attention on the applications of person identification by face recognition. The followed path led us to implement a new technique for pattern recognition at half-way between the statistical and structural approaches enabling to take in account partially hidden shapes.

At last by relying on the set of all these advances, the document end takes back the main part of a research proposal describing a new approach for coding still and moving pictures, setting up by this way a bridge between the bodies of MPEG-4 and MPEG-7 standards. It concerns self-descriptive coding and the project has been selected in a first step during the European calls for research projects OPEN-FET (proposal FP6-2002-IST-C named GROOVIES for « Geometrical RepresentatiOns fOr Video and Image indexing, Editing and CompresSion»).

[1] O.Guye. Applications Industrielles de la Reconnaissance des Formes en Vision Artificielle. Journées MODULAD : Applications Industrielles de l'Analyse de Données. Lannion-Trégastel, 14-15 juin 1990.

# 2. Industrial Applications of Pattern Recognition in Artificial Vision

## 2.1. Presentation

The artificial vision systems enable automated systems to perceive the universe in which they are evolving: they are perception systems without any touch.

Five large artificial vision application fields can be distinguished in manufacturing:

- ― metrology with the help of capture devices ;
- ― localization enabling for instance a manipulator arm to catch an object ;
- ― object sorting and identification ;
- ― inspection whose aim is to check the quality of an object being currently manufactured or ending its manufacturing ;
- ― environment watching and control in order to prevent accidents and to protect manufacturing means.

We will show how statistical data analysis can be applied on metrology, inspection and object identification, the localization will be one of its fallouts. It will be illustrated with a concrete and real example.

## 2.2. Perception procedures in pattern recognition

### 2.2.1. Common perception models in artificial vision

Research works in the domain of artificial vision have successively been carried on the three following methodologies:

- ― pattern correlation ;
- ― pattern recognition based on decision theory ;
- ― structural shape recognition.

The topic of this presentation relies on this second methodology. This one is interested in the situations where several objects may appear in the viewing space, but without any contact with each other, while structural recognition is dealing with taking in account unsorted planar objects (partial covering).

### 2.2.2. Methodology followed in statistical pattern recognition

Two successive steps can be found with a pattern recognition system based decision theory :

- ― the learning phase, mode in which the system collects, under the supervision of an operator, the necessary experiences for building a classification model ;
- ― the recognition phase, operating mode where the system classifies by itself every new experience that is submitted to him.

The perception procedures in computer vision work as a « funnel » : the sensor provides after digitalization a huge information, several processing tasks are sequentially performed for reducing this volume without losing the pertinent information that should enable it to give the awaited interpretation about the captured scene.

The processing tasks are the following ones:

- preprocessing that enables the system to take in account some evolutions about the external environment and the working hazards of the capture stage ;
- segmentation whose aim is to identify the homogeneous data sets belonging to the image ;
- attribute computation that enables to assess various measurements on these sets ;
- classification that applies the recognition model built during the learning phase on these measurements in order to deliver an interpretation.

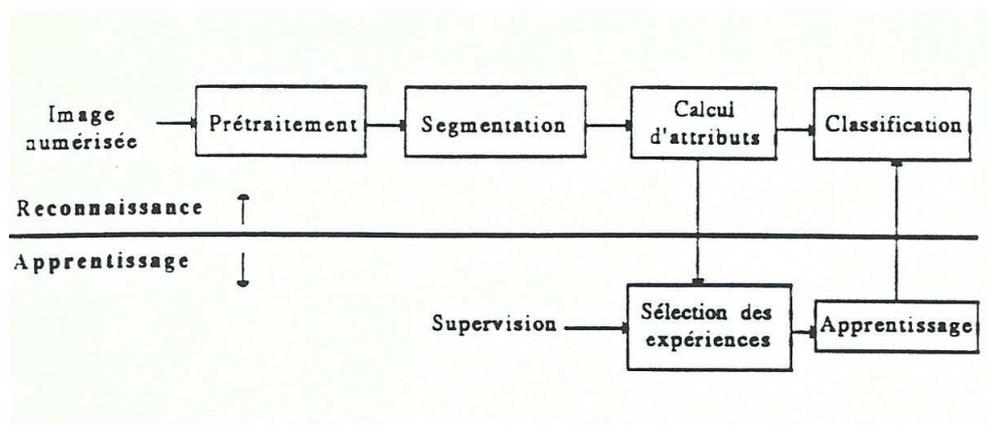

**Figure 1 : Perception procedure in statistical pattern recognition**

Before describing the different steps of a perception procedure in statistical pattern recognition, let us define the common environment conditions in which works an artificial vision system in an industrial application.

## *2.3. Lighting and shooting geometry*

### 2.3.1. Shooting geometry

According to the kind of application, the number of dimensions of the space to handle may vary from one to four (by including the temporal dimension):

- in metrology, mono-dimensional sensors may be enough ;
- in bi-dimensional spaces, it is used vidicon or solid state matrix cameras ;
- tridimensional spaces are hardly manageable without any artifice and do not accept for the moment satisfactory response times in industrial applications.

It is why, we will only focus on planar sensors. These ones are used in a tridimensional environment and need some usage cares: they are usually placed orthogonally to the scene work space and depth of field is tuned as they would be at an infinite distance in order to reduce geometric distortions due to the shooting perspective.

### 2.3.2. Lighting control

The industrial applications of artificial vision can be distinguished from natural scene analysis by the fact that the lighting is partially or fully controlled.

The use of lighting sources complementary to the view capture system may be at the core of some troubles like:

    — the presence of large shadows ;
    — the appearance of parasitical reflections,

These last ones are generated by a direct lighting of the scene. It can be solved by using a ring-shaped or diffusing lighting.

Usually, the implemented lighting is studied in such a manner that the object of interest can be clearly distinguished from the scene background according to its luminous contrast.

## 2.4. Image preprocessing

The artificial vision systems enable the automated systems to perceive the universe in which they are evolving.

The preprocessing represents a preparation phase to segmentation:

    — on one hand, it must enable to remove les degradations that an image suffers (image restoration, improvement) ;
    — on the other hand, it serves to prepare data before segmentation in such a manner that the homogenous image areas can be easily identified.

In practice, the information volume being huge, only the second point is taken in account so as to reduce computation time: the degradations altering an image will be managed by the decision stage.

Segmentation preparation is performed in converting the multi-level image into a binary image where the sets of interest are clearly separated from the background. The binarisation of an image applies by computing a binarisation threshold, then by comparing each image element with this threshold in order to class it either under, or above it. This operation is only valid when the object or the objects of interest got a luminous response opposed to the one of the scene background. As it has already be seen, the lighting control enables to insure this condition.

The computation of the binarisation threshold is performed by analyzing the luminescence histogram of the image. Theoretically the image histogram shows two modes, a local maximum linked to the object and the other one to the background, and a valley (minimum) separating the two modes. This valley is chosen as a threshold for separating the two luminescence distributions. When the lighting is generating shades on the scene background, a third mode is appearing: the valley the closest to the object is then chosen.

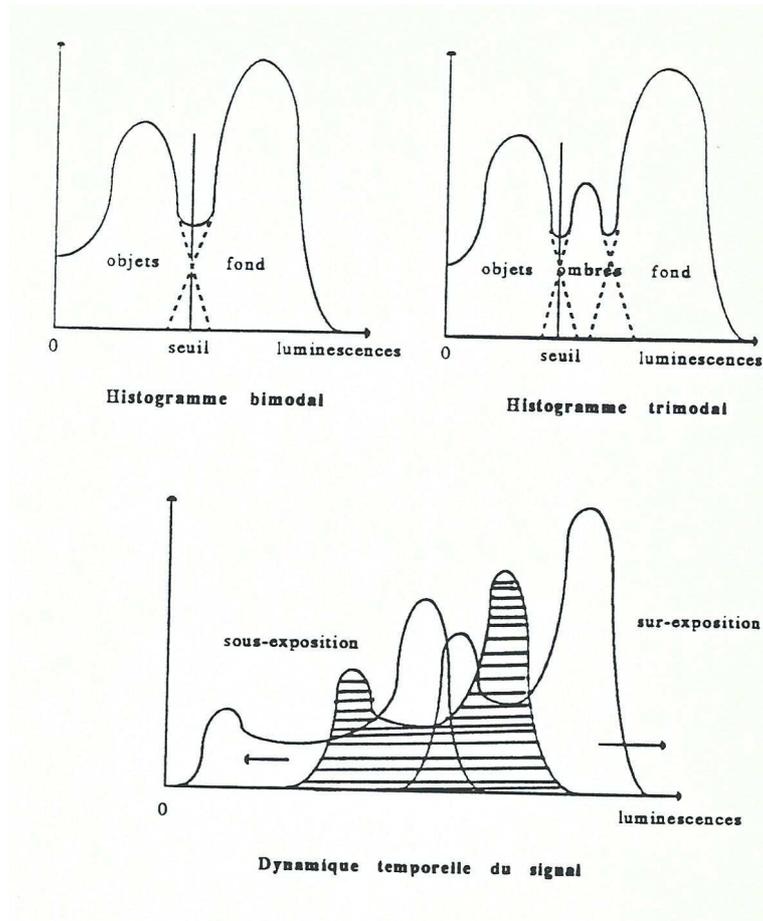

**Figure 2 : Luminescence distributions in an image**

The threshold computation is performed on the smoothed histogram version so as to remove the parasitical modes due to the shooting conditions. This one is updated at each image capture so as to be insensible to the temporal modifications of the luminous ambience and in order to be adapted to the thermal shift of the analogical stages of the capture system.

## 2.5. Image segmentation

Having at one's disposal a binary image, the search for homogenous components can be summarized to the one of the iso-colored connected components. A set $V$ of image points will be connected if $\forall (A_1, A_2) \in V \times V$, $\exists$ a string of adjacent points linking $A_1$ and $A_2$ verifying the iso-coloring predicate. On a square mesh, two distances enable to define the adjacency of two points:

- $d_1(A_1, A_2)$, sum of the absolute values of the coordinate differences of the two points ;
- $d_\infty(A_1, A_2)$, maximum of the same values.

The points $A_1$ et $A_2$ will be told:

- 4 - connected if $d_1(A_1, A_2) \leq 1$ ;
- 8 - connected if $d_\infty(A_1, A_2) \leq 1$ ;

The search for 8-connected components is used for identify objects of unitary thickness.

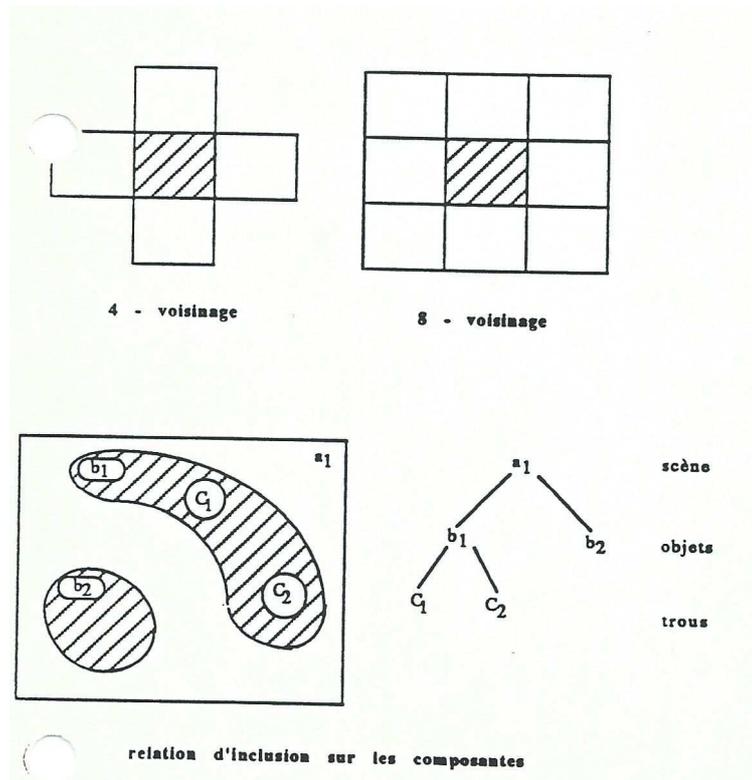

**Figure 3 : Segmentation of a binary image**

In addition, the relation of inclusion between sets is computed on the segmentation result. This one enables to provide a hierarchical representation of the scene:

- − where this last one corresponds to the root of this hierarchy ;
- − the root sons, the objects belonging to the scene ;
- − the grand-sons, the holes in the objects.

At the opposite of the partitioning methods used in multidimensional data analysis, the segmentation of an image is easier to lead, knowing that:

- − the space is planar ;
- − data is regularly sampled in this space.

## *2.6. Attribute calculus*

### 2.6.1. Generalized moments

After having isolated the objects existing in the image, it is remaining one following step to perform before applying a decision procedure: it is to measure the objects stemming from segmentation. It is the aim of attribute calculus whose result will be to associate each object with a measure vector.

In this aim, we will rely on generalized moments :

$$M_{r_1 \cdots r_n}(X) = E(X_1^{r_1} \cdots X_n^{r_n}) = \int_{R^n} X_1^{r_1} \cdots X_n^{r_n} dX \text{ , moment of order } r = r_1 + \cdots + r_n \text{ of the}$$

random variable $X = (X_1, \cdots, X_n)$ on $R^n$.

These ones are mutually independent, since they are also the derivates of the indicator function of the random variable $X$. It is used to restrict their computation to the first orders in order not to magnify digitalization noise. Their use will be shown in two cases:

- — in localization ;
- — in particle size analysis.

## 2.6.2. Localization

The computation of spatial object moments enables:

- — on one hand to localize objects existing in the vision space ;
- — on the other hand to provide about them measurements independent from the geometrical transformations that they may undergo.

In a plane, these transformations are the similarities:

- — translation ;
- — rotation ;
- — scaling.

When the camera is located at a given distance from the scene, this last transformation is not taken in account. The two first transformations are only telling that the object can be put and directed in a random manner in the work space.

These uncertainties will be solved by the computation of spatial moments that are providing:

- — the information about the object localization;
- — and a vector of measurements independent from this one

The moments of an object described in the image reference system are the following ones:

- — at order 0 : $M(1) = \sum_{X,Y \in object} dm$,

- — at order 1 : $M(X) = \sum_{X,Y \in object} X dm$, $M(Y) = \sum_{X,Y \in object} Y dm$,

- — at order 2 :
  $$M(X^2) = \sum_{X,Y \in object} X^2 dm, \; M(XY) = \sum_{X,Y \in object} XY dm, \; M(Y^2) = \sum_{X,Y \in object} Y^2 dm$$

- — at order 3 : $M(X^3)$, $M(X^2 Y)$, $M(XY^2)$, $M(Y^3)$

The moment of order 0 is the object surface: $S$.

The moments of order 1 divided by the object surface are the coordinates of its gravity center : $X_G$, $Y_G$.

After division of higher order moments the object surface and centering relatively to its gravity center, the moments of order 2 represent the inertia ellipsoid of the object from which can be extracted, after solving its second degree equation:

- the angle $\theta$ between the main object inertia axis with the abscissa axis of the image reference frame ;
- $\lambda_1$, $\lambda_2$ the major and minor inertia axes of the object measured in its Eigen reference frame.

After a rotation of an angle $\theta$ of the moments of order 3, these ones:

- enable to solve the uncertainty at more or less $\pi$, by making positive the asymmetry positive along its major inertia axis (direction of the strongest eccentricity) ;
- provide the object asymmetries.

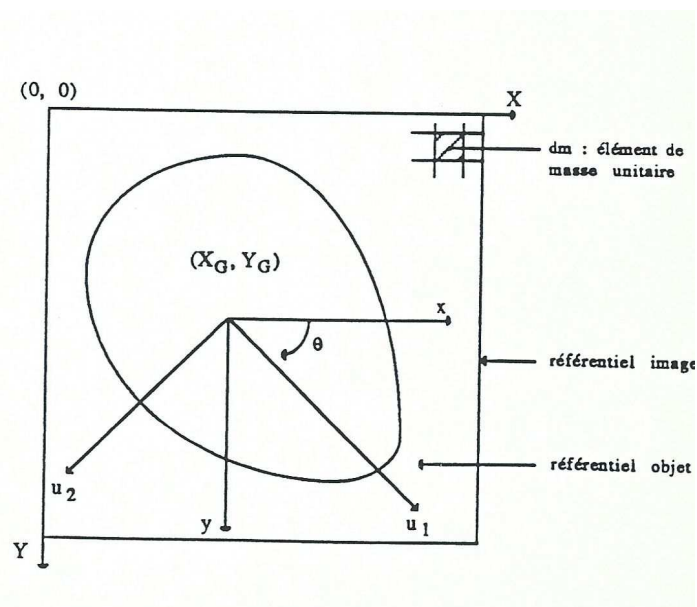

**Figure 4 : Object localization in a visual space**

The gravity center and the Eigen reference frame angle of the object enable to localize it in the visual space. The surface, the inertias and the asymmetries provide a vector of attributes invariant in translation and in rotation.

### 2.6.3. Particle size analysis

In particle size analysis, the interest is not focused on some given objects, but on the distribution of objects in the image : it is the case when it is tried to estimate the state of an area (for instance in mineralogy or in crystallography).

Numerous objects can be seen in the image and only their surfaces are computed. Having got these values, it can be deduced:

- $\mu$ the mean surface of the objects ;
- $\sigma$ the dispersion around the mean surface (standard deviation) ;
- $\alpha$ the asymmetry relatively to the Gaussian distribution $N(\mu, \sigma)$ .

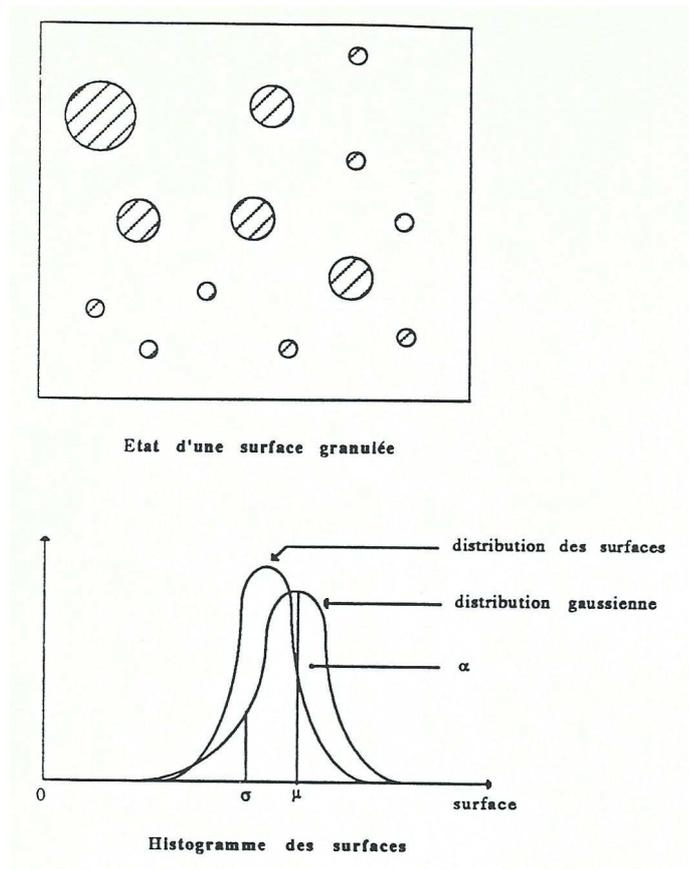

**Figure 5 : Particle size analysis**

## *2.7. Statistical pattern recognition*

### 2.7.1. Presentation

We will now pay attention to the use of data analysis methods for solving decision problems raised by the three following application domains:

- metrology ;
- inspection ;
- identification.

In metrology, we are going to show how to perform the calibration of a vision sensor. In inspection, we will describe an authentication method. For these two application domains, we will rely on the technique of least squares and for identification on discriminant analysis.

### 2.7.2. Calibration

Concerning this problem starting from observed measurements, it is tried to make a vision system able to define a new measurement, straightly inaccessible. It is for instance the case in crystallography where, according to the particle size analysis of a crystal surface, it is willing to determine the time reached during a crystallization process.

Let us assume that the observed measures are the moments of the distribution of the surfaces in an image, by fitting in the least squares meaning (linear regression), we are able to model the crystallization duration according to the observed measures In order to do it, it must be done a learning step over several crystallization processes, where it will be registered the measurements performed by the sensor and the time associated to each view take.

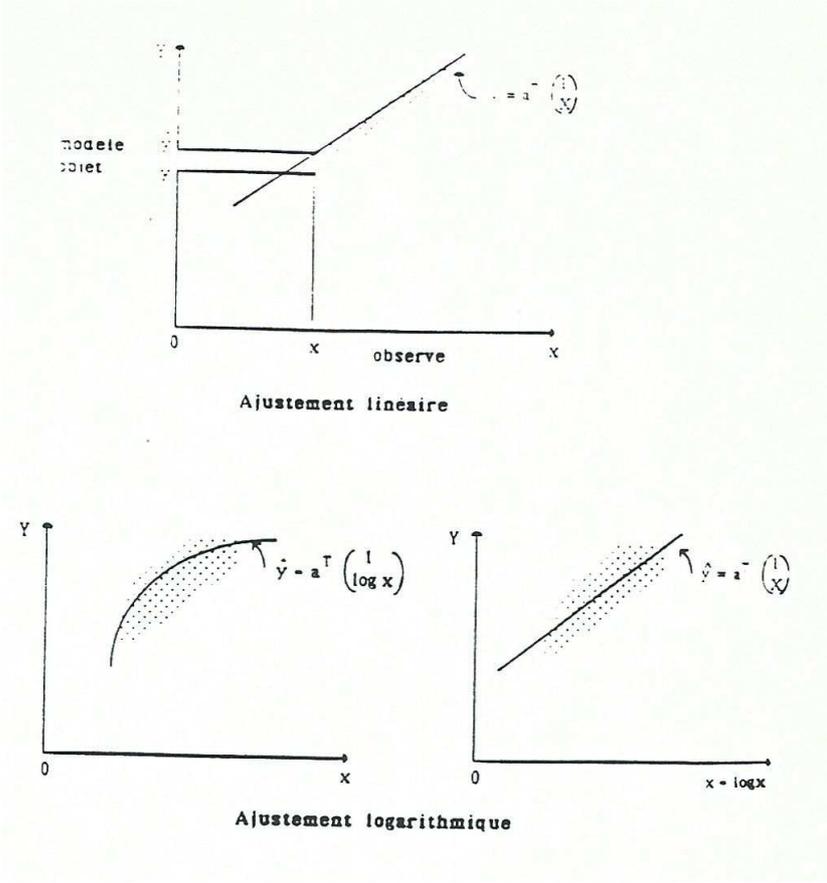

**Figure 6 : Sensor calibration**

Several view takes are so performed for each process. The training set is made from experiences $y_i, (x_{1i}, x_{2i}, \cdots, x_{pi})$, where:

- $y_i$ is the time observation of the $i$-th experience ;
- $(x_{1i}, x_{2i}, \cdots, x_{pi})$, the vector of measures performed by the sensor during the same experience ;
- $i$, the number of the experience in the learning set.

The use of a least-squares adjustment will enable to find out the linear dependency that exists between the variable to be estimated $y$ according to the explicative variables $(x_1, x_2, \cdots, x_p)$ while minimizing the adjustment error $e$ :

$$\hat{y} = a_0 + a_1 x_1 + a_2 x_2 + \cdots + a_p x_p + e$$

When the dependency is not linear, the learning data are transformed in order to come back to a linear model:

$$\hat{y} = a_0 + a_1 \log x_1 + a_2 \log x_2 + \cdots + a_p \log x_p + e \text{, for a logarithmic adjustment.}$$

### 2.7.3. Authentication

In inspection, it is tried to make a vision system able to discriminate if an object is compliant or not to a standard shape. In order to do it, it can be once more to rely on a least-squares adjustment but now for evaluating the error comparatively with the standard model.

First an initial learning step is performed on compliant objects in order to determine the linear dependency between the measurements applied on the views of these objects. That is when it is tried to model one of the measures relatively to the other ones, for instance the object surface comparatively to the moments of higher order:

$$\hat{S} = a_0 + a_1 \lambda_1 + a_2 \lambda_2 + a_3 \alpha_1 + a_4 \alpha_2 + a_5 \alpha_3 + a_6 \alpha_4$$

When the object is compliant, the approximation error $\left| S - \hat{S} \right|$ is minimal.

A second learning step is performed using compliant and not compliant objects by labeling them. This learning step then enables to compute an estimate of the compliance threshold. If the histogram of error distributions for the two object classes is analyzed, a plausible value for this estimate is:

$$s = \frac{\sigma_2 \mu_1 + \sigma_1 \mu_2}{\sigma_1 + \sigma_2} \text{, for two Gaussian distributions } N(\mu_1, \sigma_1) \text{ and } N(\mu_2, \sigma_2)$$

The authentication decision procedure is if $\left| S - \hat{S} \right| < s$ then the object is compliant, else the object is not compliant.

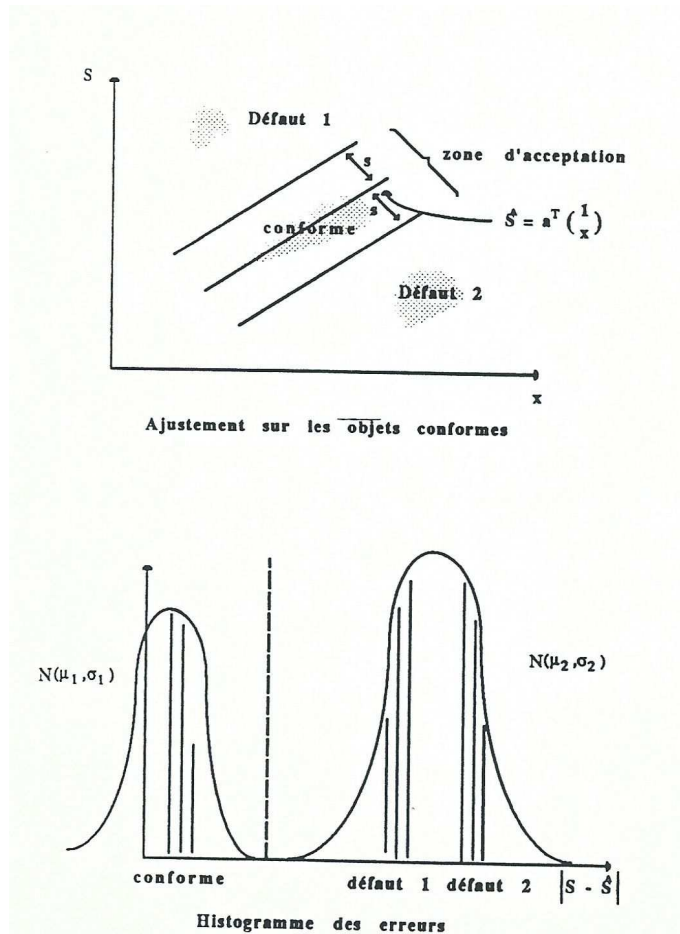

**Figure 7 : Authentication**

### 2.7.4. Identification

In object sorting or identification, it is expected to distinguish different objects between each others.

To each distinct object, it is assigned a class $C_k$. The recognition universe is then the set of the $m$ classes $\{C_1, C_2, \cdots, C_m\}$.

For these m objects, we have at our disposal a representation space, the space of measures like for instance the moments' space. For building it, it will be relied on discriminant analysis, so as to evaluate the probabilities of object belonging to the classes according to their attribute vectors.

The membership probabilities of a measure vector $X$ to the classes $\{C_k\}$ are represented by the vector of dimension $m$ :

$$V = \big(P\big(X/C_1\big), P\big(X/C_2\big), \cdots, P\big(X/C_m\big)\big)$$

During the learning phase, it is assigned at each experience a priori values as probability vector:

$$W_1 = (1,0,\cdots,0) \text{ if } X \in C_1 ;$$

$$W_2 = (0,1,\cdots,0) \text{ if } X \in C_2 \,;$$
$$\cdots$$
$$W_m = (0,0,\cdots,1) \text{ if } X \in C_m \,.$$

If the learning set is the set of measures $\{x_1, x_2, \cdots, x_n\}$, then it can be deduced the extended learning base:

$$\{(x_1, v_1), (x_2, v_2), \cdots, (x_n, v_n)\} \text{ where } v_i = w_j \text{ if } x_i \in C_j$$

Applying discriminant analysis, it is consisting in computing the linear estimate $\hat{v}$ of $v$ in the least-squares meaning. The estimate is made from the set of discriminant functions $f_j$ :

$$\hat{v} = \begin{pmatrix} f_1 \\ f_2 \\ \vdots \\ f_k \end{pmatrix} \begin{pmatrix} 1 \\ x \end{pmatrix} \quad \text{where the } f_j \text{ functions are row vectors..}$$

By computing the expression :

$$\hat{v} = R_{vx} R_{xx}^{-1} (x - M_x) + M_v \,,$$

where

$M_x$ is the mean of attribute vectors,

$M_v$ is the mean of vectors of a priori distributions,

$R_{vx}$ is the covariance matrix of variables $v$ and $x$ ,

$R_{xx}$ is the variance matrix of the variable $x$ .

The recognition procedure is performed by looking for each new experience which $w_j$ is the nearest from $y$ , that is to say which component of $\hat{v}$ is getting the maximal value.

As for authentication, it is done a second learning step in order to assign a threshold $s_j$ at each discriminant function. They are computed by comparing each class $C_j$ to all other classes. This procedure enables to bound the recognition space to the known objects of the recognition space.

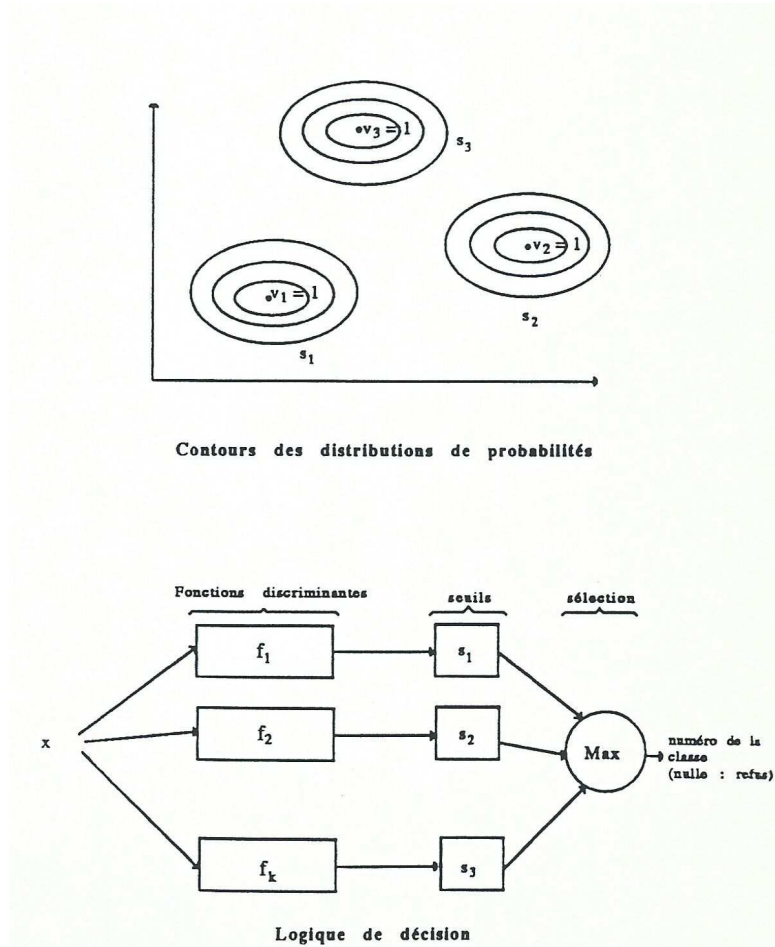

**Figure 8 : Identification**

## 2.8. Description of an industrial application

It is the automation application for decorating boxes of chocolate candies implemented in collaboration with an Alsace automation enterprise.

The aim was to automatize a work station in a manufacturer of the chocolate making field. This manufacturer is carrying out boxes of chocolate candies usually bought to be offered.

Most of purchases are made for two annual events, at Christmas and at Easter, and the product is perishable. So this activity is strictly seasonal as the manufacturing staff. For the remaining periods in the year, the tasks to be performed are mainly market researches in order to define new chocolate candies and new packaging models so as to find new clients or to keep the older ones.

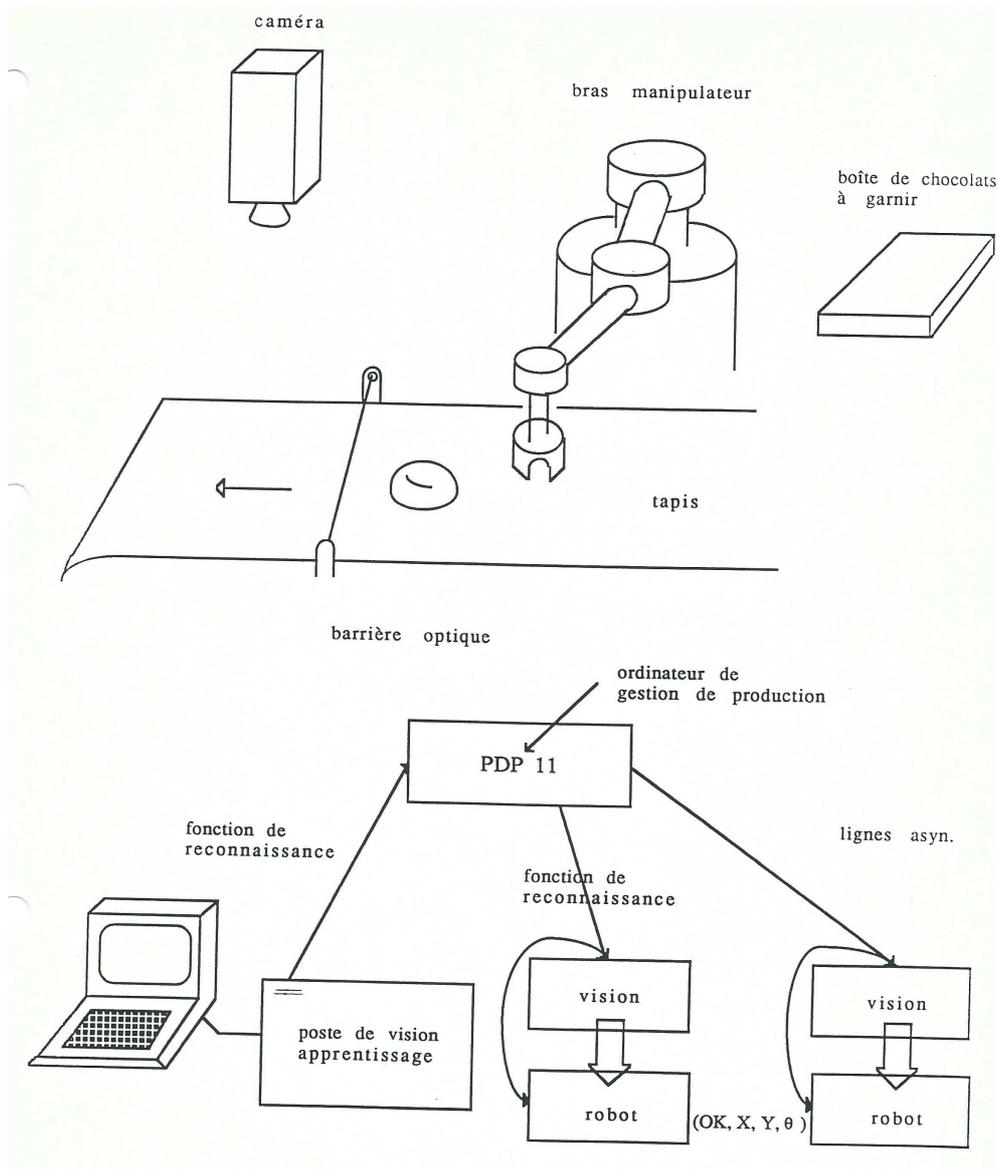

**Figure 9 : Decorating boxes of chocolate candies**

After being molded, the chocolate candies are dressed with a paper or aluminum packing, then put down in dedicated alveoli in an inbox. The work station consists in :

— inspecting of the candy after dressing ;
— putting the candy in its alveolus.

The decorating station is composed of:

— a transport belt including a ginning machine ;
— a vision system with an optical sensor for detecting a candy arrival in the field of view ;
— a fast mechanical arm of SCARA kind (all rotating axes, except the last one) ;
— a receiving stand in which several boxes can be put.

The full cycle lasts one second for processing one chocolate candy.

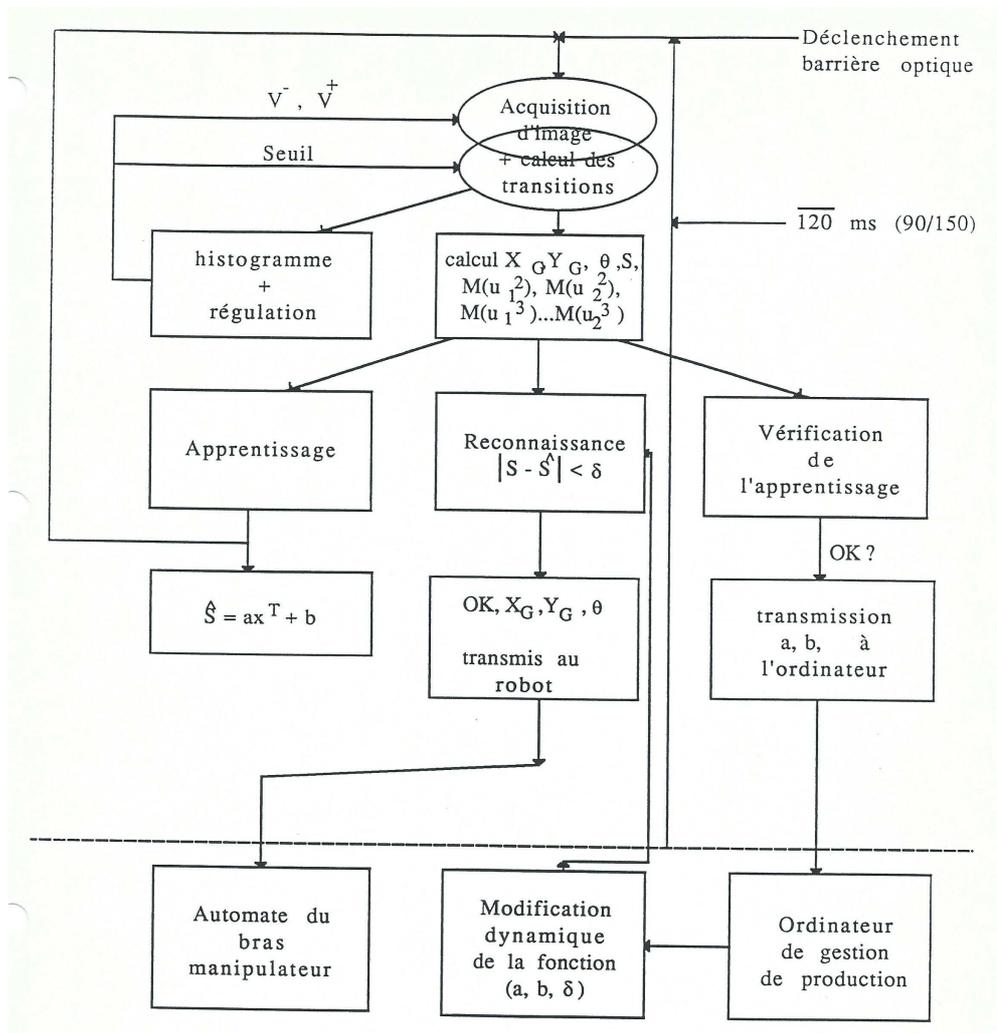

**Figure 10 : Processing scheme**

500 chocolate candy kinds are used for filling boxes. The decorating stations are controlled by a manufacturing management computer. The engineering and design department has at its disposal its own vision system in order to perform the learning steps for any new candy to take in account.

The vision system localizes the chocolate candies in position and in orientation and checks its compliance (authentication). If the chocolate candy is compliant, it is picked and put down at its right place in a box. The supply of a work station is made from preprogrammed sequences selected by the manufacturing management computer. This one is holding the recognition functions and dynamically reconfigures the vision systems between each sequence.

When an error occurs, this one usually leads to the destruction of the receiving chocolate candy box. Concerning this particular application, the authentication error rate has been valued at less than 1/10000.

## *2.8. Conclusion*

The methods of statistical data analysis can be efficiently used in pattern recognition for applications in artificial vision.

Concerning the described methods, the recognition universe should have preferably a low dimension (classes made from only a few units).

Other methods can be envisioned, but these ones are less easy to implement.

# 3. Cataloging System of Satellite Data

## 3.1. Presentation

### 3.1.1. Survey aim

Development of a system capable to quickly perform a preprocessing task on satellite images enabling to build a catalog of the data stored by an archiving center and providing the opportunity to propose a second level archiving facility.

### 3.1.2. Research objectives

It is developing a cataloging system for satellite images enabling to:

  — provide preprocessed data catalogs on the whole registered images;
  — develop a content-based query function ;
  — control the description precision of data belonging to the catalog;
  — propose a solution for enabling a second archiving functionality.

Data registration and query in an archiving center can be organized around the following activities:

  — data archiving ;
  — data cataloging ;
  — and data querying.

Data archiving can be dividing in two levels:

  — at first level, it is recent information stored by keeping the data capture format ;
  — at second level, it is replacing information stored in a compressed form.

Data cataloging includes three distinct operations:

  — data signaling that enables to know in which conditions and when data has been gathered;
  — data indexing, necessary for enabling any future search ;
  — data summarization for having at one's disposal an insight on it.

It can be query data according to :

  — the kind of data access, address-based or content-based query;
  — the request complexity, simple or complex query.

Address-based interrogation is usually performed with the help of signaling information.

Content-based interrogation is assuming that it is possible to address simultaneously the whole data in the base in order to compare it with some given patterns and to select pieces that get closer to them: for instance, query by example offers a content-based access for a data base..

The production of complex queries can be led in combining simple queries within a relational algebra: that is the case of structured interrogation languages.

A content-based query needs to have settled an information thesaurus enabling to index the pieces of information belonging to the registered documents, satellite images in the present case.

Such a thesaurus can be settled by:

— decomposing images to be archived into homogenous components ;
— computing geometric or radiometric characteristics enabling to measure the components stemming from this segmentation process.

These components will enable to characterize and to sort the shapes in such a manner that it will be possible to simply identify and localize them among the pieces of information registered in an image base.

At half-way from the representation by a vector of measures computed over the surface of objects stemming from image segmentation and from the one made from a table of luminance values sampled in row and column representing an image, the luminance means and the decomposition into line segments of the contour of objects coming from segmentation, enables to provide an intermediate representation model whose size is varying according to the approximation precision (contour vectorization).

With the help of this representation model, it can be then provided an overview about the content of an archiving system, that is compliant with the usual graphical models, as for instance it is for the information handled by a cartographic information system.

Luminance means and decomposition into line segments are approximations:

— at order 0 for image luminosity ;
— at order 1 for object contours stemming from image segmentation.

That is also the result of a finite development in series of a piecewise surface.

So in order to insure a coherence between segmentation and approximation, the segmentation of images, during their registration, should be applied according to a regularity predicate (in the meaning of surfaces regular up to a given differentiation order). It would then enable to include in the future the results got in approximation of line and surface structures.

After the development of a procedure measuring objects produced by image decomposition and their storage in data base according to the order induced by this computation result, it can be envisioned to set up a content-based query system of an image base :

— by nearest neighbor querying, in computing the measures of a prototype shape in order to retrieve the closest occurrences from the image base ;
— by multi-criteria selection, in order to extract the subset of shapes belonging to the base whose measure vector could belong to the Cartesian product of value intervals related to these measurements.

In the two cases, only connected shapes, which are simple in term of complexity, will be delivered after a query. They can be used as a basis for building a query system for searching for complex shapes, consisting in the union of several simple shapes sharing between them closeness relationships: such queries could be described using a language based on a relational algebra.

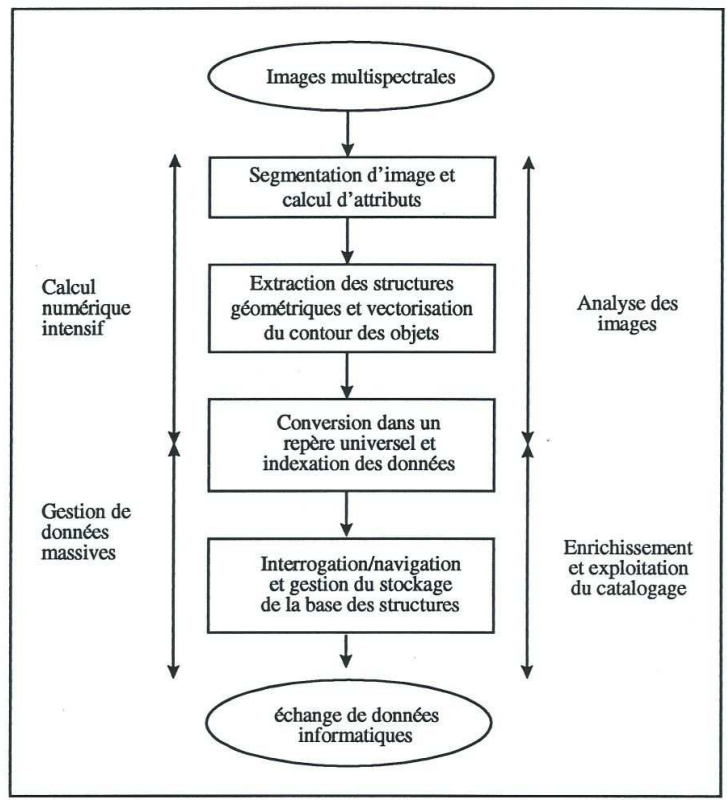

**Figure 11 : Cataloging satellite images**

The set-up of a measure thesaurus enabling to index the set of simple shapes belonging to an image collection induces the following facts:

— any image having to be registered must be segmented and the objects coming from this decomposition measured ;

— despite the expectancy that can be awaited in information compression, an effort in system architecture must be envisioned if it is willing to be able to master the response time of queries.

So some specific efforts have been provided during the study:

— in process parallelization for image analysis and for registering data ;

— in parallelization of data management procedures for interrogating the image catalog.

Three points of view should be conciliated during these operations:

— the location of the observation system during data capture (satellite reference frame) ;

— the recording reference frame (universal reference frame) ;

- the observer location when rendering the results of a query (display reference frame).

The geometric transformations to be performed for moving from one reference frame to another one are also addressed.

### 3.1.3. Previous works history

The project of satellite image cataloging is relying on the previous following developments:

- a software for processing multidimensional information by the means of tree-like structures (the KDTREE software - Contracts CELAR-ADERSA n°005/41/84 and n° 004/41/88) ;
- a library for image processing suited to the generation of cartographic data from images captured by an earth resource observation satellite (the STCART library - Contract CELAR-ADERSA n° 012/41/88).

The KDTREE software has been designed in the aim to provide a tool enabling to model multidimensional numerical objects. The used modeling approach is based on the "divide and conquer" paradigm and leads to the building of trees of order $2^k$ in order to represent numerical objects belonging to spaces of dimension k. It is a generalization of quaternary and octernary trees ("quadtrees" and "octrees") that can more often met in the plane or the 3D space for image analysis.

The STCART library has been developed in the aim to provide a tool enabling to generate information of cartographic kind from aerial or satellite images. This one is relying on the KDTREE software for providing a thematic conversion procedure and is made from planar image analysis functions. The analysis result is provided in a vectorized way so as to be easily inserted into a geographical data base

Thereafter, works about the parallelization of the KDTREE software have been performed on two different computing systems (Contract DRET-ADERSA n° 90/34/106) :

- a distributed memory synchronous parallel computer, the CONNECTION MACHINE 2 ;
- a distributed memory asynchronous parallel computer, the T.NODE.

These works have led to the full parallelization of KDTREE software on the T.NODE computer (Contract CELAR-TELMAT INFORMATIQUE n° 028/41/91).

Besides, TELMAT INFORMATIQUE has contributed to the European project MARS, dealing with the remote sensing in the service of the geographical information used for agricultural purpose. TELMAT INFORMATIQUE has participated to the parallelization of the preprocessing software dedicated to NOAA-AVHRR images on T.NODE computer (radiometric and geometric corrections, simplified thematic classification).

### 3.1.4. Previously obtained results

Attribute calculus is already used for a long time in ADERSA, especially in the framework of the industrial applications of artificial vision ("Processeur d'Image Micro-programmable", Research contract ADI-ADERSA n° 82/227). It is mainly relying on the computation of generalized moments. The centered and reduced generalized moments enable to have at one's disposal a vector of measures invariant to similarities and mutually independent, as well as localization information about the location and the direction of objects in their observation reference frame.

In connected component search, two distinct approaches can be envisioned:

- a border approach, where it is relying on object contours ;
- a regional approach, where the object interior is straightly processed.

The structure of industrial scenes and the demand of minimal response times lead us to master in a first step image segmentation favoring the border approach.

Then the works focusing on the development of an image processing library suited to the generation of cartographic data allow us to deal with the regional approach through the implementation of a blob labeling algorithm.

The border approach has been complemented with the development of a border vectorization algorithm with error control relying on the "strip-trees" technique.

During this study, the interest was mainly focused on the segmentation of multispectral images. It was proceeded by thematic classification and conversion.

The thematic classification consists in identifying in the radiometric space the luminance distributions linked to the multispectral responses of terrestrial covering (urban areas, vegetation, rivers ...). This classification is performed by using the capabilities of hierarchical modeling and multidimensional segmentation of the KDTREE software. The same approach applied on a mono-spectral image is equivalent to perform a monochromatic segmentation by multi-thresholding.

With the works done for developing the KDTREE software, it has been possible to create the means enabling to model regularly sample data sets in spaces of any dimension. It has been taken in account of the works that has been already made with this kind of data structure in order to handle cartographic information and to model tridimensional environments, especially in the aim to apply geometric reasoning tasks.

So we have been interested in the applications of computing inter-visibility graphs relatively to a digital terrain (simulation of aero-terrestrial battles, tactical management of electromagnetic spectrum of an army corps, contract SEFT (Section d'Etudes et de Fabrication de Télécommunications) n°92.50.135.92 performed in collaboration with the enterprise ADV TECHNOLOGIES).

Moreover, it has been provided a significant effort to the mastery of distributed synchronous and asynchronous parallel computers. We have been led to revise some algorithms, especially by developing a multidimensional algorithm of region growing for which the transposition in two dimensions have been easily realized, enabling to reduce or to suppress the irregular communication schemes between processors implied in regional segmentation, that are at the origin of message collisions during communication.

On asynchronous, developments have been made on the range of T. NODE computers of TELMAT INFORMATIQUE. The used implementation model is a SPMD kind model including the simulation of a global memory and the emulation of an OMEGA multi-stage interconnection network using a communication protocol by message of "store-and-forward" kind.

### 3.1.5. Research progress

#### 3.1.5.1. Presentation of the different steps

This research is built around three major stages, among which the first ones have been performed simultaneously:

- prototyping of the chain of data storage and preprocessing ;
- parallelization of the storage system ;
- testing and synthesis.

The methodology proposed for cataloging satellite images is developed on a sequential computer during the first step. Its adaptation to parallel architecture computers is performed during the second step. The results of these works are assessed during the last step. Let us detail each of these major stages.

#### 3.1.5.2 Prototyping of the chain of data storage and preprocessing

**Selection of data sets**

An evaluation data set is made up with the help of the Cellule d'Étude en Géographie Numérique (CEGN).

The structure the exchange formats or the reading procedures have been communicated for enabling data exploitation.

Notably, it is necessary to access to the signaling information of images:

- location and attitude of the observation point ;
- field of view and date of the view take.

**Implementation of the thematic conversion chain**

Implementation of a thematic classification procedure so as to realize an image segmentation based on homogeneous luminance and radiometric responses classes.

**Implementation of a vectorization chain**

Implementation of procedures enabling to:

- compute the attributes of the components stemming from segmentation (i.e. surface, gravity center, inertia axes, orientation angle, asymmetries of objects) ;
- find the object contours and vectorize this information.

**Development of a geometric conversion procedure**

A geometric conversion procedure, taking in account the location of the observation center, is developed for replacing vectorized data into a universal reference frame.

**Development of a data indexing procedure**

A two-scale indexing procedure relying on a data hierarchical modeling enables to provide :

- a major indexing, based on the image location in the catalog universal reference frame and on the attribute histogram of objects belonging to the image.
- a minor indexing, based on the localization of geometric structures belonging to the image with their attributes.

**Implementation of a registration procedure and development of a catalog geographical browser**

The aim is to implement the data registration procedure that can build up a catalog and to develop the querying procedures that will restitute the information belonging to it.

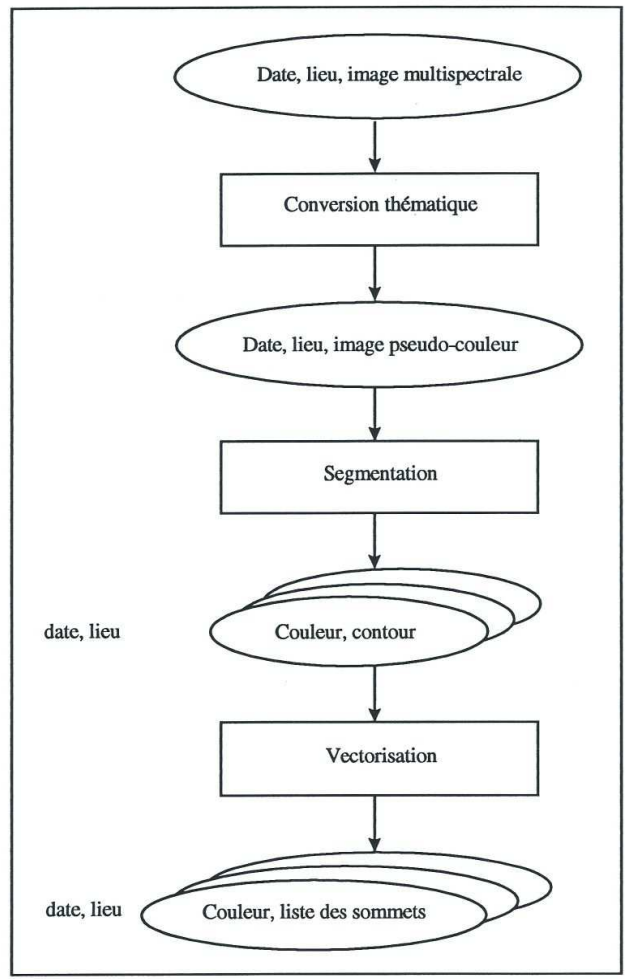

**Figure 12 : Processing description (I)**

### 3.1.5.3. Parallelization of the storage system

**Migration the KDTREE software**

The KDTREE software parallelized on T.NODE computers, designed around the INMOS T800 processor, has been ported on the new TELMAT range of TN parallel computers, based on the SGS-THOMSON - T9000 processor.

**Data recording system on parallel computer**

The data recording system, built from a parallel file management system capable of handling simultaneously several disks, is studied in order to optimize the access time to image data.

**Distribution of image data over a disk parallel system**

The distribution of image data over a disk parallel system is studied in order to optimize image access performances.

**Parallelization of the image analysis software**

It consists in parallelizing the image analysis functions necessary for building an image catalog.

**Tests on parallel computer**

The whole set of parallelized function is integrated and evaluated.

**Processing optimization**

A performance analysis which is highlighting the ratio between the time spent in computing and the time spent in communication, is performed.

### 3.1.5.4. Test and synthesis

**Generation of a catalog and performance evaluation**

A significant set of images is acquired from the C.E.G.N. A catalog is generated on two different computing systems, one with a sequential architecture, the other one with a parallel architecture. The obtained results are compared and analyzed.

**Interrogation of a catalog and evaluation of performances**

Having at one's disposal of a same catalog handled with two distinct computing architectures, data interrogation trials are realized and the behavior differences are analyzed.

**Synthesis**

A demonstration enabling to build and to interrogate an image catalog is set up.

The result synthesis has allowed to:

– define the system dimensioning formula according to the processes to be performed ;
– list the complementary works that could be envisioned for extending the application spectrum of the system.

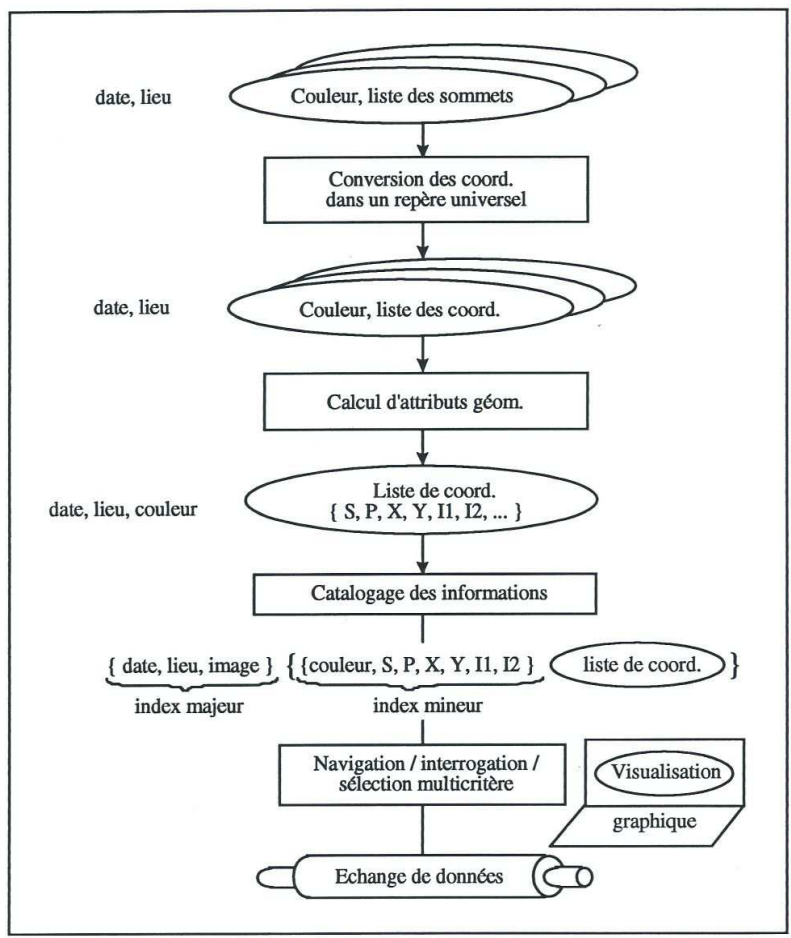

**Figure 13 : Processing description (II)**

## 3.2. Thematic conversion

### 3.2.1. Thematic conversion goals

The initial goals of thematic conversion have been to provide the means enabling to decompose a multispectral image into a partition of homogenous components from which it could be retrieved:

  − their radiometric responses ;
  − their shapes ;
  − their locations.

The continuity of radiometric responses is the predicate which is used to perform an image segmentation into homogenous components.

This property can be expressed in performing a classification of radiometric responses in their Eigen space. This classification is led by analyzing the proximity of these data vectors. When these response classes have been identified, the thematic conversion then consists in producing a new image made from the labels of classes to which radiometric vectors are belonging.

From a multispectral image, it is got a thematic monochromatic image. The responses, that can be associated to these themes, are the responses corresponding to the classes' centers.

After looking for the connected components of the thematic image, the shape of components can be evaluated:

  − globally, by attribute calculus ;
  − locally, by decomposing the component boundaries over a set of geometrical primitives, usually line segments (piecewise linear decomposition of a contour).

From attribute calculus, it can be deduced information about the location of components in the image.

The component boundary decomposition using a set of geometrical primitives enables to provide a description of the components into geometric structures.

A classification of geometric structures belonging to an image is got by indexing the structures by their location in the image and a vector of measurements independent from its location stemmed from attribute calculus. This classification can be extended to an image collection by complementing the information of location in an image by the information of the image place inside some given reference frame.

If we are coming back to the thematic conversion and the classification of image points' responses in the radiometric space, it can be seen that the use of this procedure is restricted to the use of multispectral data. It is widely exploited in earth resource observation, but it is not easy to extend and apply it on a classification performed over a single image up to a collection of images taken at different dates and at different places. Actually, in passive imagery, the radiometric responses of a same theme, especially for those belonging to the vegetal coverage, are evolving according to:

  − its date during a year (seasonality) ;

- its time during a day (solar height) ;
- the atmosphere conditions (humidity) ;
- the relief (specular or diffuse reflection).

The statistical indices that take in account measurement ratios between different spectral bands may be favored to a global statistical analysis when it is tried to follow the coverage evolution of some vegetal species. These calculations can be correlated with ground surveys done at the same places and the same periods

In digital cartography, the interest is naturally focused on the search and the watching of engineering structures or human manufactured industrial goods. This kind of buildings gets a geometrical shape much more regular than natural ones. The identification of such objects leads to favor high-resolution mono-spectral images to middle range multispectral ones.

The development of new satellite observation systems and the multiplication of observation systems standing in orbit lead to try and superimpose data stemming from different sensors and to attempt and pair geometric structures discovered in these observations. Particularly, it can be tried to compare artificial structures registered in a geographical information system with those observed by a satellite or to transfer this information over a digital elevation model.

| Spectral bands | LANDSAT-TM | SPOT |
|---|---|---|
| UV 290 nm-400 nm<br>visible RGB 400 nm-700nm<br>near IR 700 nm-900 nm<br>mid IR 900 nm-5,5 nm<br>thermal IR 8 μm-14μm<br>* | 1   450-520 nm<br>2   530-610 nm<br>3   620-690 nm<br>4   780-910 nm<br>5   1.57-1.78 μm<br>6   2.10-2.35 μm<br>7   10.4-12.6 μm | 1   500-590 nm<br>2   615-680 nm<br>3   790-890 nm<br>P   510-730 nm |
| geometric resolution | 30 m x 30 m<br>except last band<br>(120 m x 120 m) | 20 m x 20 m   XS<br>10 m x 10 m   P |
| average observed area | 180 km x 180 km | 60 km x 60 km |

(*) : UV for ultra-violet -  RGB for red, green, blue - IR for infra-red

**Table 1 : Sensing spectral bands for two different passive observation satellites**

The passive observation satellites do not always analyze a signal in the same frequency range; at the opposite radar satellites can capture images independently from solar illumination and meteorological conditions.

In these conditions, it may be preferable to keep analysis capabilities on each spectral band that is taken alone, within a multispectral image, rather than to apply at first a classification mixing information belonging to several distinct bands.

### 3.2.2. Methodology applied for analyzing images

We have applied a thematic conversion procedure based on the statistical analysis of data in its radiometric space. This approach has been used on two main data sensing sources:

- the multispectral images provided by the SPOT satellite range ;
- the images produced by LANDSAT satellites and its derivates.

In order to keep the same approach on mono-spectral data, we have developed an image decomposition based on the analysis of the first derivate orders of a monochrome image. The spectral bands previously processed using statistical analysis are then replaced by the following images:

- average image (integration) ;
- image differentiated at order 1 (Sobel filtering) ;
- image differentiated at order 2 (Laplacian filtering).

The classification applies no more on the identification of similar radiometric responses, but on the identification of similar integral-differential behaviors.

### 3.2.3. Multispectral image analysis

In planar multispectral image analysis, we have used until now multidimensional statistical analysis methods with the help of the KDTREE software, in order to implement the thematic conversion stage of a multispectral image.

It has led us to interpret the image analysis as a sub-case of multidimensional statistical analysis, where the data is regularly sampled over a planar support.

It has then be possible to set a bridge between operators used in two dimensions and operators used in regularly sampled multidimensional spaces and to retrieve, with the help of this approach, several classical results from statistical data analysis (hierarchical classification, data partitioning, factorial analysis, etc.).

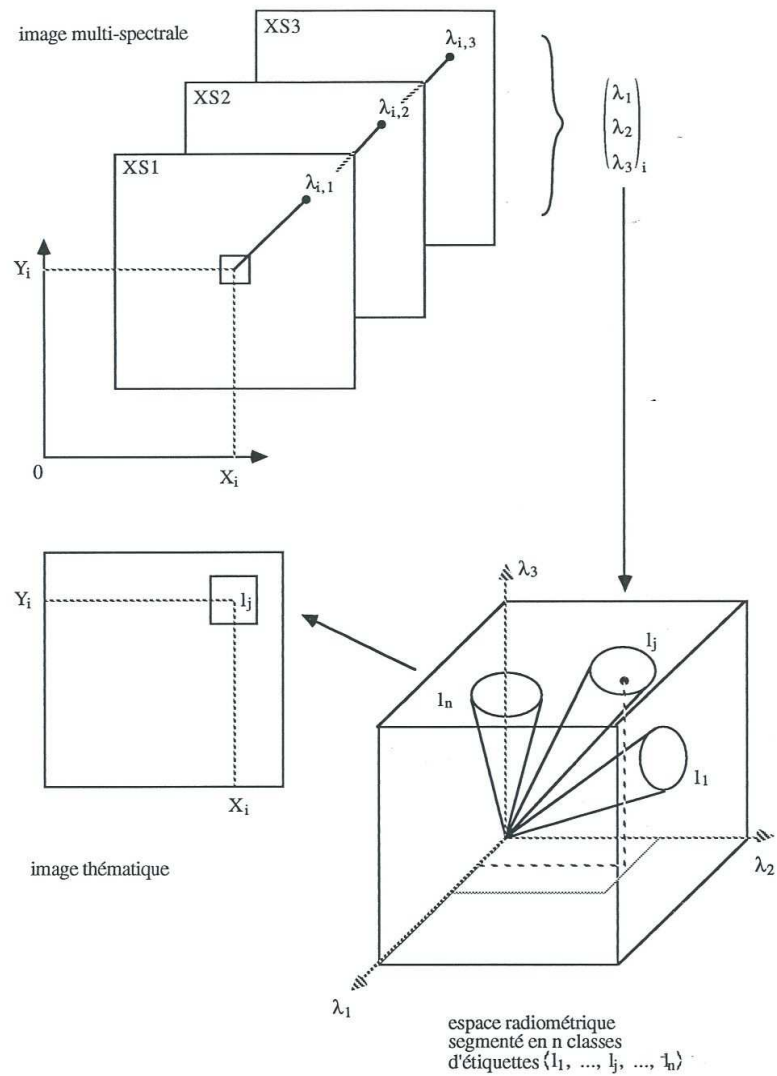

**Figure 14 : Thematic conversion**

In such a framework, it is then possible to mix data from different sources by taking some precautions and to process these ones in a more general way: luminescence, region labels, attributes computed over these regions.

Some troubles occur when using this approach: if it was actually possible to handle huge information volumes according to this strategy, some processing operators offer polynomial response times but not linear or as if linear from the space dimension and have reduced the use that can be done with this kind of representation (for instance the search for adjacencies in a multidimensional regularly divided space).

In order to solve this problem, we found out that the description axes handle independent information; then it was possible to preserve their relation by analyzing the space dimensions sequentially two by two ones, but with introducing from the second step the result of the processing of the two previous dimensions as one of the two new operands.

So, it has been possible to develop a technique of classification based on the analysis of k-1 planar projections by re-introducing the labels produced by previous projections as one of the two current dimensions during the processing of the k-th projection.

The segmentation tools used in two dimensions can be then extended to any number of dimensions while offering a processing cost which is linear of the dimension of the handled space.

At the opposite, it implies to process each dimension orthogonally to the other ones, which is restricting it to an analysis done in $d_1$-adjacency and splitting line structures belonging to the image.

In this aim, it has been used tools enabling to create artificial images whose :

- planar support is built around one axis devoted to the labels belonging to the previous image and the other one to the grey levels or attribute values belonging to an image which is directly related to ;
- the luminescence values of the new image are counting the number of occurrences of identical couples of values discovered in the two initial images.

It is a bi-dimensional histogram, on which can be applied the whole set of functions used in image analysis, notably for performing a segmentation and a connected component labeling

Restarting with the two initial images, it is then possible to generate a new image labeled according to the segmentation of the bi-dimensional histogram relying on the support space common to the two initial images.

In image synthesis, it is possible to reconstruct images modeled by:

- polygon filling, for vectorized geometric structures ;
- digitalization of piecewise constant or linear surface models.

### 3.2.4. Mono-spectral image analysis

According to the conditions of view takes and the processed scenes, it can be distinguished several cases:

- when the luminous ambience is controlled and when the scenes are built on the observation of manufactured objects, the analysis of the image grey level histogram is enough for properly classify image elements ;
- when the luminous ambience is not controlled and when the scenes are built on the observation of natural situations, the analysis of the image grey level histogram is not enough for classify the information belonging to the image.

The analysis of the image grey level histogram enables:

- when the thematic is summarized to a binary situation, to find a threshold separating the objects from the background, to binarize the image and to perform a border following of the objects belonging to the image ;

– when the thematic is multi-valued, to identify the modes existing in the histogram and to segment the image according to the identified maxima.

When the analysis of the grey level histogram does not enable by itself to retrieve the information standing in the image, two approaches based on local analysis enable to process this image:

– the search for singularities standing in the image, that can be got by identifying the continuity breaks inside luminous surfaces ;
– the search for regions sharing a same integral-differential behavior.

The search for singularities consists in identifying the continuity breaks:

– in the initial image, when it is looking for identifying the boundaries of luminous surfaces ;
– in differentiated images, when it is focused on discontinuities of higher order.

If it is the tried to approximate regular areas by surface polynomials, it is then necessary to take in account the discontinuities of the initial image and all its derivatives up to the chosen approximation order. It would rather remain on low orders of polynomial approximation in order to minimize the computation load. This approach insures a high quality to model fitting, especially in the neighborhood of irregularities.

Looking for regions sharing the integral-differential behaviors should enable to improve the fitting quality of models that would be interpolated over these regions.

The perception procedure is taken in account the functionalities of bi-dimensional histogram computing and image coding starting from the labeling performed on this histogram that have been presented in the previous chapter :

– computing images by integration, then differentiation up to the expected order using convolution ;
– at each order incrementation, computation of the bi-dimensional histogram associated to the couple of values currently processed (last labeling, modulus of the image differentiated up to the current order) ;
– histogram thresholding (quantification) or looking for local maxima (classification), then extending labels of found classes for filling the lack of information in the label image ;
– re-encoding the initial image with these labels by scanning the image couple having enabled to compute the bi-dimensional histogram.

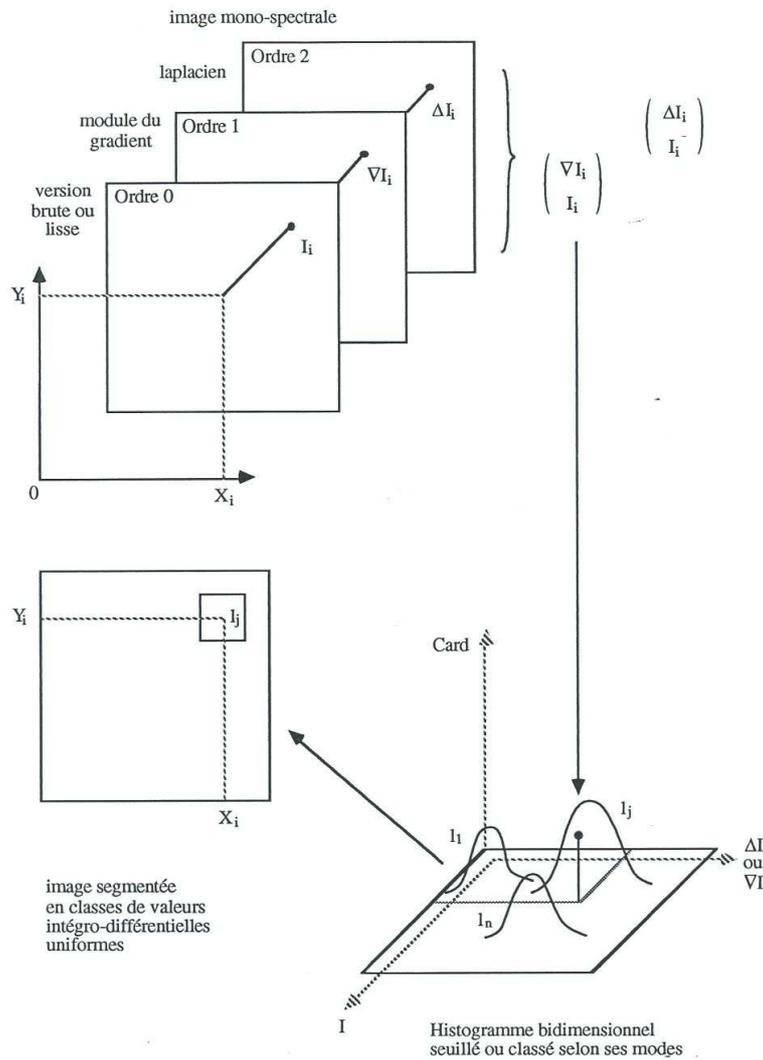

**Figure 15 : Mono-spectral segmentation**

## 3.3. Different used representation models

### 3.3.1. Presentation

Initially images are available under a cellular representation model: the data is made of digital values sampled over a regular mesh in the capture reference frame. This representation model is kept in regional analysis: after blob labeling, it provides a pseudo-color image as a result.

In the case of real-time applications, it may be preferable to use only the information related to the boundary of objects. For compact and planar objects, this data only occupies the square root of the size of the information inside an object.

On the other side, for computing measures on this information, two distinct access manners are necessary:

- a first access by contour following, enabling to perform measurements on the object boundary, as for instance its perimeter ;
- a second access row by row or column by column, enabling to perform surface measurements by evaluating a surface covering made from parallel line segments.

The first access can be provided by moving direction coding, as FREEMAN coding. The second one can be illustrated with run-length coding (RLE).

Based on this former information, it can be built a third representation model which is consisting in vectorized data: the contour of objects is represented using polygonal shapes. It is made from a piecewise linear approximation of the contour: the polygonal vertices are only kept in this kind of model. It is only a specific case among the piecewise polynomial approximation and the polygonal vertices can be seen as control points on which are relying these geometric shapes.

Applying geometrical transformations on all these representation models do not insure the preservation of an exact correspondence (bijection) between the initial data and the transformed data: some information pieces may be gathered, some ones else split.

### 3.3.2. Boundary representation

The representation model, which has been used, is initially divided into two lists:

- the list of row transitions between the scene background and the objects of interest ;
- the list of column transitions between the scene background and the objects of interest.

The first list is registering the coordinates of the horizontal line segments that are recovering an object. The second one corresponds to those of the vertical line segments that are producing the same result. For each of these lists, it is registered in a separate way, the first transition of each row or each column of the image.

By chaining in row or in column the following transitions, the horizontal or vertical line segments can be scanned until the row or column exhaustion. Using contour following, the whole set of points belonging to the two lists be chained so as to be able to visit once and only once all the points of a contour.

In an image, it is registered in a separate way the first transition of each contour existing in the image. By chaining the transitions between them until their starting point, it can be performed a contour following of the object of interest. The row transition list is got by the equivalent of an horizontal gradient. The column transition list is got by that of a vertical gradient. The coordinates of the contour corners appear simultaneously in the two lists. It can be come back to a cellular representation by digitalizing one of the two transition lists (contour filling).

### 3.3.3. Vectorization of boundaries

The used algorithm is based on the "Strip-Trees" technique and allows to control the approximation error according to $L_\infty$ (maximum of errors).

The algorithm follows a hierarchical procedure of decomposition:

- at the beginning, a bounding box (the rectangle enclosing data) is computed ;
- while this one has not got a width lower than the approximation error, its point list is divided into two sub-lists by identifying the point the most distant from the line segment joining the ends of the current list and by using it as midpoint for realizing this division.

When this procedure stops, the ends of the lists holding in each terminal sub-box of the tree deliver a piecewise linear approximation of an object boundary points. The procedure can also work on curves.

Concerning the contour of an object, the points are previously put in the order of border following.. The error computation is known by computing the distance of a point to a straight line, for each concerned list point.

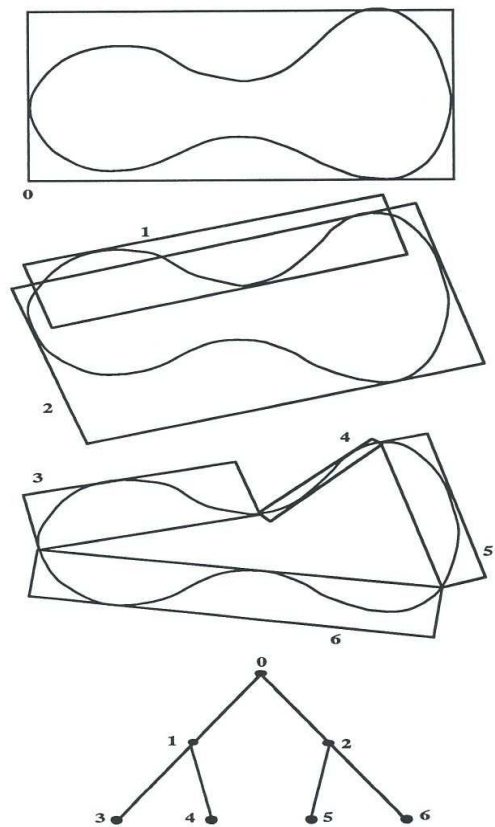

**Figure 16 : Contour vectorization using a « strip-tree »**

### 3.3.4. Vector digitalization

In order to retrieve the lists of contour points, it is necessary to digitalize the line segments of the vectorized representation. In this aim, it has been used a variant developed by LUCAS from BRESENHAM algorithm. Its principle is the following one:

— evaluation of the slope of the line segment ;
— evaluation of the progression direction of coordinates according each axis in order to move from the origin to the end of the line segment ;
— production of as much elementary moves as they are steps to run through by balancing steps relatively to each axis according to a ratio proportional to the slope of the line segment.

Among the elementary moves, it can be distinguished:

— horizontal moves ;
— vertical moves ;
— diagonal moves.

The LUCAS variant make a difference between the cases where:

— the slope of the line segment is below the first bisector line, in order to make only horizontal and diagonal moves ;
— the slope of the line segment is above the first bisector line, in order to make only vertical and diagonal moves.

Diagonal moves are more numerous than moves parallel to axes when it is going closer to the first bisector line. The LUCAS variant has got the property to uniformly spread the interpolating error above the whole set of digitalized points.

The points are then introduced in:

— the row transitions, concerning vertical and diagonal moves.;
— the column transitions, concerning horizontal and diagonal moves.

The duplicate points that may appear at segment ends are deleted, by only keeping only one occurrence of the same coordinates in the transition lists.

### 3.3.5. Use of the representation models

Two targets are looking for with the use of these representation models:

— on one hand, the capability to compress information ;
— on the other one, getting high computing performances in performing several processes.

The first aimed processes are the geometrical transformations.

It is easy to understand that more the information volume is reduced, more quickly these transformations are performed.

At the opposite, it can be observed that some artifacts may occur during these transformations and that it will be necessary to process them:

- — the first defaults are aliasing effects that happen during linear transformations like rotations ;
- — the other troubles get born with no-linear transformations that may destroy the connected structure of shapes.

So :

- — by looking for transitions, an image is converted into transition lists ;
- — by vetorization, contour lists are converted into vector lists ;
- — by vector digitalization, it can be come back to contour lists ;
- — and by filling contours, the initial image is retrieved.

## 3.4. Attribute calculus

### 3.4.1. Object moments

The moments of an object at the order 0, 1, 2 et 3 are given by the following formulas for their continuous expression:

- — $M(0) = \iint_{object} dS$ ,

object surface area ;

- — $M(X) = \iint_{object} X dS$ , $\qquad M(Y) = \iint_{object} Y dS$ ,

from which is retrieved the object gravity center ;

- — $M(X^2) = \iint_{object} X^2 dS$ , $\quad M(XY) = \iint_{object} XY dS$ , $\quad M(Y^2) = \iint_{object} Y^2 dS$ ,

are describing the object inertia ellipsoid in the observation reference frame ;

- — $M(X^3) = \iint_{object} X^3 dS$ , $\quad M(X^2 Y) = \iint_{object} X^2 Y dS$ ,
- — $M(XY^2) = \iint_{object} XY^2 dS$ , $\qquad M(Y^3) = \iint_{object} Y^3 dS$ ;

expressing the different object asymmetries according to the reference frame axes.

From order 2 moments it can be deduced the major inertia axis and the angle that this one is making relatively to the observation reference frame. This one is delivered more or less 180° and the uncertainty is solved by examinating the sign of the asymmetry according to the main axis after being replaced in the object reference frame.

### 3.4.2. Moments computed on a cellular representation

The discretized expression on regularly sampled meshed lattice of these measures becomes:

- $M(0) = \displaystyle\sum_{X,Y \in object} dm$ ;

- $M(X) = \displaystyle\sum_{X,Y \in object} Xdm$    et      $M(Y) = \displaystyle\sum_{X,Y \in object} Ydm$ ;

- $M(X^2) = \displaystyle\sum_{X,Y \in object} X^2 dm$ ,      $M(XY) = \displaystyle\sum_{X,Y \in object} XYdm$ ,

   $M(Y^2) = \displaystyle\sum_{X,Y \in object} Y^2 dm$ ;

- $M(X^3) = \displaystyle\sum_{X,Y \in object} X^3 dm$ ,      $M(X^2 Y) = \displaystyle\sum_{X,Y \in object} X^2 Ydm$ ,

   $M(XY^2) = \displaystyle\sum_{X,Y \in object} XY^2 dm$ ,      $M(Y^3) = \displaystyle\sum_{X,Y \in object} Y^3 dm$ .

### 3.4.3. Moments computed on a boundary representation

This computation mode enables to speed up the attribute calculus when having at one's disposal a representation based on transition lists or on run-length coding.

From vectorized information, it must be come back to this kind of representation using vector digitalization.

If an object is described by a covering of horizontal line segments ((X1, Y), (X2, Y)), the moments get for values:

- $M(0) = \displaystyle\sum_{((X_1,Y),(X_2,Y)) \in object} dM(0)$ ;

- $M(X) = \displaystyle\sum_{((X_1,Y),(X_2,Y)) \in object} dM(X)$ ;

- $\ldots$

- $M(Y^3) = \displaystyle\sum_{((X_1,Y),(X_2,Y)) \in object} dM(Y^3)$ .

Where each increase $dM(X^i Y^j)$ can be expressed according to the following formulas:

- $dM(0) = \Delta X$

- $dM(X) = X_1 \cdot \Delta X + \displaystyle\sum_{k=1}^{\Delta X} k$

- $dM(Y) = Y \cdot \Delta X$

- $dM(X^2) = X_1^2 \cdot \Delta X + 2X_1 \cdot \displaystyle\sum_{k=1}^{\Delta X} k + \displaystyle\sum_{k=1}^{\Delta X} k^2$

- $dM(XY) = Y \cdot dM(X)$

- $dM(Y^2) = Y \cdot dM(Y)$

- $dM(X^3) = X_1^3 \cdot \Delta X + 3X_1^2 \cdot \displaystyle\sum_{k=1}^{\Delta X} k + 3X_1 \cdot \displaystyle\sum_{k=1}^{\Delta X} k^2 + \displaystyle\sum_{k=1}^{\Delta X} k^3$

- $dM(X^2 Y) = Y \cdot dM(X^2)$

– $dM(XY^2) = Y \cdot dM(XY)$

– $dM(Y^3) = Y \cdot dM(Y^2)$

and where

– $\Delta X = X_2 - X_1$ and $X_1 \leq X_2$

### 3.4.4. Moment translation to the object gravity center

The translation moment invariance can be deduced by correction of these ones according to the coordinates of the object gravity center.

$M(0)$ represents the surface area $S$ of the object.

The gravity center coordinates can be obtained with the help of moments of order 1 :

– the abscissa of gravity center $X_G = M(X)/S$ ;

– the ordinate of gravity center $Y_G = M(Y)/S$ .

The values of the higher order moments in the new reference frame $(\overrightarrow{X_G x}, \overrightarrow{Y_G y})$ (cf. figure underneath) become :

– $M(x^2) = M(X^2) - X_G^2 S$

– $M(xy) = M(XY) - X_G Y_G S$

– $M(y^2) = M(Y^2) - Y_G^2 S$

– $M(x^3) = M(X^3) - 3X_G^2 M(x^2) - Y_G^3 S$

– $M(x^2 y) = M(X^2 Y) - Y_G M(x^2) - 2X_G M(xy) - X_G^2 Y_G S$

– $M(xy^2) = M(XY^2) - X_G M(y^2) - 2Y_G M(xy) - X_G Y_G^2 S$

– $M(y^3) = M(Y^3) - 3X_G^2 M(y^2) - Y_G^3 S$

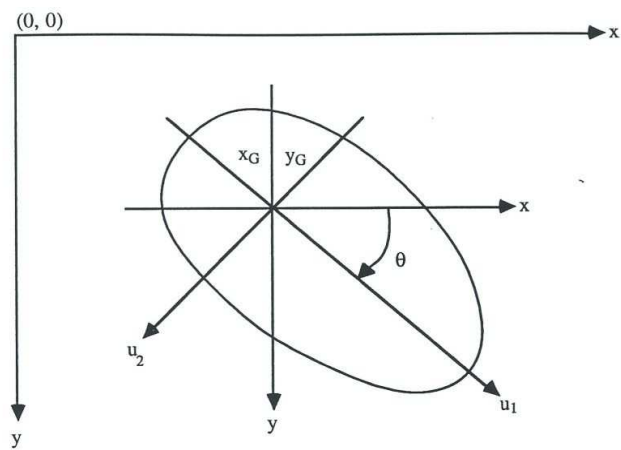

**Figure 17 : Image and object reference frames**

### 3.4.5. Moment rotation in the object reference frame

From the moments of order 2 can be deduced the inertia axes $u_1, u_2$ of the object:

$$M(u_1^2) = \frac{1}{2}\left( M(x^2) + M(y^2) + \sqrt{\left(M(x^2) - M(y^2)\right)^2 + 4M(xy)^2} \right)$$

$$M(u_2^2) = \frac{1}{2}\left( M(x^2) + M(y^2) - \sqrt{\left(M(x^2) - M(y^2)\right)^2 + 4M(xy)^2} \right)$$

This operation comes to assimilate the object to its inertia ellipsoid of major axis $\overrightarrow{u_1}$ and minor axis $\overrightarrow{u_2}$.

The rotation angle enabling to get in the object Eigen reference frame $(\overrightarrow{X_G u_1}, \overrightarrow{Y_G u_2})$, is worth :

$$\theta(\overrightarrow{X}, \overrightarrow{u_1}) = arctg\left( \frac{M(u_1^2) - M(x^2)}{M(xy)} \right) \text{ more or less } \pi.$$

Actually, it can be only deduced from the object inertia ellipsoid the orientation of the major axis and not its direction.

The crossed moment $M(u_1 u_2)$ cancels oneself and the moments of order 3 in the directions $u_1$ and $u_2$ become:

$$M(u_1^3) = \cos^3\theta \cdot M(x^3) + 3 \cdot \sin\theta \cdot \cos^2\theta \cdot M(x^2 y)$$
$$+ 3 \cdot \sin^2\theta \cdot \cos\theta \cdot M(xy^2) + \sin^3\theta \cdot M(y^3)$$

$$M(u_1^2 u_2) = -\sin\theta \cdot \cos^2\theta \cdot M(x^3) + \cos^3\theta \cdot M(x^2 y)$$
$$- 2 \cdot \sin^2\theta \cdot \cos\theta \cdot M(x^2 y) + 2 \cdot \sin\theta \cdot \cos^2\theta \cdot M(xy^2)$$
$$- \sin^3\theta \cdot M(xy^2) + \sin^2\theta \cdot \cos\theta \cdot M(y^3)$$

$$M(u_1 u_2^2) = \sin^2\theta \cdot \cos\theta \cdot M(x^3) - 2 \cdot \sin\theta \cdot \cos^2\theta \cdot M(x^2 y)$$
$$+ \sin^3\theta \cdot M(x^2 y) + \cos^3\theta \cdot M(xy^2)$$
$$- 2 \cdot \sin\theta \cdot \cos^2\theta \cdot M(xy^2) + \sin\theta \cdot \cos^2\theta \cdot M(y^3)$$

$$M(u_2^3) = -\sin^3\theta \cdot M(x^3) + 3 \cdot \sin^2\theta \cdot \cos\theta \cdot M(x^2 y)$$
$$- 3 \cdot \sin\theta \cdot \cos^2\theta \cdot M(xy^2) + \cos^3\theta \cdot M(y^3)$$

The direction of axes $\overrightarrow{u_1}$ et $\overrightarrow{u_2}$ is set by making the moment $M(u_1^3)$ to have a positive value.

This is equivalent to give as direction for $\overrightarrow{u_1}$ the one providing the strongest eccentricity according to this axis.

The eccentricity according to the object major axis is a characteristic proper to its shape and its value is stable when are known the inertia axis directions.

The uncertainty more or less 360° is then solved in the following way:

$$M(u_1^3) < 0 \Rightarrow \theta = \theta + \pi$$

And the moments of order 3 are corrected by a sign change:

$$M(u_1^3) = -M(u_1^3)$$
$$M(u_1^2 u_2) = -M(u_1^2 u_2)$$
$$M(u_1 u_2^2) = -M(u_1 u_2^2)$$
$$M(u_2^3) = -M(u_2^3)$$

These values represent the different asymmetries of the object in its Eigen reference frame.

### 3.4.6. Spatial characteristics deduced from the moments

We have just described how getting:

- — the coordinates of the gravity center $X_G, Y_G$ of the object;
- — the angle $\theta$ made by the object in the image reference frame.

The other moments computed in the object reference frame $(X_G, Y_G, \theta)$ enable to have at one's disposal object spatial characteristics invariant in translation and rotation in the view take plane.

These characteristics are:

- — $M(0)$ the object surface area;
- — $M(u_1^2), M(u_2^2)$ the object inertias;
- — $M(u_1^3), M(u_1^2 u_2), M(u_1 u_2^2), M(u_2^3)$ the object asymmetries.

These characteristics can be used for performing statistical pattern recognition.

### 3.4.7. Complementary attributes

Other attributes can be computed over the object supports, as well as their radiometric responses.

On the support, it can be got:

- — the coordinates of the bounding box;

- the object compactness that is measured by the ratio of the perimeter square over the surface area ;
- the object eccentricity which is delivered by the ratio of the major axis on the minor axis.

The following characteristics can also be evaluated according to the object radiometric response:

- value interval (minimum and maximum) ;
- average value ;
- dispersion ;
- and asymmetry.

They are corresponding to the moments of the luminescence distribution over the object surface.

## 3.5. Geometric transformations

### 3.5.1. Problem position

The geometric transformations take place in two distinct situations in the present research:

- on one hand in order to replace in a universal reference frame objects observed by a satellite ;
- on the other hand, in order to geographically browse and to render referred objects according to a given viewpoint.

The information captured on the Earth surface can be mapped on a plane with the help of a planispheric projection, which enables to link a space point standing on a sphere with a cartographically referred point

The universal used reference frame is a polar coordinate system made from angle measures applied from the sphere center started from the equatorial plane and from the meridian plane crossing Greenwich in United Kingdom: these angles are the latitude and the longitude of a geographical point.

Let us notice that a planispheric projection is not a linear transformation, which can be at the origin of some geometric distortion. Besides that, the spherical model is only providing an approximated shape for modeling the Earth geoid. One can then be led to take in account the geoid flattening at the poles. Actually, any processing will depend on the location precision that is expected to get at the end of the processing and on the conditions with which the view takes will have been caught.

In the present case which is concerning the building of catalog, it will be possible to restrict the processes to the first approximation orders. It would not be the case, if we should be able to localize objects with a very high geographical precision.

### 3.5.2. Nature of used sensors

Our interest is focused on earth resource observation satellites.

The satellite view takes are mainly conditioned by:

- the elevation from which they are observing the Earth surface;
- the viewing angle or the observed field of the data capture system ;
- the resolution of the same capture system.

The table shown underneath lists the values of these parameters for the main civilian earth resource observation satellites.

Among these ones, it will be found:

- a meteorological geostationary satellite able to capture a whole hemisphere within a single observation ;
- a second meteorological satellite able to scan a continental area ;
- two solar-synchronous earth resource observation satellites that can capture data over regional areas.

| Satellite | Elevation | Observed area | Resolution |
|-----------|-----------|---------------|------------|
| METEOSAT | 36 000 km | hemisphere | 2,5-5 km |
| NOAA | 850 km | 3 000 x 3 000 km | 1-17 km |
| LANDSAT | 700-900 km | 185 x 185 km | 30-80 m |
| SPOT | 800 km | 60 x 60 km | 10-20 m |

**Table 2 : Characteristics of the main earth resource observation satellites**

Concerning continental size areas, it seems hard to avoid the application of a geometrical correction in order to correct the distortions due to the Earth roundness. At the opposite, this one has less weight on the regional view takes, because at this scale it is nearly about one hundred meters for one hundred kilometers. The influence of relief prevails then on the Earth roundness in mountain areas for such view takes. It is why data satellite providers may propose two distinct levels of geometrical corrections, for which one is taken in account the relief of the observed area.

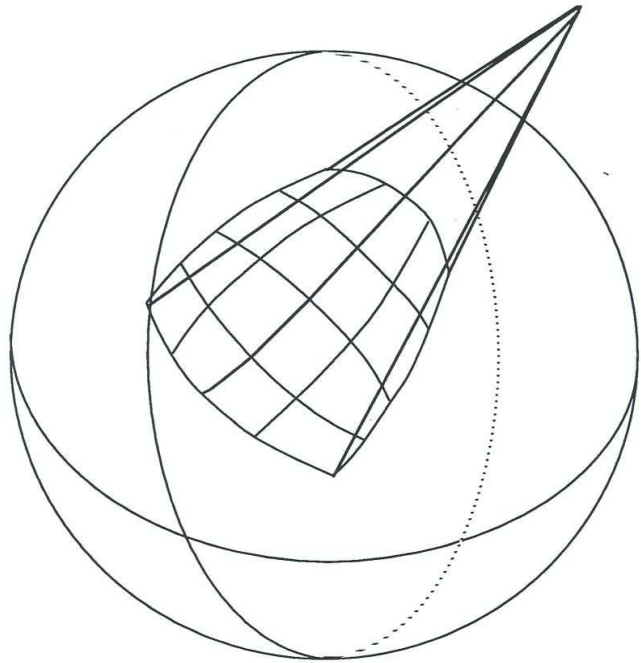

**Figure 18 : Continental area observed by a geostationary satellite**

The basic geometric corrections deal mainly with view taking conditions that are:

- — the path followed by the satellite when it is moving (solar synchronous satellite case) ;
- — the geometric distortion generated by the data capture system.

Especially in this last case, the NOAA and LANDSAT observation satellites are using a mirror-based scanning system which is producing a pillow shaped distortion orthogonally to its rotation axis. The method used for re-sampling images in their corrected observation reference frame may change notably image content

### 3.5.3. Information available for defining a planimetry transformation

Two distinct situations enable to perform a data geometric correction procedure:

- — either, information about the satellite position and attitude are a priori known as well as the physical characteristics of the view take system at scene capture time;
- — or, it is already precisely known the geographical location of the information pieces in the view take.

The first situation corresponds to the one met by the data provider: it is then necessary ask him data with the expected level of correction.

The second situation corresponds to the one of an user having to his disposal capabilities in photo-interpretation and geographically referred data: this one will be able to geometrically rectify these information pieces according to the data at his disposal.

A third situation occurs when it cannot be held one or the other information series, it is then necessary to only rely on the information belonging to the header of the images delivered by the data providers. In this last case, it can usually be retrieved the geographic location of the corners and the center of the captured image, but only linear or bilinear transformations can be applied on this data. This transformation only enables to recalibrate images in position and orientation into a cartographic reference frame. The error due to the Earth roundness will not be taken in account on regional view takes and the data will keep a certain uncertainty in location.

### 3.5.4. Method usually applied for specifying a geometric transformation

This method is applicable in the case, when observable and geometrically referred information is available, as well as in the case, when we can only use information belonging to image headers. In the first case, the transformation order could be higher than in the second one and the applied processes will be much more precise.

This transformation must enable to move from a planar mesh indexed $(p, q)$ to a new planar mesh indexed $(r, s)$. The two meshes being regularly sampled, it is consisting then to determine the relation:

$$r = f(p, q);$$
$$s = g(p, q),$$

which is checked by a set of points which coordinates relatively to the two meshes are already known:

$$\left\{ \left( (p_m, q_m), (r_m, s_m) \right), \quad 1 \le m \le M \right\}$$

For a transformation approximated by polynomials of order k, it is finding the series expansions:

$$\widehat{f}(p, q) = \sum_{0 \le i+j \le k} a_{ij} \, p^i q^j$$

$$\widehat{g}(p, q) = \sum_{0 \le i+j \le k} b_{ij} \, p^i q^j$$

so as the mean square error

$$\sum_{m=1}^{M} \left( r_m - \widehat{f}(p_m, q_m) \right)^2 + \left( s_m - \widehat{g}(p_m, q_m) \right)^2$$

can be minimum.

If      $a = [a_{00}, a_{10}, a_{01}, \ldots, a_{0k}]^T$,

$b = [b_{00}, b_{10}, b_{01}, \ldots, b_{0k}]^T$,

$r = [r_1, r_2, \ldots, r_M]^T$,

$s = [s_1, s_2, \ldots, s_M]^T$,

$$A = \begin{bmatrix} 1 & p_1 & q_1 & \cdots & p_1 q_1^{k-1} & q_1^k \\ 1 & p_2 & q_2 & \cdots & p_2 q_2^{k-1} & q_2^k \\ \vdots & \vdots & \vdots & \cdots & \vdots & \vdots \\ 1 & p_M & q_M & \cdots & p_M q_M^{k-1} & q_M^k \end{bmatrix}$$

then:

$$a = A^+ r \qquad \text{and} \qquad b = A^+ s ,$$

with:

$A^+ = \left(A^T A\right)^{-1} A$, which is the pseudo-inverse matrix of $A$.

### 3.5.5. Method used for applying a geometric transformation parametrically defined

If it is applied the parametric transformation

$$(p, q) \rightarrow \left(\bar{f}(p, q), \bar{g}(p, q)\right)$$

Onto the image mesh, it is usually resulted that the transformed vertices are not able to insure and provide a data regular covering in the transformed space. In order to re-sample the transformed data over the whole mesh, it remains possible to assign values to mesh elements:

   –   by assignment the value of the nearest neighbor ($C_0$-continuity extension) ;
   –   by polynomial interpolation using data standing in a given neighborhood.

This second method gets the disadvantage to smooth data and consequently to delete part of the image information. This one becomes harder to analyze. The nearest neighbor assignments are preferable in this situation.

### 3.5.6 Implementation of a geometric transformation

The transformation is specified under the shape of a set of point couples whose coordinates are given in the departure and the arrival reference frames.

The control points of such a transformation are graphically shown underneath in their transformed reference frame. This point seeding enables to specify a transformation up to the order 3. In the following pages, it will be so found :

- a system of nested centers displayed in their original reference frame ;
- the system transformed using an approximation computed at order 1, then at order 2 and at last at order 3.

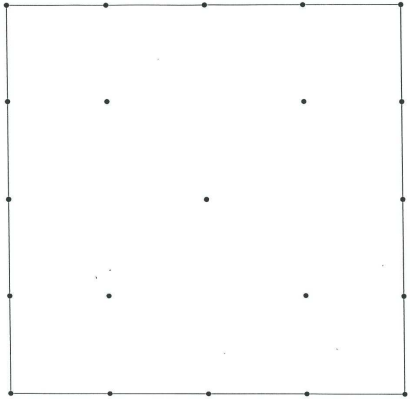

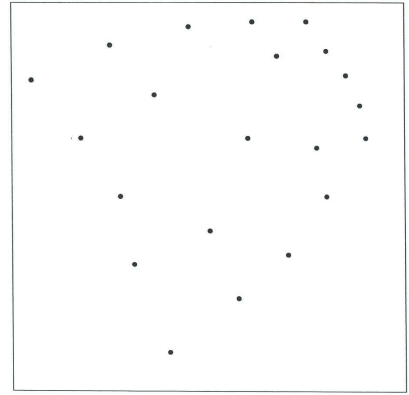

a) Departure reference frame           b) Arrival reference frame

**Figure 19 : Seeding of points defining a geometric transformation**

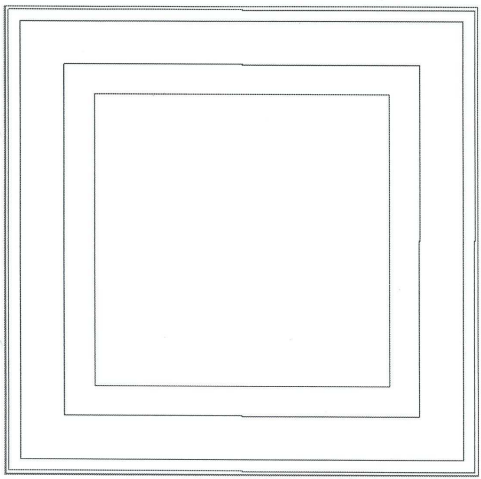

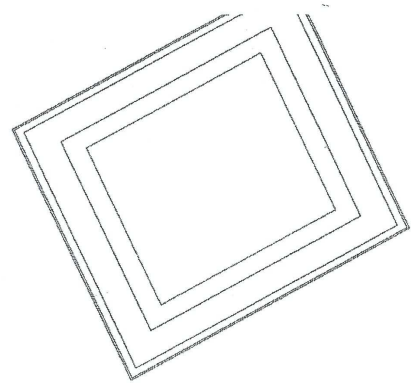

a) Initial shape           b) Transformed at order 1

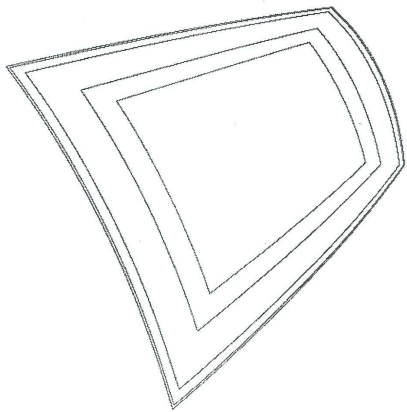 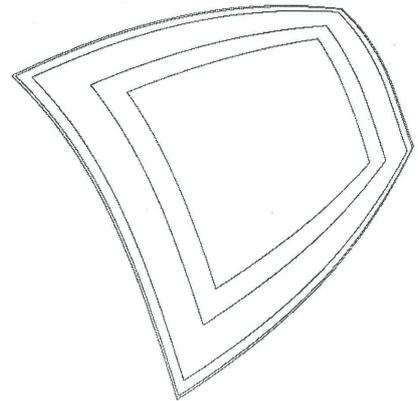

c) Transformed at order 2                    d) Transformed at order 3

**Figure 20 : Geometric transformed of a shape at different orders**

### 3.5.7 Spatial observation and cartographic projections

The transfer of observations made from space about Earth and the rendering of data referred in a cartographic reference frame imply the use of similar but distinct geometric transformations.

If objects are observed from space from some observation point, two different cases must be distinguished:

- the observation point stands at a finite distance, a conical perspective is applied on objects and distorts their shapes according to this distance ;
- the observation point is at the infinite (or nearly), it is a cylindrical perspective that is not distorting the shape of objects.

In the first case the transformation is a homography, in the second one it is an orthography. In the two situations, only the parts visible from the observation center can be perceived.

By projection, these pieces of information are brought back in a plane: cameras are performing this operation on a photo-sensitive planar surface.

If it is expected to get information about the Earth surface registered in a planar model, it is necessary to make a planispheric transformation. Information about Earth surface is initially defined in a polar coordinate reference system, relatively to the Earth center. This planispheric projection enables to transfer information over a plane surface.

In this kind of transformation, it is tried to preserve some geometric characteristics like:

---

- — the distances between different points (equidistant representation) ;
- — the angles between line segments (conformal representation) ;
- — the surface areas (equivalent or equi-areal representation).

It does not exist any planispheric projection that remains unchanged all these characteristics. This observation leads the cartographer to propose several different projection systems enabling to preserve one or the other one of these properties:

- — the first one of these properties enables to compute the shorter path joining two airports in air navigation ;
- — the second one enables to establish routes in sea navigation ;
- — the last one enables a geometer to measure the areas of terrain plots.

A projection needs the specification of a projection center and a surface shape tangential to Earth in a given place or according to a given curve on which information will be projected. This surface shape is then spread on plane in order to build up a map. The used tangential surfaces are either planes or cylinders, or also cones. The projection centers are either the earth center or one of the Earth poles, or also some point in the Space.

In navigation, the more used projections are:

- — the cylindrical projection;
- — the orthographic projection;
- — the gnomonic projection (conical perspective centered at the Earth center) ;
- — the Mercator projection which is a variant of a cylindrical projection supported by the equator ;
- — the transversal Mercator projection which is again a cylindrical projection, but supported by a series of meridians ;
- — the Lambert projection, used for instance in France for building maps and that provides a conformal transformation.

In order to implement these projections, it has been used the free software GCTP (General Cartographic Transformation Package), developed by the National Mapping Division of US Geological Survey and that is proposing about thirty standard transformations.

### 3.5.8 Image transfer into a universal reference frame

The used universal coordinate system is the geographic coordinate system, made from the latitude and the longitude of a point on the Earth surface.

Concerning satellites with a wide observation field, it will be necessary to identify landmarks in the image and to compute, starting from these points, a transformation with an enough high order, in order to map the image into a universal reference frame.

Concerning satellites with a narrower observation field, it can be accepted a linear transformation for building an image catalog relying on the information found in the image headers.

The LANDSAT images are referenced in an own coordinate system based on the indices of tracks and lines, from which can be exactly computed the image center and corners in a universal reference frame. This function is performed with a free software named LSF (LANDSAT Scene Finder), developed by Computer Sciences Corporation on behalf of the EOSC (Earth Observation Satellite Company) and available on Internet.

The same coordinates are directly available in the header of SPOT images. Unlike LANDSAT whose view takes are made without incidence (orthogonally to the Earth), the SPOT satellite can make view takes with incidence (by lateral sight). In this last case, the position error involved in a linear transformation may increase.

### 3.5.9 Transformations used in data base browsing

For implementing a browsing system in a base of images captured by Earth observation satellites, the orthographic projection seems at usage to be the most convenient for performing this task. It enables to simulate what could be seen by a satellite at a location, with an elevation and with a given observation field. According to these two last parameters, it must be kept in mind, as previously, that the used transformation order may reduce the impact of possible distortions. Especially concerning continental observations, it will be needed to use high orders.

## 3.6. Cataloging data from satellite observation

### 3.6.1. Data registration

The information produced by image analysis is stored in the archiving base according the following manner:

- a "header" file records main information data about the archiving base ;
- an "objects" file records the image number and all the attributes that have been computed on the objects belonging to the archiving base ;
- a "vertices" file records the lists of vertices stemming from the boundary vectorization of all the objects belonging to the archiving base;
- an "image" file can be used for storing all the images that are taking part to the constitution of the archiving base.

The use of this last file remains optional in order to master the data size of the resulting archive.

In the "header" file is specifically stored the following information:

- the number of objects belonging to the archiving base ;
- the number of vertices ;
- the number of images from which they have been extracted.

When this information is archived, the geometric transformation, that enables to come into a universal reference frame, is specified.

The chosen reference frame is normalized between -1 and +1 and can be mapped over a geographical reference frame. The transformation applies on the object vertices and does not concern the attributes as long as it is only used affine transformations. The attribute values are invariant with these transformations (nearly from a homothetic factor for some ones). At registration time, it can be indicated among all the attributes that can be computed, which ones should be stored in the archiving base.

It can be found underneath the list of archiving functions that have been developed.

| Name | Function |
|------|----------|
| arcinit | creation of a new archive |
| arcwrhdr | modification of the archive header |
| arcrdhdr | reading of the archive header |
| arcaddob | initialization of the addition of an object |
| arcwrobj | addition of information about an object |
| arcwobja | addition of object attributes |
| arcwobjv | addition of object vertices |
| arcwrvtx | addition of a vertex belonging to an object |
| arcxtrob | initialization of the extraction of an object |
| arcrdobj | reading information about an object |
| arcrdobja | reading object attributes |
| arcrobv | reading object vertices |
| arcdvtx | reading a vertex belonging to an object |

**Table 3 : List of archiving functions**

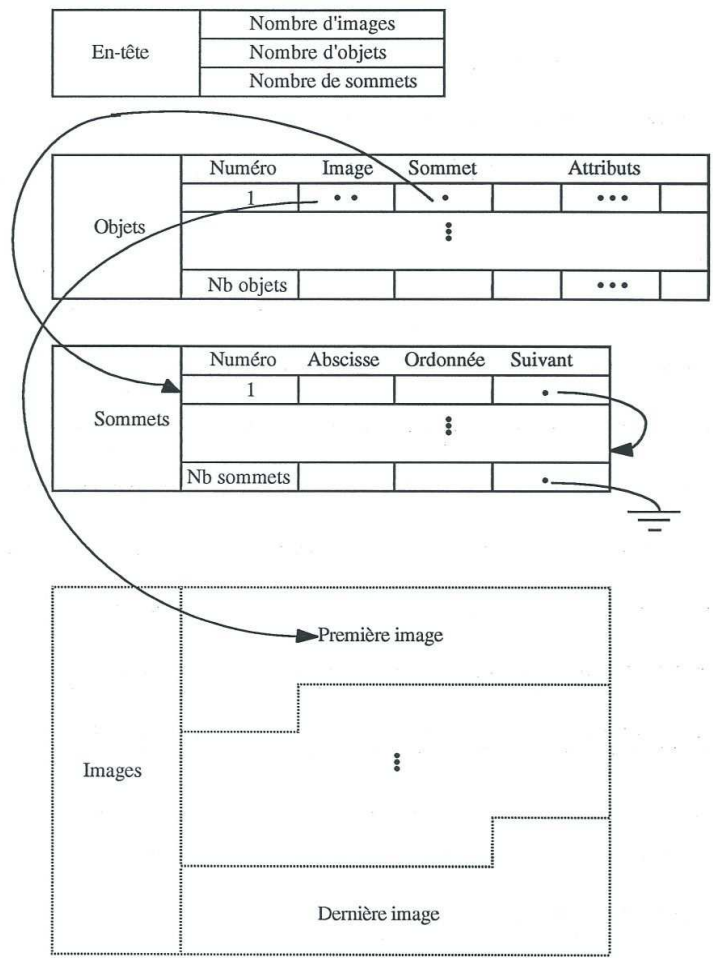

**Figure 21 : Data recording structure**

### 3.6.2. Data indexing

The data indexing is directly applied on all the information belonging to the archiving base. The attributes linked to each data base object are serving as key for referring to them.

A tree of the dimension of attribute vectors is created and for each object from the archiving base is recorded at the location defined by its attribute vector in the so generated space.

So, if it is chosen as attributes:

- the coordinates of the gravity center,
- the mean radiometric response,
- the object surface area and compactness.

It will be possible to build an indexing system enable to refer objects in position, in luminosity and in shape.

Some other attributes may enrich this first selection in order to have at one's disposal a richer indexing system. So, beginning with a same archiving base, more or less rich indexes can be generated.

The precision until which the indexing tree must be developed is given when the index is built.

The object numbers are recorded in lists hanged at each terminal node of the indexing tree. The size of these lists varies according to the building precision of the index

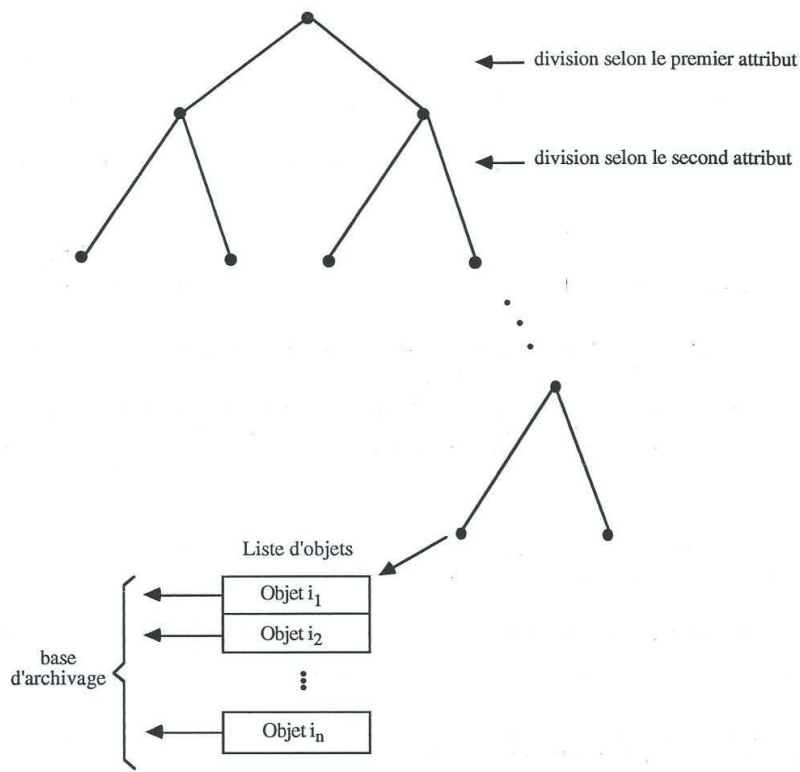

**Figure 22 : Object indexing tree**

The object indexing is handled using a $2^k$-tree with the help of the KDTREE software.

If spatially objects are more or less uniformly spread above the emerging inland, it is not the case when it is looked to their attributes in shape and in luminous response. Unbalanced trees are then generated when it is tried to compute an index. On parallel computer, it is at the origin of a load unbalancing when running parallel processes. For avoiding this trouble, it has been proposed to perform a histogram equalization for each attribute before starting to generate an index.

Currently the lower and higher bounds, between which attributes are sampled, are given when the archiving base is built.

### 3.6.3. Content-based querying

*3.6.3.1. Presentation*

The data recording using an archiving base is prepared by applying image analysis functions. These same functions enable to retrieve the recorded data from an archiving base.

The indexing operations are made with the help of the KDTREE modeling software. It is so possible to read an archiving base and to build its index, but also to provide a new archiving base that will be linked to a sub-index extracted from the initial index.

The archiving bases represent the medium through which image analysis functions and the KDTREE software are able to communicate for data cataloging. In order to reach this aim, some complementary function have been developed for performing multi-criteria interrogation, histogram computing and sub-index extracting so as to retrieve information about the objects of interest from the data base (cf. Table underneath).

*3.6.3.2. Multi-criteria selection*

It is a first use case: from an archiving base it is expected to extract an information sub-set inside the Cartesian product of several variation intervals. These intervals apply on attributes on which is relying the index enabling to access to the data base. They are defining in the attribute space a parallelotope whose faces are parallel to those from the space reference frame.

The computing of histogram according to any dimension of an index enables to provide global information about the distribution of an object population according each index attribute.

| Name | Function |
| --- | --- |
| kdrcix | construction of the index of an archive |
| kdadix | addition of an object into an indexing tree |
| kdhsix | histogram computation of an index |
| kdhspy | index histogram according to a given dimension |
| kdslix | tree building of a query defined by a multidimensional interval system |
| kdxtix | extraction of the index corresponding to a selection |
| kdanix | reconstruction of an archive from an index |

**Table 4 : List of indexing and selection functions**

### 3.6.3.3. Nearest-neighbor query

It is the second usage case: from an archiving base, it is expected to retrieve the information belonging to the neighborhood of a point or a given element. The point coordinates, which are the values of its attributes for the indexing system, and the neighborhood size are known at query time

The tree-like indexing system handled under KDTREE induces a topological structure of Borel algebra over the registered data. This topology is less fine than the metric $d_k$-topologies more often used on discretized spaces, but it relies on a ultra-metric distance that has the advantage of enabling to compare two sub-sets between them.

Within this topology, at each point of the discretized space is associated a neighborhood system which is the series of the nested parts linked to the tree traversal that enables to retrieve the tree root from the queried point: according to the depth reached in the branch traversal, the reached neighborhood will be more or less small in the modeling space. So the tree of the point developed down to the neighborhood size will enable to model an index used for performing a nearest-neighbor query.

| Name | Function |
|---|---|
| kdirvc | creation of a vector |
| kdcrbt | creation of a tree |
| kdarvt | addition of a vector to a tree |
| kdplvi | projection of a tree into a sub-space |

**Table 5 : Auxiliary functions used for querying**

### 3.6.3.4. Constructive approach and similarity index

Nearest-neighbor query and multi-criteria selection enable to build primitive questions that can be answered with such an indexing system.

The KDTREE software provides Boolean operators that enable to combine several trees in order to set new ones the result can be interpreted as set operations applied on subsets of the modeled space (Cf. Table underneath).

In the attribute space to which indexes are belonging, it is then possible to structure primitive queries into more complex ones.

The Boolean algebra can also put in order the answers from a nearest-neighbor query according to the similarity index associated to the ultra-metric distance induced by the indexing hierarchical structure:

- the objects extracted at the precision of index building are similar to the one from which is built the query ;
- those, that are belonging to the neighborhood of just higher size, have a dissimilarity index equal to the neighborhood size ;
- and so on in an increasing way up to the index root where the index will have a maximal value and will measure the volume size of this space.

By normalizing by the space volume size, it is got by difference a similarity index.

By putting in order the answers according to the nested parts, answers will be sorted in decreasing order according to this similarity index.

In order to keep only the contributions owned by each part, it will be got the right query by erasing from the present querying tree, the contribution corresponding to the previous query. So by building trees at variable precision and using set difference, it can be put in order according a similarity index the nearest-neighbor answers inside a neighborhood of any size.

| Name | Function |
|------|----------|
|      |          |

| kdass | assertion of a tree |
|-------|---------------------|
| kdnot | negation of a tree |
| kdunio | union of two trees |
| kdintr | intersection of two trees |
| kdexcl | exclusion of two trees |
| kddiff | difference of a tree by another one |

**Table 6 : Boolean algebra functions**

### 3.6.3.5. Sorting an archiving base

The modeling of spaces of dimension $k$ regularly decomposed by trees of order $2^k$ mapped into binary trees shows that this representation model has got the continuity power, that is to say that it can be found a continuous transformation that maps sub-sets of $R^k$ straight into $R$ .

Consequently it should be possible to find paths enabling to visit once and only once by continuous moves the whole set of a data base: in the case of data base managed with $2^k$ -trees, by nearest-neighbor moving according to the Hausdorff distance.

The Hilbert scans satisfy for instance to this usage as it is shown in the following figures : the target is to find a visiting path for places defined by the geographic coordinates that minimizes the elementary moves made from one to another one.

Concerning a multidimensional data base, it is equivalent to find a visiting path maximizing step by step the similarity index of registered data, which is to sort base data according to its similarity.

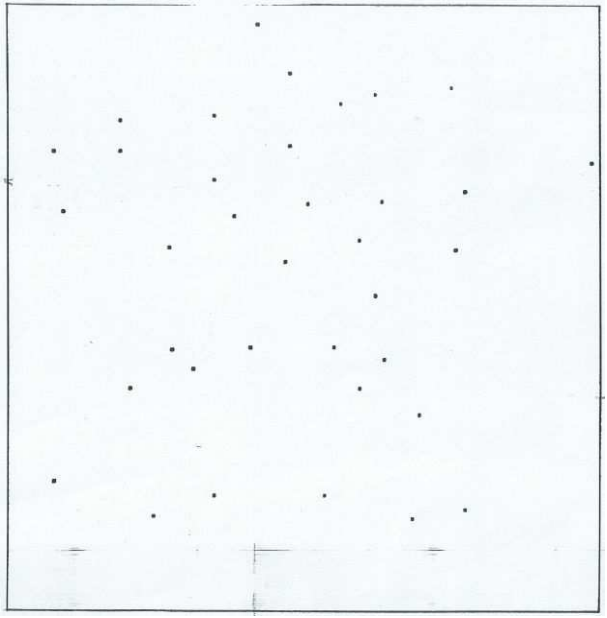

**Figure 23 : Map of the main France towns referenced in polar coordinates**

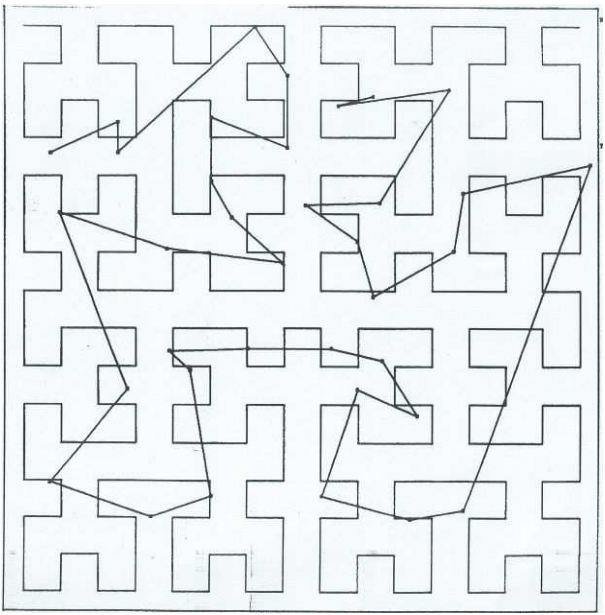

**Figure 24 : Looking for a visiting path using a Hilbert scanning**

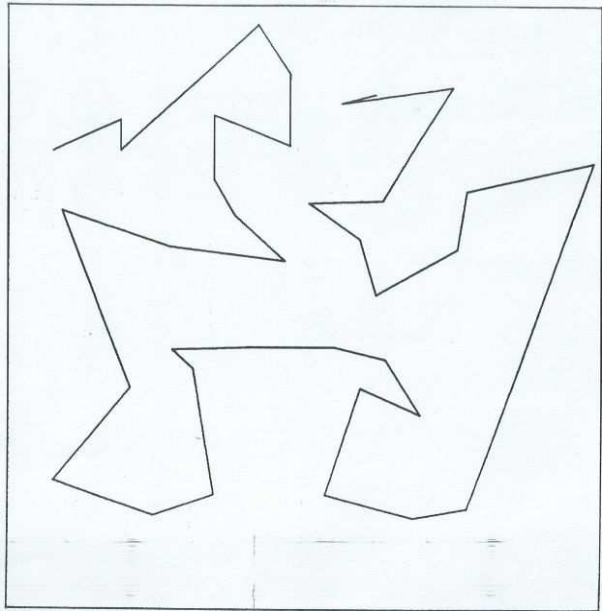

**Figure 25 : Nearest-neighbor visiting path of the main France towns**

## *3.7. Obtained results*

### 3.7.1. Processing of a multispectral image

We have tried to process a multispectral image using the different approaches that have previously been presented.

The image on which is relying this analysis is a SPOT 1 A view take made over the Landes region in France, located just at the North of Bayonne city.

The three bands of this multispectral image are successively shown in the following pages. They are nearly covering the B, G, R bands in the visible field, including a sensible shift towards high frequencies in comparison with a usual TV picture.

On this scene, the two first bands present a lower intensity than the third one and show significantly correlated information. On the third band, clearly appear the Adour River and the Biarritz airport.

The image is made from a 1024 x 1024 window with side of 20 km extracted from the original image.

It can be noticed that averaging grey levels when displaying is masking the information richness included in the high number of components: especially the Bordeaux-Bayonne highway appears nearly clearly on the labeled image on the border of the Landes forest.

A sensor correction default appears in the image middle on the second band.

The third band seems actually richer in luminosity variations than the two previous ones.

It has been processed to the image classification in its radiometric space made from the set of the three bands.

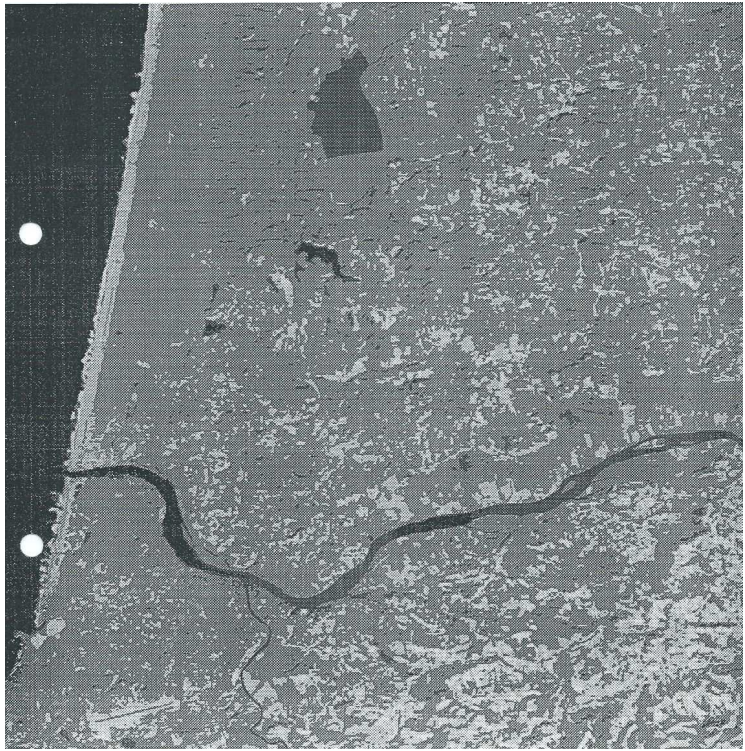

**Figure 26 : Result of the segmentation of the spectral band triplet**

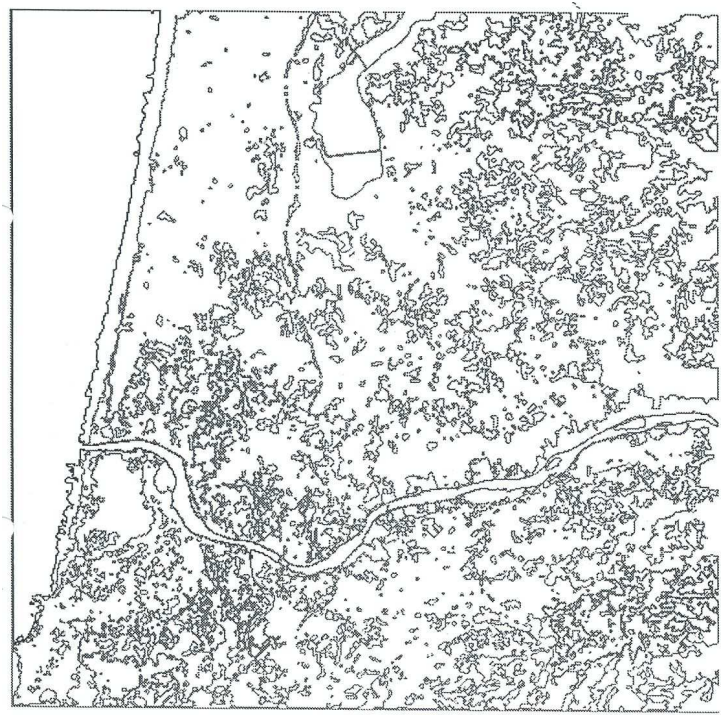

**Figure 27 : Contours of the selected regions**

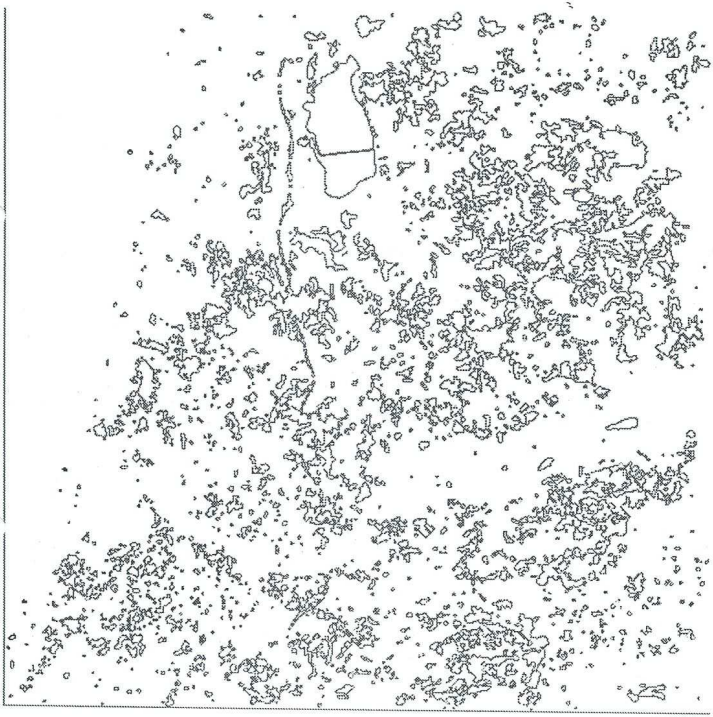

**Figure 28 : Internal contours to the selected regions**

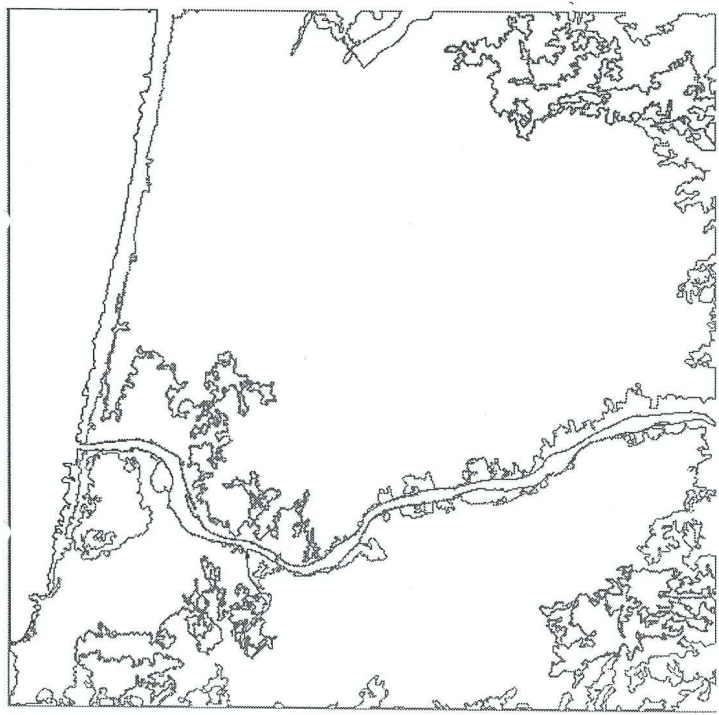

**Figure 29 : External contours of the selected regions**

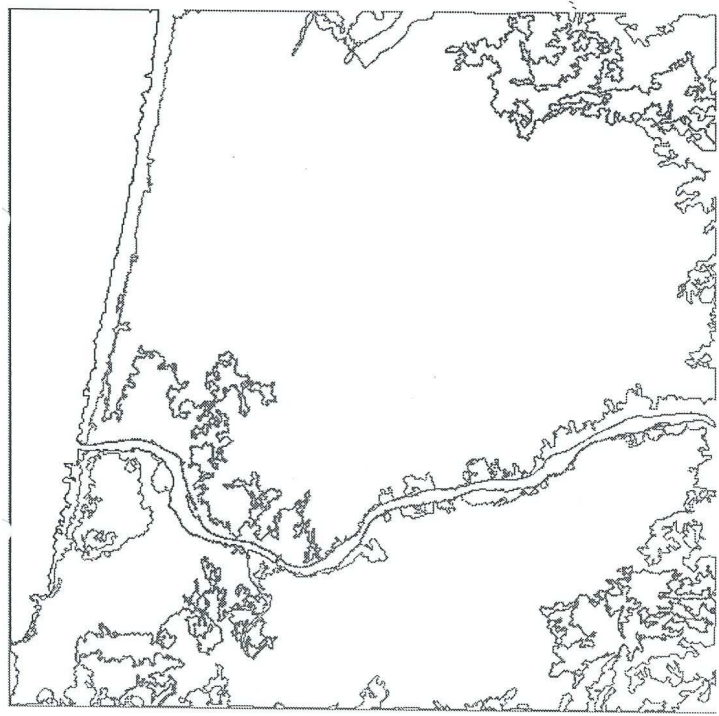

**Figure 30 : Vectorization of the external contours**

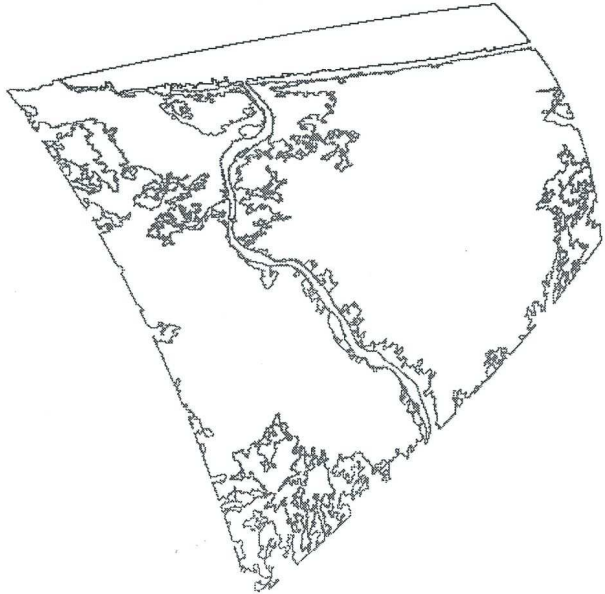

**Figure 31 : Quadric transformed image of the vectorized regions**

### 3.7.2. Processing of a mono-spectral image

We have tried to apply the same process to a mono-spectral image, except the multispectral classification stage.

It is concerning a SPOT 1B panchromatic image corresponding to a view take made over Syria. As previously it has been chosen a 1024 x 1024 window holding an observation of the Ghab plain over a surface about 10 km of side.

This plain is composed of an agriculture area standing on the right side of the image and a mountain relief on the left two third of the image. At the beginning of this relief, it can be seen a set of regular buildings from which is starting a climbing lane.

In the following pages are shown the series of processing operations made from the segmentation of the image couple composed by the initial image and its differentiated version at order 1.

It can be noticed that the contrast improvement applied on the initial image does not increase the information volume and that this one is mainly standing in the differentiation at order 1 rather than order 2.

Usually, the flatness of response surfaces in an image concentrates the image information in the early first differentiation orders and could lead to an identity between image discontinuities and high image gradient variations.

We have selected the result of the segmentation of the couple of the original image with its gradient image.

After displaying the original image, it can be then seen:

- the image of the connected components assigned with their average grey level ;
- and the image of the labels of the same components.

The components attributes are computed and a sub-set of them is selected.

Concerning the selected sub-set, it will be found:

- the image of the components colored with their labels ;
- the component contours ;
- the external contours.

The structures selected using their surface area and their compactness are:

- in a first case, the most important structures in surface area ;
- in a second case, the high size geometric structures similar to linear structures.

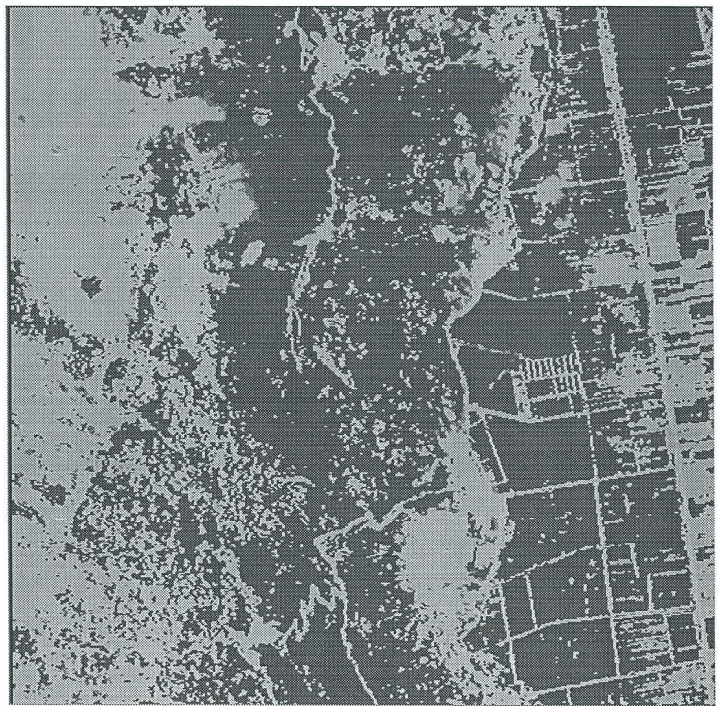

**Figure 32 : Image of the average grey levels of the connected components**

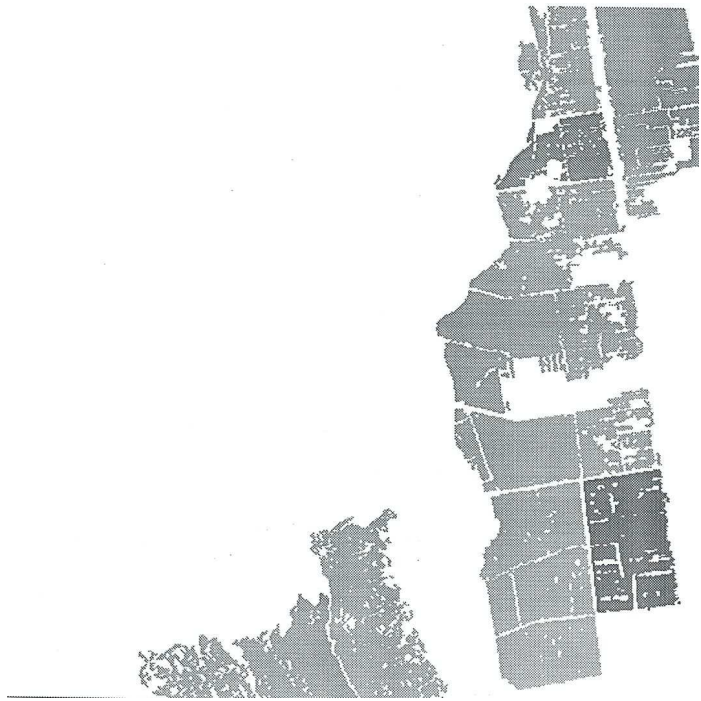

**Figure 33 : Regions selected on their surface area**

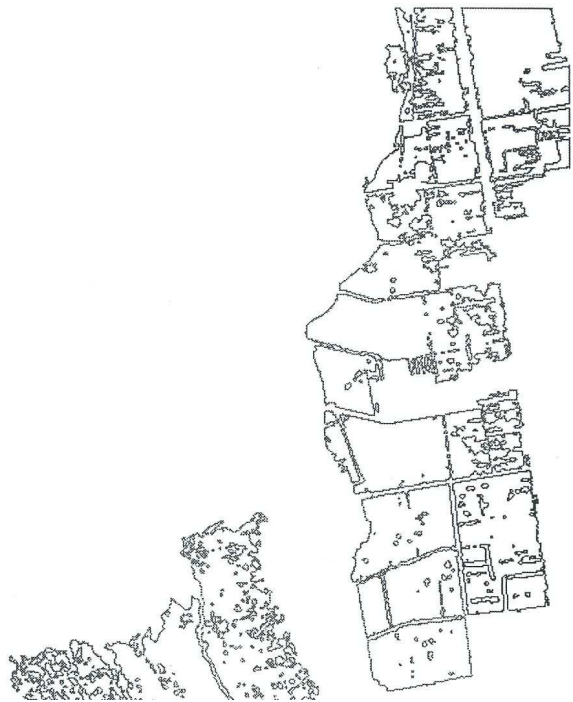

**Figure 34 : Contours of selected regions**

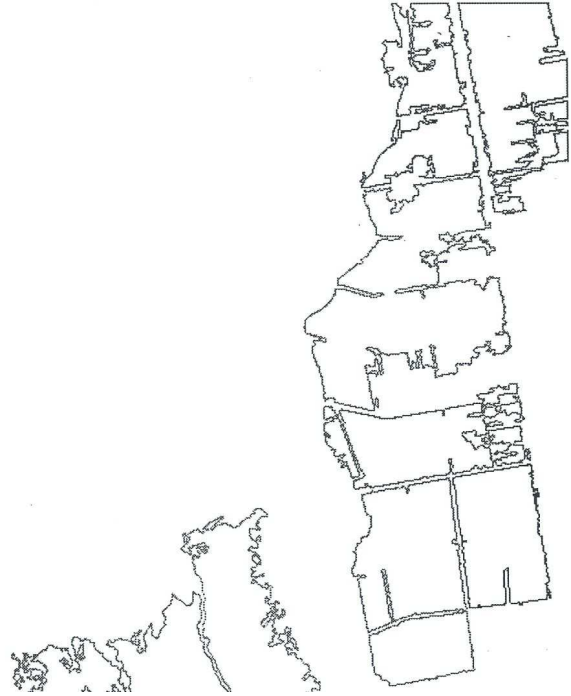

**Figure 35 : Region external contours**

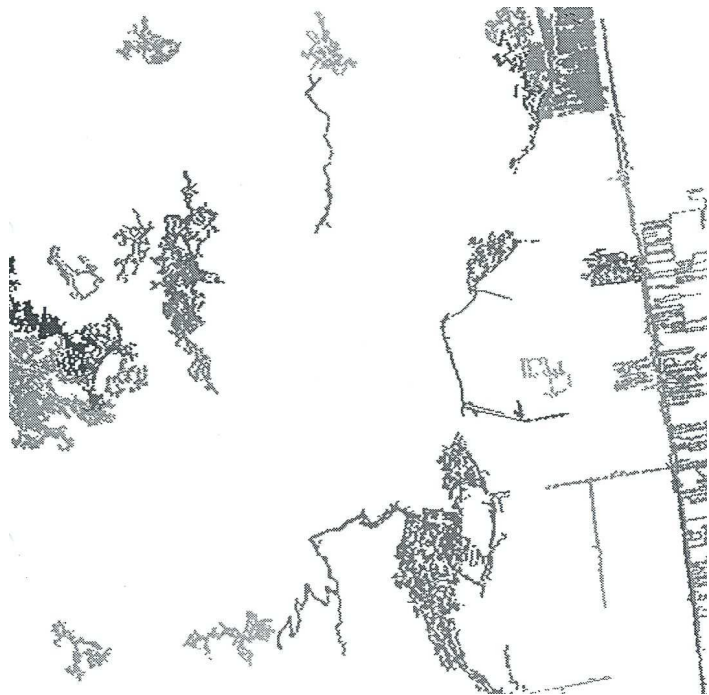

**Figure 36 : Regions selected on their compactness**

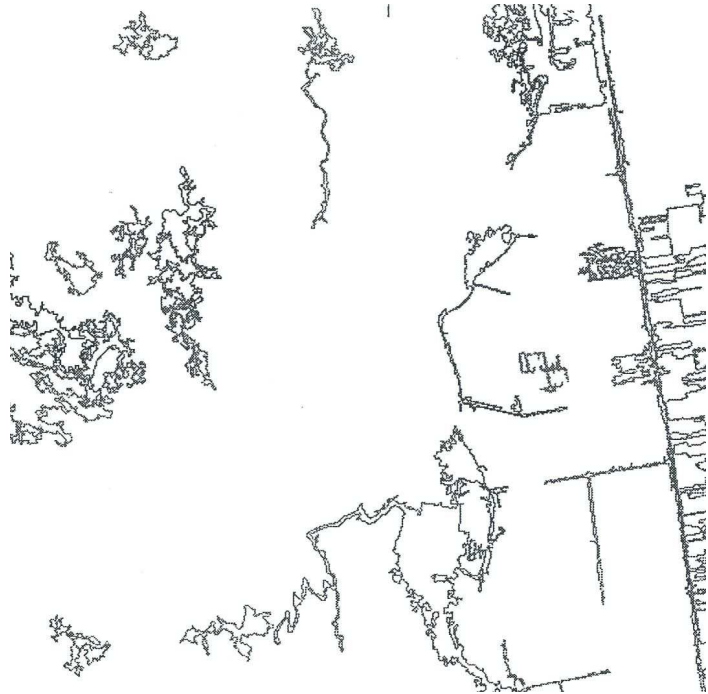

**Figure 37 : External contours of the selected regions**

## 3.8. Results achieved during the research

<u>Selection of a data set</u>

For the sake of the research, it has been possible to access to a set of images coming from different sources (SPOT, LANDSAT, ERS), and also to the reading procedures concerning SPOT images.

The LANDSAT image can easily be read without any specific procedure and the ERS image was delivered in an intermediate format directly readable.

We also got signaling information on a printed state at a first step, then by editing the image header with the help of the reading procedure for SPOT images.

Concerning LANDSAT images, Computer Sciences Corporation put in public domain on Internet a software named LANDSAT Scene Finder (LSF) that allows to retrieve the geographical coordinates of a LANDSAT scene.

In the two cases, the scene localization is given by the four image corner and optionally by the center, in geographic coordinates.

Without having at our disposal a landmark list, only linear geometric transformations can be applied on.

**Thematic conversion**

The image decomposition scheme initially proposed was well fitted to multispectral image analysis, but much less to mono-spectral images.

By relying on the works made in automatic classification by J-G. POSTAIRE at U.S.T.L. (Université des Sciences et Techniques de Lille), we have developed a technique for image segmentation into regular components (in the meaning of surface approximation). Its principle is the following one:

— if necessary, the picture is smoothed, then differentiated several times until reaching the desired order enabling to detect the image singularities, by preserving aside intermediate images (in practice, it used to stop at a not too high order) ;
— similarly to a multispectral classification, the multidimensional histogram made from the luminance vectors, taken from each processed images by forgetting the point positions on the planar support, is built (usually, by keeping only a couple of images in order to build a simple bi-dimensional histogram ;
— the histogram maxima, that are representing regular luminous surfaces, are labeled ;
— the series of differentiated images are re-examined and labeled using the labels of the histogram maxima, according to a nearest-neighbor assignment, in order to provide a map of the regular areas of the initial image.

If it is dealt with a multispectral image, each radiometric band can be processed as a mono-spectral image and the process may offer a result richer than a usual thematic classification.

**Attribute calculus / Vectorization**

The labeling of an image into regular areas and the nature of scenes observed by a satellite – mainly composed of a lot of very small objects- lead to favor regional approaches for searching connected components.

The connected components produced at image decomposition constitute the objects of interest, on which the following processes are performed during the analysis procedure, then at registration time and at last at rendering one when querying.

At the opposite, vectorization is relying on the border of objects and leads to perform a connected components search based on a boundary approach, in the present case by the use of the border following technique.

Whatever is the followed approach, attribute calculus is applied on the two representation kinds.

It can be come back to a regional representation with the help of a boundary filling operator.

## Geometric conversion

Three points of view may appear in the handling of information:

- the sensor position relatively to the observed scene when capturing an image ;
- the reference frame used for registering data ;
- the position that an observer is expected to have at data rendering time when querying.

A geometric transformation is implied in each reference frame change. Theoretically, they are planispheric projections, consequently non linear transformations. In practice, three different situations can be distinguished according to the altitude, the field of view and the resolution of the sensor:

- hemispheric or continental observation satellites as METEOSAT and NOAA ;
- mid-resolution regional observation satellites (LANDSAT and SPOT) ;
- high-resolution regional observation satellites (HELIOS, ...).

For continental observation satellites, a planispheric projection must be performed in order to provide data transposable in a universal reference frame.

Concerning regional observation satellites, data is less sensitive to distortions due to the terrestrial geoid and plausible measurements can be performed.

Concerning high-resolution observation satellites, the relief and the buildings relying on the ground notably distort data captured using monocular sensors.

The geographic coordinates of the four image corners enable to replace these ones into a universal reference frame with an enough small error for cataloging data and measuring objects (but not for setting up geographic maps).

In the case of high-resolution observation satellites, it is necessary to get previously some knowledge about the relief and the elevation of building on the ground in order to geometrically correct images.

In order to implement planispheric transformations, it has been developed planar polynomial transformations working up to order 3. The coefficients of these transformations are determined by mean-square approximation of a cloud made from couples of points, made from their values before and after transformation.

For building the couples corresponding to a cartographic projection, it has been used the transformation library GCTP (General Cartographic Transformation Package), developed by the National Mapping Division and put in the public domain by the U.S. Geological Survey.

## Object indexing and data base browsing

The object indexing is implemented in a tree-like manner by using the KDTREE software.

The attributes computed on objects enable to set up a dictionary indexing all the objects belonging to the data base. Such attributes enable to have at one's disposal information about :

- the object localization (in position and in orientation) ;
- the luminous response (mean, dispersion, asymmetry) ;
- the object geometric shape (surface area, inertia axes, asymmetries).

These attributes are handled either in their capture reference frame, or in their registration reference frame

Four index types can be built according to the nature of generated attributes and the wished order of expansion:

- one index built on a-dimensional attributes (compactness, eccentricity) ;
- three indexes built around the generalized shape and luminous response moments at order 1, 2 and 3 including their linked localization information.

The attribute vector is used as an indexing key for retrieving an object and appears like a point in the space of the vector dimension: the localization information constitutes a sub-space of the indexing space.

By recursive decomposition, half by half along each space dimension, a binary tree is generated and the object identifier is recorded in a list hanged to the node corresponding to the object attribute vector.

The tree development precision is given at building time and defines the size of the dictionary of shapes handled with this indexing system. This procedure follows the same principles than the algorithms of vector quantification based on tree-like structures used in image compression.

The implementation was enough easy for enabling us to master a full data base with only one index and not an index decomposed in two sub-indexes as it was planned at the research beginning.

The structuring of archiving files enables by pointing out any object to retrieve the image to which it is belonging and all the objects that are included in. Two content-based querying modes have been developed:

- the one by multi-criteria selection where it is possible to specify a value interval for each attribute;
- the other one by nearest-neighbor finding where, after having given as parameter an attribute vector, it is possible to retrieve the similar shapes ranked according to a similarity index.

These queries work simultaneously on the object localization in the base and the shape parameters.

The nearest-neighbor search enables to implement a complementary mode of interrogation by example:

- an object observed in a view take independent from the base or independently digitized is analyzed in order to compute its attributes ;

- these ones are used to build a nearest-neighbor query and all the answers respecting more or less some error are retrieved and sorted according to their similarity ;
- the object localization, the belonging image and the response sorting rank enable to define a visiting path in the images of the data base for authenticating the objects of interest.

The localization information belonging to the attribute vectors enable:

- to geographically browse the data base ;
- to select the orientation of objects of interest, if necessary.

The answer is served as an extract of the data base that can have its own index and holding the information of interest under the same recording scheme.

**Data registration**

After analysis, images are registered according to the following manner :

- for a multispectral image, it is recorded as many images as there are spectral bands ;
- the attributes of each object are registered ;
- the vectorization vertices are coming after ;
- images can be gathered, if necessary, into a single file.

If necessary, it is also possible to restrict data registration only to the recording of attributes. With only them, it can be generated compact data bases.

For mastering the size of recorded data, it is also possible to select objects according to the values of their attributes, especially concerning their sizes.

This capability should enable to deliver periodical reports about the data available at a given while in an archiving center.

**Migration of KDTREE software**

This work has been achieved without too much troubles, the T9000 processor taking over using a 32 bits architecture the functionalities of T800 running on 16 bits.

To the T9000 is associated a communication processor named C104 that has been able to electronically take in account the Omega network developed on T800 and its message routing procedure.

**Data recording system on parallel computer**

Hard disks have been mounted in order to be accessible using SCSI interfaces from parallel processors according to the scheme shown underneath.

Above the own communication protocol of SCSI interfaces, it has been develop a disk management, enabling to access by blocks to data standing on them.

Tests have been performed for validating their parallel access on TN computer including 4 processors each equipped with a disk.

**Distribution of image data in the parallel disk system**

The testing images have been stored on this system.

A bank organization has been favored so as to be available for the preprocessing operations used during image analysis.

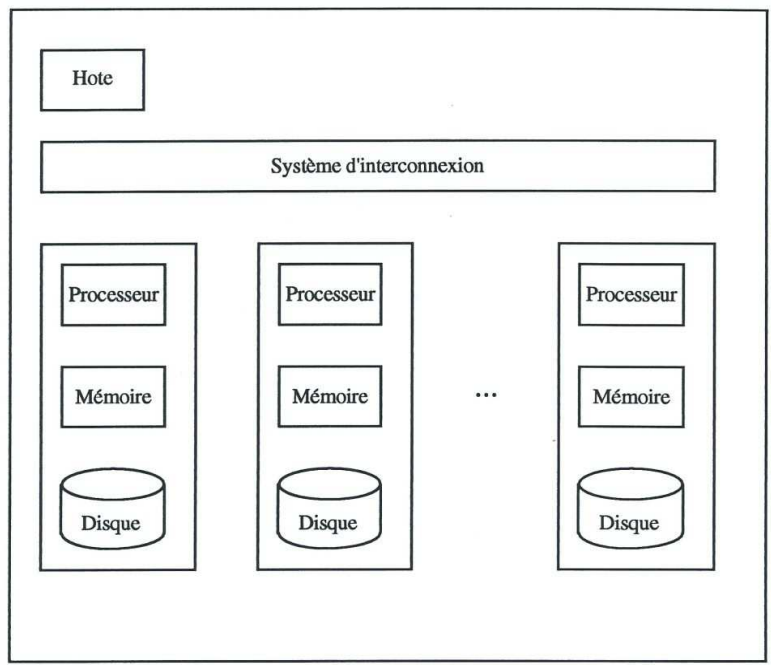

**Figure 38 : Parallel file system management**

**Parallelization of the image analysis software**

The algorithms have been classified into four categories according to the kind of memory accesses used in matter of inter-processor communication:

— the functions that are requiring no access to data external to the current processor (local processing) ;
— those that are requiring some access to neighbor processors (regional processing) ;

- those that are needing to access to data distributed all over the processors, but by using regular communication patterns (regular global processing) ;
- finally those that are randomly accessing to the whole data stored in the computer (irregular global processing).

In image analysis, the study of algorithms to be parallelized showed that it is encountered different kinds of algorithms according to the increasing degree of structuring for the information to be processed.

So concerning the functions of image analysis, taking in account the kind of parallel architecture (distributed memory), it has been preferred to use a farming model rather than a simulated global memory in order to minimize the time spent in communication.

In this kind of model, one processor is dedicated to the control of operations running on slave processors: between two operations, the master processor is reorganizing data belonging to the memory of slave processors, in order to balance the computing load over all of them for next operation.

**<u>Tests on parallel computer</u>**

Tests have been performed on a computing configuration made from 12 parallel processors.

The functions of image analysis immediately showed interesting performances on local or regional processing.

At the opposite, the KDTREE software provide less good performances comparatively to the image analysis functions : the simulation of a global memory providing much more communications based on small messages than in the farming model, where the data reorganization is based on blocks.

At the opposite, when regular or irregular global processing functions are performed in image analysis, the data reorganization is generating an increasing response time proportionally to the size of the exchanged data.

So if two processors are reaching similar performances to an ordinary work station concerning local or regional processing by using a farming programming model, it is necessary to get up to eight or sixteen processors for hoping to reach the same level of performance for global processing functions.

Despite that, this programming model allowed us to observe performance progresses varying linearly with the number of processors.

It was not the case for a programming model based on the use of a global memory, where the small size of messages produces the occurrence of a performance saddle point located around four processors concerning the evaluated system. For short messages, the communication initialization time is then overcoming on the communication duration.

It must be noticed that going from T800 to T9000 associated with C104 has allowed us to observe a true quantitative performance improvement step on parallel computers developed by TELMAT.

**Processing optimization**

The writing of the KDTREE software has been deeply re-engineered so as to decrease the number of messages sent. The result has enabled to get a speed-up ratio of about 3 after this work.

The sequential release has been also adapted and shows speed-up gains in the same ratio.

**Generation of a catalog and performance evaluation**

We have got a set of 32 mono- and multispectral SPOT images covering France. The images were provided with two different levels of correction, we have only used a linear transformation based on the coordinates of the four corners, in order to put them in a geographical reference frame limited to France.

During the tests we have at our disposal the following equipments:

- — a work station SILICON GRAPHICS INDIGO2 ;
- — a work station SUN MICROSYSTEMS IPC ;
- — a parallel computer TN 300 with 12 T9000 processors.

Concerning the tests, the set of hard disks connected to the TN computer have been temporally moved on the SUN work station in order to load the set of images.

**Catalog interrogation and performance evaluation**

The tests performed using the evaluation image set showed that, if the building of an index over the data base has not the response time of an interactive function, queries on the data base were at the opposite very fast to perform.

On parallel computer, we have met troubles with the distribution of trees in processor memories; it has been improved by modifying the bounds of the attribute space.

Otherwise, it has been observed a saddle point around 4 processors when using KDTREE, situation after which the response time is once more increasing with the number of processors.

The use of shared memory parallel computers should solve this issue and enable to find back a performance growth quasi-linear with the number of processors.

**Demonstration implementation**

A demonstration has been implemented on a platform at TELMAT. It enables to perform the following operations:

- visualization of all the objects stored in an image base ;
- interrogation using multi-criteria selection by defining, with the help of small rulers, the interval limits of each index attribute;
- visualization of all the selected structures using their vectorized representation ;
- listing in a table the object identifiers and attributes;
- displaying into another table the list of images in which a selected structure has been found ;
- for a given image, visualization of the image including the selected objects shown in superimposition.

As geometric transformations enabling to move from the image reference frame to the base reference frame was unfortunately not recorded in the archiving structure, it has given us some troubles for performing easily this last operation.

During the development of this demonstration, it has appeared that it was lacking in the recording data structure the registration of direct and inverse geometric transformations used for saving the images in the universal reference frame of the base.

**Summary of achieved results at the research end :**

- implementation of an image catalog searchable by address and by content ;
- development of a procedure of image decomposition in regular regions (image segmentation available for mono-spectral images as well as multispectral images);
- description of objects stemming from image decomposition by their attributes and their shape vectorization;
- when objects are only described by their attributes it can be got high compression rates, but without having the possibility to decode them for the moment ;
- the catalog indexing based on attributes enables to browse in a multidimensional way the data base according to object localization, luminance and geometric shape information;
- the content-based interrogation is implemented using multi-criteria selection or by nearest-neighbor querying;
- the nearest-neighbor finding enables to define a visiting path in the image base according to a decreasing similarity index ;
- by computing the attributes of a prototype shape, the nearest-neighbor querying is transformed into an example-based interrogation ;
- implementation of a demonstrator that can show all of these results.

**Further development perspectives**

At the end of the experiments lead in the framework of this research, some development efforts are remaining:

- in interrogation, in order to develop a relational algebra for building a querying language that can detect complex shapes in an image base ;
- in piecewise polynomial approximation, so as to have at one's disposal compact representation models enabling to model luminous surfaces and region contours, from data provided by an image segmentation in regular components ;

- in compression, in order to improve the performances achieved in indexing the attributes of regular components, while succeeding in implementing the synthesis of this information for rendering.

In system architecture, we have dealt until now with distributed memory parallel computers; it seems that higher performances should be observed on shared memory computers.

# 4. Person Identification and Authentication

## 4.1. Introduction

Biometrics refers to the automatic identification of a person based on her/his physiological or behavioral characteristics. Various types of biometric systems can be implemented for person identification like speaker verification, fingerprint matching or face recognition [6].

Depending on the usage context, a biometric system can be either a verification (authentication) system when it must be checked who is going out or entering in a given area or an identification one when it should be recognized who is staying at a given location.

In the framework of project EUREKA ITEA 00003 AMBIENCE, we have been more specifically interested in the provision of face recognition techniques in order to develop a person identification and authentication system with the support of the French State Secretary for Industry.

Concerning face recognition, mainly two approaches can be distinguished:

- the first one, is based on planar color image processing [1], [3], [4];
- the other one relies on 3D face modeling [5].

Planar color image processing has the main advantage not to be sensitive to illumination variations, but it does not propose to recognize objects independently to position and attitude. 3D face modeling may offer such a geometrical independence, but requires using a stereo set of cameras for capturing the necessary data to be processed. In the presently described work, it is shown how color image processing can be used for reaching this geometrical independence: from only one processing step using a single view coming from a video camera capture.

Most of video cameras are providing images in YUV format where luminosity information stands in Y band and chromatic information in a near de-correlated manner in complementary U and V ones. In order to provide some independence to illumination changes, it has been developed an image segmentation tool robust to illumination variations relying on the analysis of chromatic information.

Concerning the position and attitude face independence, a solution composed of two steps is proposed:

- first by computing geometrically independent features that insure to provide measures independent from a given set of visible planar movements;
- secondly by taking advantage of the image stream captured by a video camera and recording all feature variations that can occur in a 3D world with moving objects.

As feature calculation assumes that no object part may be occluded and that the object of interest must appear as a single connected component. These conditions are not met in the present application field: only part of a face can be viewed by a single camera and image color segmentation would decompose a face in several connected components. In comparison, 3D face modeling approach uses an elastic graph-matching tool that does not suffer from this drawback and allows recognition of partially occluded objects. To solve such an issue, it is been implemented an intermediate step based using a holistic scheme: nearby regions are successively aggregated in compound regions until the whole face can be wholly recomposed. Feature calculus is so generalized to compound regions and a single face can be then described by a series of nested subcomponents. Each of them gets its own feature vector and can then be individually recognized: so it is possible to recognize a face or an object without wholly seeing it.

This new approach is presented according to the following manner:

- in a first step, it is described how a color segmentation can be implemented;
- then, how the skin of a face can be detected in a image ;
- after that, how attribute calculus can be extended to composed objects ;
- at last, it is described a procedure for object recognition relying on multidimensional data modeling and the development of a similarity measure.

## 4.2. Luminance varying independent color image segmentation

The main variations observed during image capture about natural scenes are generally due to illumination condition modifications: with respect to reflectance and diffusion material properties, the object luminance changes accordingly to the scene illumination conditions. On the contrary, object colors stay unchanged in the visible light spectrum. It has been taken in account in the building of most digital color cameras which deliver data in a YUV data format: primary R, G, B colors are mapped in Y luminance and in $C_b$ and $C_r$ chrominance information which is relative to differential data concerning blue and red colors and provide nearly normalized color information

Gray level image segmentation will lead to take in account the Y image plane information, but luminance varying independent image analysis would prefer to do U and V planes image processing. Multi-threshold image segmentation is well suited to decompose gray level images according to luminance distributions: to achieve it a gray level histogram is built and a maximum search is applied onto it. Mono-spectral multi-threshold segmentation can be easily extended to color images following a bi-spectral approach:

- first, establish a bi-dimensional histogram that registers U,V couple value counts (it looks like a gray level image where U,V counts stand for image grey levels recorded at U and V row and column position);
- then, search for all local luminescence maxima (they stand for color distribution summits) and label them as it;
- erase all information except the maxima from the histogram and diffuse their labels all over the histogram unlabelled support;
- use this resulting bi-dimensional histogram as a look-up table for labeling U, V color planes into a pseudo-color image made from color distribution labels.

The bi-dimensional histogram maxima search and the planar diffusion of color distribution labels can be done using a watershed morphological gray level operator. Concerning label diffusion, faster processing times are afforded through computing the Voronoï diagram of luminescence maxima in place of a morphological extension.

## 4.3. Face localization using skin detection

Most of researchers agree that human skin gets a single response in chrominance and varies only in luminescence (melanin pigments are at the source of the difference by absorbing more or less light). Based on this material property of skin, face recognition systems using color imagery first tries to detect faces (or hands) and then computes geometric features or shape attributes latterly used at recognition step.

As each video camera does not necessarily respond in the same manner within color space, a camera calibration must be done at installation time in order to learn a camera to recognize skinny parts of a human being.

This task is being led in the following way:

- several people images are selected;
- each person image is recorded face centered in the camera's field of view;
- after regional analysis, chrominance couples of central components are stored as skin in a bi-dimensional histogram.

At learning ending, a two-class histogram is providing allowing to distinguish regions from subsequent images between skinny and not skinny parts. Used as a look-up table, it allows to build binary mask images for skin localization in color images. So, further image analysis can be restricted to the facial regions in an image.

The following images show how this process takes place. They hold a planar view of a two-class histogram where skin parts are displayed in white and others in gray. Skin detection is shown with the final view.

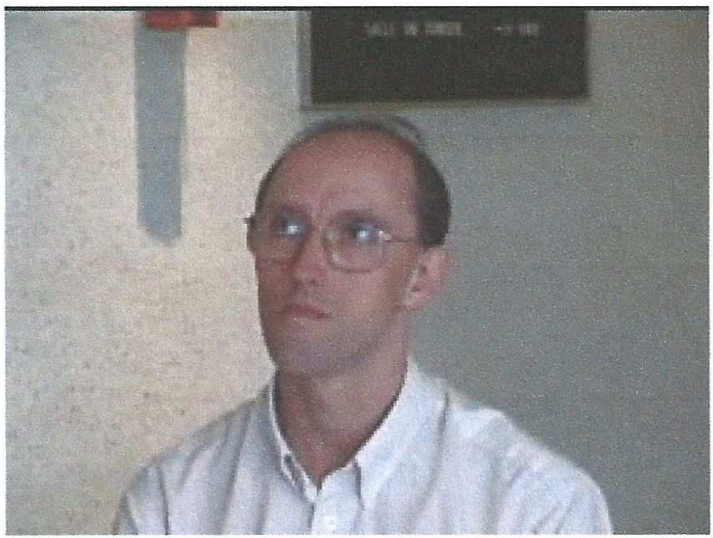

**Figure 39 : Color camera calibration: original view**

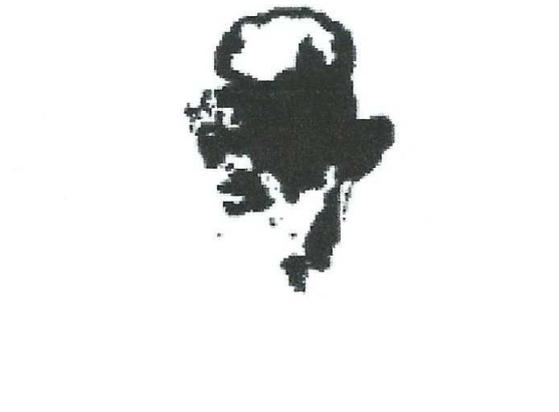

**Figure 40 : Color camera calibration: central region mask**

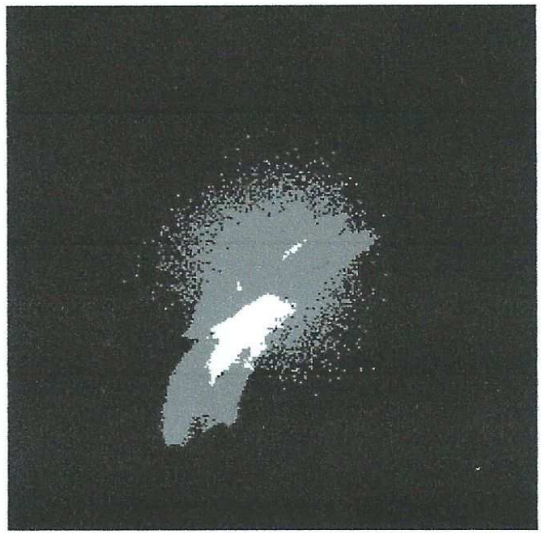

**Figure 41: Color camera calibration, bi-dimensional histogram**

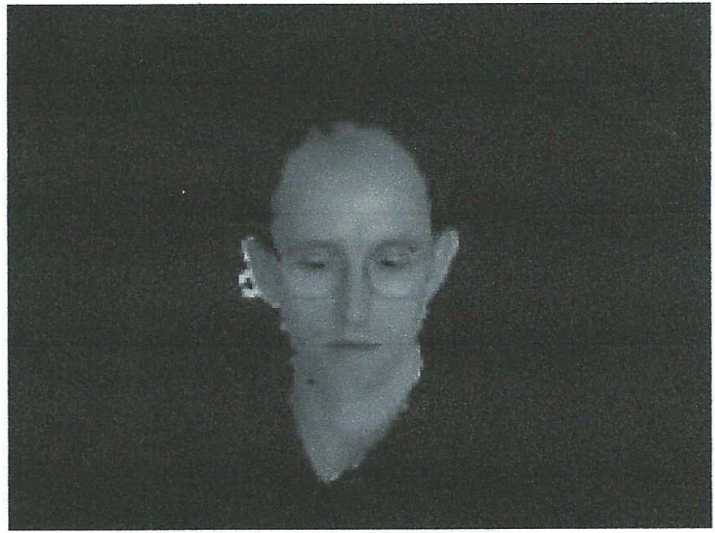

**Figure 42 : Skin parts detected in a new view**

## *4.4. Connected component attribute calculus*

Alter camera calibration, the person profile enrolment as well as the person identification or authentication includes face segmentation and component attribute calculus.

As skin detection, binary image analysis belongs to a two-class classification problem set. In this kind of application, foreground objects can be generally easily extracted from background information. At the opposite, histogram multi-threshold analysis stands in multiple-class classification problem set that provides a finer image decomposition of visible objects. So, inside a single image, small color variations can be detected and an image segmentation will lead to a decomposition of objects in more numerous regions. This property is being used in order to get richer information about a face to recognize.

So a bi-dimensional color histogram is built and a pseudo-color image is provided according to the histogram color distribution labels. An iso-colored connected component analysis is then applied in order to provide a face description in skinny region areas.

On the resulting components, a vector of measurements based on generalized moments is calculated. Using summation formulas, moments are computed until third order and the following information can be provided concerning an object:

- − at null order, object surface in image referential system;
- − at first order, object gravity center coordinates;
- − at second order, inertia axes and main axis angle nearby 180';
- − at third order, asymmetries.

The gravity center is used to define the object position, the main axis angle for object orientation and the main axis length for object scale: they compose the localization attributes of an object in an image plane.

Alter centering, normalizing and scaling, remaining values provide a feature vector whose components are independent to similar transformations (scaling, translation and rotation). They consist in shape attributes and are composed of six mutually independent values that can be reduced down to four ones by eliminating the crossed third order moments without a great loss of statistical representation power: surface with main axis ratio, eccentricity (second with first axis ratio), normalized asymmetries along main axes. With such a set of measurements, objects can be read as sheared elliptical objects like for instance ovoid shapes (asymmetric along main axis, symmetric along the other one).

## 4.5. Multidimensional data modeling

With feature vectors at one's disposal, face recognition can be led using a statistical approach:

- at learning time, features extracted from known faces are recorded in a training set;
- at recognition time, features are computed on every new image and objects are labeled according to information found in the training set.

Labeling can apply using some suitable statistical data analysis process.

It is used an indexing scheme for registering labeled features of a training set, based on binary tree structures. For modeling a numerical data set standing in a four-dimension space, data can be recorded in a binary tree using feature vectors as geographical keys. Feature vectors can be used for placing data at a right position in such a tree describing a value set. To define vector location, a similar visiting traversal is applied as in a quad-tree or oct-tree search:

- according to the divide and conquer paradigm, the multidimensional space is divided half by half successively along each space direction until reaching a modeling precision value given as a parameter search;
- at each space division step, vector coordinates are examined and a decision is taken about which half-space side, among left and right, must be visited to achieve successfully a search until the analysis precision is reached.

So a multidimensional space can be viewed as a series of overlapping quadrants, octants, hexadecants, according to the feature space dimension. Their union gets back the whole initial space in which they are mapped and a simple binary search allows to locate any given point or subset in this modeling space. The precision is a parameter used for construction or retrieving: it corresponds to the tree building or visiting depth.

So a training set can be modeled by building at a given precision a person identification label tree indexed by their feature vectors. If the precision is small then the recognition space will be coarse and at the opposite, if the precision is high, it will get finer.

When training sets are overcrowded, a high modeling precision can be achieved. But when training sets are sparse, it is possible to keep a good precision in adding a label diffusion process by merging alone regions with empty nodes in the modeling tree. The process is similar to a Voronoï diagram construction, but produces a different result because the process implies an ultra-metric distance in place of an Euclidean one. The labels can be then extended up to space bounds according to their relative positions. As feature vectors are made from normalized data, the space bounds can be set to unity.

The image samples used to build a learning base are shown in the figure underneath: three distinct persons are recorded using different positions and attitudes. The face attributes will be used to index people names in an identification tree.

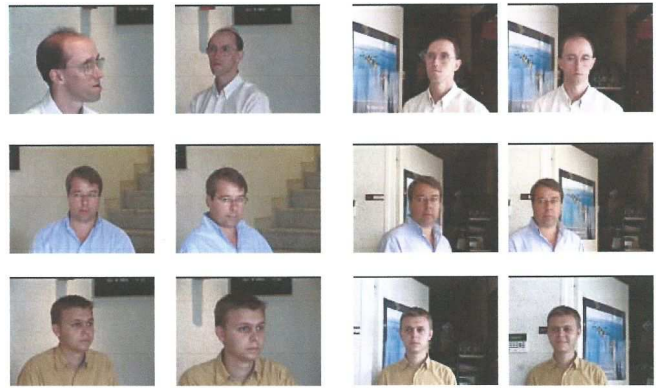

**Figure 43 : Image samples captured for building a learning base**

## 4.6. Similarity measure and recognition threshold setting

Precise multidimensional data modeling can be afforded with overcrowded training sets. Dealing with sparse training sets, it is necessary to include similarity measurement constrained to some acceptance thresholds. Threshold calculus generally implied to divide a training set into two parts for avoiding over-fitting issues: one of them is dedicated to the recognition database construction and the other one to the threshold estimation.

Multidimensional data set modeling through tree-like hierarchical structures implies that data sets are managed under an ultra-metric distance. Inside regularly divided spaces, the structuring distance is a Hausdorff distance: the distance between two points may straightly valued as the minimum subset size that can hold simultaneously these points. It is not a true distance for point nearness measurement, but more rather a pseudo-distance used for subset comparison.

As the features are normalized, it can also be used an un-weighted Euclidean distance in the aim of building a similarity measure. The effective feature coordinates are held in tree structure and allow to compare straightly a new feature vector occurrence with a previous one recorded in the training set. So the conditions are then more compliant for measuring similarity between known and unknown objects. In addition, such a similarity measure allows to sort all the objects of a multimedia database according to their similarity with a sample (query by example) or to do it according to their self-similarity (database sort).

## 4.7. Shape aggregation

Moment-based attribute calculus assumes that objects to recognize are viewed on a whole. It is not actually fulfilled with facial person recognition applications. Statistical learning made from the recording of several views of a same person observed with distinct head positions and attitudes, allows to create a sufficient dense set of feature data for expecting to recognize any incoming new view standing near some previously recorded ones.

But the moment-based attribute calculus using low-order formulas allows only to identify compact shapes. High-order formulas would introduce more digitalization noise in measurements and would decrease recognition precision. On another side, image color segmentation provides a richer object decomposition in more numerous regions. To take benefit of this property, it has been designed an aggregative process that enables to take in account most of region groupings that may appear in a person face. Usually region merging proposes to aggregate adjacent regions. To avoid building up a regional adjacency graph, it has been developed an alternate aggregative approach based on region nearness. During region aggregation, attribute calculus is performed without centering, normalizing and scaling. As moments are integral formulas, attributes can be then straightforwardly summed up before being post-processed. Using gravity center information, a regional gravity center quad-tree can easily be built and allows to merge nearest regions using a bottom-up tree traversal. As the merging process is recursively repeated up to tree root, a person face can then be successively decomposed in a nested set of regions.

After attribute summation, features are centered, normalized and scaled at every decomposition level of a face model. A face can be then described as a variable length series of feature vectors. At training time, the labels are applied on vector series allowing to provide a recognition step running on wholly or partially visible objects. This approach was firstly defined by psychologists who have developed the Gestalt Theory of perception in the 1900s [2] and offers a new outlook for machine learning: it can hold simultaneously the advantages coming from statistical as well as structural pattern recognition.

The shape aggregation process is shown underneath on a chosen example. The attribute list of different objects produced during the aggregation of the corresponding images is displayed just after. They consist in location and shape attributes. Gravity center coordinates, object angle and scale are location attributes in the image reference frame system. Surface, eccentricity and asymmetries compose shape attributes. They consist in centered, normalized and scaled values, which are independent from any translation, rotation and scale change in the capturing reference frame.

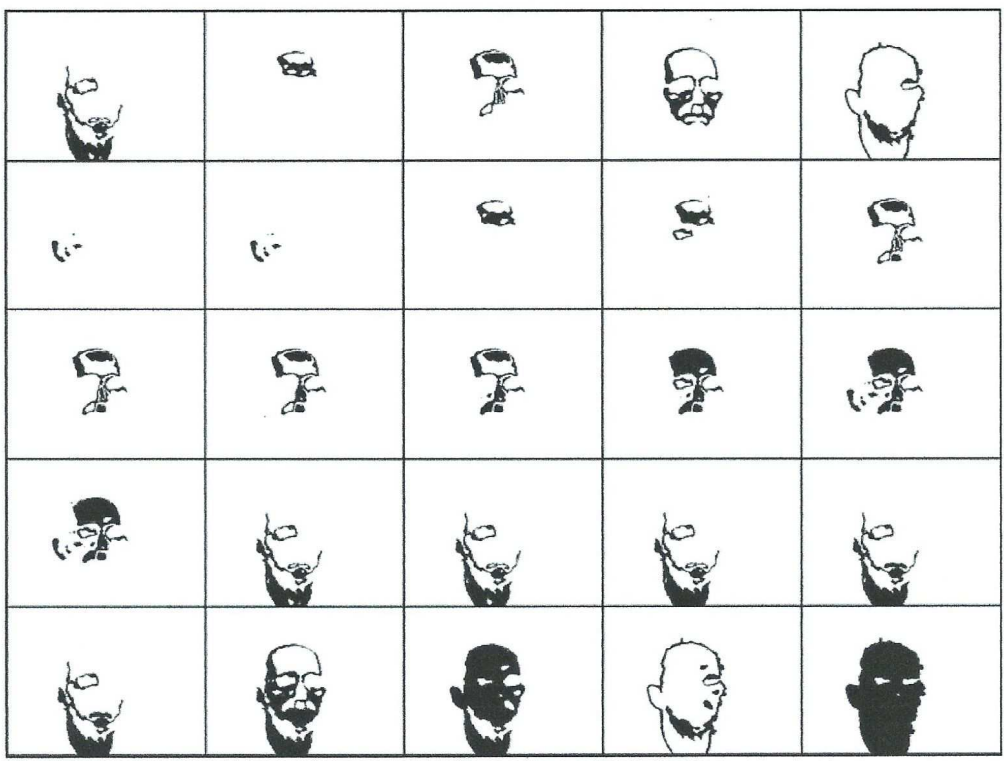

**Figure 44 : Region hierarchical aggregation**

| Object n° | Center Abs. | Center Ord | Angle | Scale | Surface | Eccentric. | 1st Asym. | 2nd Asym. |
|---|---|---|---|---|---|---|---|---|
| 0 | 169.279 | 190.920 | 273.545 | 59.559 | 0.4747 | 0.7057 | 0.8579 | 0.3849 |
| 1 | 168.971 | 160.698 | 238.319 | 39.068 | 0.7220 | 0.7628 | 0.7980 | 0.5896 |
| 2 | 176.130 | 121.729 | 77.629 | 34.416 | 0.6487 | 0.7082 | 0.7952 | 0.1673 |
| 3 | 178.040 | 105.748 | 15.728 | 20.979 | 0.7661 | 0.5030 | 0.1476 | 0.3890 |
| 4 | 153.887 | 226.375 | 249.587 | 47.281 | 0.6608 | 0.5542 | 0.8516 | 0.4414 |
| 5 | 109.085 | 173.296 | 350.625 | 16.242 | 0.6306 | 0.7493 | 1.0041 | 0.2672 |
| 6 | 109.190 | 173.068 | 348.652 | 16.265 | 0.6320 | 0.7594 | 1.0039 | 0.3471 |
| 7 | 177.521 | 106.380 | 14.977 | 20.360 | 0.8164 | 0.5211 | 0.3752 | 0.4272 |
| 8 | 173.352 | 112.506 | 339.856 | 20.794 | 0.8888 | 0.8004 | 0.4257 | 0.5166 |
| 9 | 176.674 | 126.672 | 79.848 | 37.619 | 0.6162 | 0.6263 | 0.7472 | 0.2434 |
| 10 | 176.769 | 127.016 | 79.740 | 37.723 | 0.6167 | 0.6225 | 0.7359 | 0.2379 |
| 11 | 177.078 | 128.370 | 79.654 | 37.591 | 0.6319 | 0.6125 | 0.6774 | 0.2259 |
| 12 | 176.152 | 132.086 | 84.307 | 38.740 | 0.6347 | 0.5916 | 0.5893 | 0.2757 |
| 13 | 175.141 | 125.018 | 85.145 | 33.952 | 0.9060 | 0.6508 | 0.8530 | 0.3482 |
| 14 | 168.515 | 129.846 | 115.188 | 37.001 | 0.8765 | 0.7405 | 0.8251 | 0.2190 |
| 15 | 168.470 | 129.799 | 115.066 | 36.988 | 0.8773 | 0.7421 | 0.8255 | 0.2340 |
| 16 | 153.939 | 226.471 | 249.548 | 47.319 | 0.6609 | 0.5533 | 0.8515 | 0.4404 |
| 17 | 153.982 | 226.550 | 249.521 | 47.341 | 0.6610 | 0.5527 | 0.8518 | 0.4396 |
| 18 | 154.134 | 227.217 | 249.691 | 47.404 | 0.6645 | 0.5487 | 0.8589 | 0.4397 |
| 19 | 154.838 | 226.784 | 249.857 | 46.836 | 0.6810 | 0.5600 | 0.8577 | 0.4307 |
| 20 | 154.742 | 226.803 | 249.846 | 46.771 | 0.6829 | 0.5613 | 0.8577 | 0.4324 |
| 21 | 160.977 | 197.839 | 264.764 | 53.025 | 0.8036 | 0.5928 | 0.5511 | 0.4162 |
| 22 | 163.727 | 172.865 | 92.444 | 57.546 | 0.9308 | 0.5386 | 0.6141 | 0.2124 |
| 23 | 172.523 | 188.593 | 275.887 | 58.201 | 0.5090 | 0.7112 | 0.8155 | 0.5023 |
| 24 | 165.788 | 176.550 | 92.379 | 58.052 | 1.0544 | 0.5853 | 0.3648 | 0.2288 |

**Table 7 : Aggregated region attribute list**

## 4.8. Conclusion

All the planned experiments have not been completely achieved during the research presented above.

The followed approach does not only apply on face recognition. Shape attribute calculus can be enhanced with luminance and/or chrominance attributes which can be also based on moment formulas but applied in response space in place of image plane. In face recognition color attributes has been distributed in two classes, skinny and not skinny regions, but they may be generalized to a multiple-class problem. Then face recognition will appear as a singular case from a more general application field. It was in the mind of the authors during the development of this application and for which a wider aim was envisioned: it was to set up a toolbox compliant with most of the MPEG-7 features for the development of visual applications.

Concerning the face recognition application field, the developed technique can be applied in two distinct ways: person identification or authentication. In each case, a training set must be set up, but with person authentication only feature vectors concerning the person, which identity must be checked, are used to build a recognition database.

If training sets are sufficiently large, person recognition or authentication has not to be constrained to an acceptance threshold. At the opposite, a threshold calculation must apply with sparse sets. In every case, a complementary data set must be acquired to validate a training set before using it for recognition.

Finally, it has been developed a facial recognition system composed of the following parts:

- camera color calibration;
- person profile enrolment;
- recognition database management;
- person identification/authentication procedure.

Camera color calibration is applied once at installation time. A series of images of distinct people are captured face centered in the image. The color pixels belonging to corresponding regions are stored and labeled as skin in a bi-dimensional histogram. The other ones are stored and labeled as not skin. Alter processing all images, the unlabeled regions are marked with the nearest label and the histogram can then be further used as a lock-up table for skin detection in any new image.

Personal profile enrolment needs also to capture series of views belonging to a same person but using different face positions and attitudes in the aim to get a representative set of person features. Inside each view, skin parts are segmented and resulting regions are successively grouped using shape aggregation. Feature vectors are calculated along all initially segmented regions as well their groupings. Labeled with person identity, they are stored in a learning base that will include information concerning:

- only one person for authentication purpose;
- several people for identification usage.

At recognition time, this feature database can be used by reapplying the process composed of image segmentation, region aggregation and feature calculus to determine which person can be identified or authenticated. Results can be constrained or not to a similarity acceptance threshold.

The application has been tested using a personal computer processing standard format video streams. The response time was close to real-time.

# 5. Self-Descriptive Video Coding

## 5.1. Scientific and technical project description

### 5.1.1. Project objectives

Visual representation and compression techniques so far have been defined independently of the needs for editing and indexing for database retrieval. Many visual processing tasks or database management tasks are therefore uneasy as they often require to first de-compress the content. The objective of the project is to define a new and flexible visual content representation paradigm to be used as a common framework for three application domains: compression, indexing and editing. This new paradigm, referred to as self-descriptive coding of visual content, aims at providing an answer to the pressing need of having a single representation amenable to storage, protection, retrieval, rendering and editing of visual information.

While current visual representation models remain close to the digitized information captured by the input sensors, the targeted paradigm will rely on a layered information model which will structure the information present in still images and videos, that is, row pixel information, geometric structures, relationships between regions and visual objects. These different kinds of information are complementary and can be retrieved one from the other along a progressive path leading naturally to a pyramidal organization of visual information.

The first objective will thus be to develop a mid-level representation model that will allow us to manipulate image and video content from its homogeneous components. A geometric invariant description of the different components will be defined. These components can be regarded as visual primitives trained and learned from the content and stored in a dictionary. The visual representation paradigm will allow us to manage the image components separately or grouped in order to facilitate content-based editing and retrieving as well as compression.

The descriptors will be used to code, store and access the different visual primitives (describing texture, shapes, and so on) and eventually the entire content. The representation model will be independent of the resolutions of capture and rendering. In the case where visual descriptors are invariant to similitude transforms (scaling, translation and rotation), the compressed representation and the corresponding syntax visual units will be sorted according to their scale factors in order to allow for scalable content access, coding and transmission.

The benefits of this new paradigm will be demonstrated for the three targeted applications: compression, editing and indexing. The success criteria will be: in compression to get compact codes preserving visual quality, in indexing to retrieve similar content by using only codes at querying and in editing to easily create and modify the content with the single help of visual dictionaries.

### 5.1.2. Background, state of the art, troubles and hypotheses

Content-based image/video retrieval systems research has been mainly focused on low-level feature extraction techniques [1]. Along this path, important results have been obtained with scale-invariant image description using multi-resolution wavelet decomposition [2]. Moreover, a lot of efforts have been addressed to automatically extract semantic information from pixel-based image representation but without a convincing success [3].

Object-based image retrieval is a possible candidate for helping in bridging the semantic gap. In the framework of MPEG-4/MPEG-7 standardization activities, efforts has also been provided in trying to extract and describe visual information [4], but visual appearance of 3D real-world objects cannot be easily recover from numerical objects produced by segmentation techniques. Applying perceptual laws, visual groupings can be retrieved and help in bridging the gap between numerical objects and real ones [5].

Multi-resolution techniques has recently evolved and result invariance properties have been extended from scaling up to a larger set of geometrical transforms that is including now also translation and rotation [6].

Applying scale-space theory [7], it is now known that object descriptors can be defined in their own description space and can afford robust descriptors for retrieving efficiently shapes in a database [8]. Scale-space descriptors lead in a natural way towards hierarchical indexing which can be used to easily create visual ontologies [9].

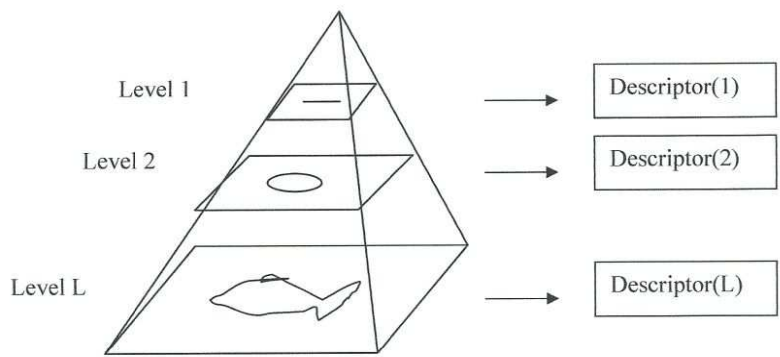

**Figure 45 : Multi-level descriptors**

In terms of data representation, it is also known that visual objects can be characterized by different types of regions (smooth, texture, edges, and so on), generally highlighted by the order of singularity of the input signal. In this respect, the literature clearly illustrates that there is no single basis (i.e., decorrelating transform) capable of providing an optimal signal representation. For instance - to quote but a few of the latest state-of-the art transforms - ridgelets [10] illustrate their strong power of approximation in particular on functions having linear singularities; wedgelet-based [11] tree-structured representations generally handle very well cartoon-image components (i.e., edges without any adjacent texture), but give poor results on ridge-alike image components; curvelets [12] and contourlets [13] are well adapted to curvilinear singularities but become highly sub-optimal on textures; classical separable 2D wavelets [14] are highly effective on zero-order singularities but fail on higher orders; finally, bandelets [15] are adaptive bases dependent on the presence of geometric flows, that behave excellently on curvilinear singularities, but revert back to the approximation power of a classical wavelet in the absence of such flows. Furthermore, given that objects are to be described by geometric invariant features, it is critical that the transform(s) synthesizing them possess shift invariance properties [16, 17].

Under all these design constraints, it is clear that constructing appropriate visual object descriptors requires appropriate tools (transforms). In this sense, in our approach we propose the combined use of a multi-scale geometric transform [17] in order to synthesize object contours, respectively of a wavelet transform [16] to synthesize high texture object regions. This will advance the state-of-the-art and will bring a two-fold benefit:

- the construction of a complete geometric invariant object descriptor based on both shape and texture information, and
- as an inherent side effect, provide an optimally sparse multi-resolution representation for the later purpose of coding [18,19], in which each transform is piece-wise optimal with respect to the type of data (object region) it aims at synthesizing.

The approaches proposed for the design of the image indexing and retrieval scheme will be further extended to video. In this respect, one of the solutions SVDC intends to follow is based on active region-based tracking with shape priors. Visual object tracking can be formulated here as the process that matches the object regions in an image sequence. Tracking is made difficult by occlusions, erratic or fast object motion, and appearance changes due illumination and pose variations, out-of-plane rotations, etc. In coping with these adverse conditions, of paramount importance is the object description itself, i.e. a set of the object descriptors (object signature) that can distinguish it from other objects and background. Typically, the object can be described by some statistics related to its color properties, texture or shape. Coming up with a good object representation is then a challenging task, as robustness and flexibility should be traded-off for complexity. Indeed, for example, using a rotation and scaling-invariant description such as the color histogram leads to a fast tracker [20] that can withstand partial occlusions but will have difficulties when illumination changes. At an increased computational colt robustness can be achieved by integrating multiple cues/features [26] and this is the direction we follow here.

SDVC will combine several object descriptors and image-derived information in a Bayesian tracking framework. More specifically, we pose the tracking problem as region-based segmentation via a level set approach. This is motivated by the capability of the level set methods [21] of providing a general and flexible framework for image segmentation [22]. In these methods, the boundary of an image region is represented as the zero level set of a two-dimensional function referred to as the level set. Starting with an arbitrary initial boundary, the level set can be evolved in time by optimizing an energy functional depending on the similarity between the prior information about the object and the characteristics of the image patch delineated by the current boundary. There are other methods to evolve a contour such as snakes [23], but only the level set methods can handle well the topological changes (split and merging of regions). Tracking based on level set approaches and multiple image descriptors was already tackled in the literature [24, 25].

In conclusion, it is envisaged that the success of the SDVC will build the settlement of a new self-descriptive video representation scheme that could be used for editing and retrieving image and video by staying at a syntactic level.

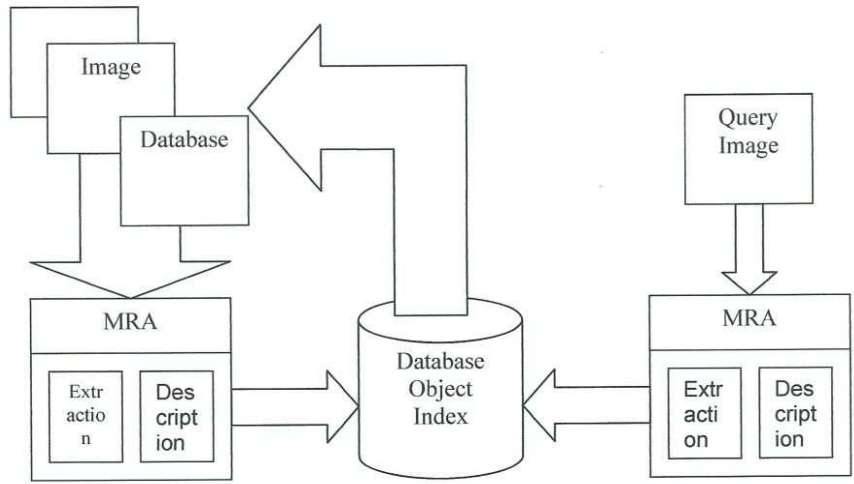

**Figure 46 : Object-based query system**

### 5.1.3. Proper project innovations

During the past 20 years, research on image and video compression has mainly focused on pixel-based representations (JPEG, H.261/3, MPEG-1/2/4, JPEG-2000, H.264). Most of these coding strategies rely on a block partition and on the use of transforms such as the Discrete Cosine Transform (DCT) or Discrete Wavelet Transform (DWT). These strategies, only driven by compression performance, are thus only based on frequency decomposition of the original signal, and disregard higher-level information such as the geometrical information. The introduction of a limited amount of higher-level information (regions, shape) has been tackled in standards dealing with content-based representation such as MPEG-4 (object profile), for compression, and MPEG-7, for indexing. In the framework of MPEG-4/MPEG-7 standardization activities, efforts have also been provided in trying to extract and describe visual information [4], but visual appearance of real-world objects cannot be easily recovered from numerical objects produced by segmentation techniques. Content-based image/video retrieval research has been mainly focused on low-level feature extraction techniques [1]. Efforts have been dedicated to automatically extract semantic information from pixel-based image representation but without a convincing success [3], due mostly to the lack of a proper framework to handle higher-level information.

Multi-resolution techniques have also recently evolved, extending invariance properties from scaling to a larger set of geometrical transforms including translation and rotation [6]. New transforms (e.g., bandelets [15], curvelets and contourlets [17]) have been designed to better take into account - and capture - geometrical patterns present in images. Applying scale-space theory [7], visual descriptors can be defined in their own description space and robust descriptors can be used for retrieving efficiently shapes in a database [8]. Detection of interest points and extraction of features invariant to geometric transform and robust to viewpoint and illumination changes is now possible with the Scale Invariant Feature Transform (SIFT) [26] on which most of efficient image retrieval systems are currently based.

The SDVC project will go beyond, first by providing scientific advances on each individual technology and second in the development of a complete representation model bringing together and fostering the convergence of these constituent technologies into a common framework amenable for different application fields. Although the SDVC project will mainly focus on compression, editing and indexing, special care will be given to the possible use of the model in a context of content protection and security.

In terms of data representation, visual units will be characterized by different types of regions (smooth, textured, edges, and so on), generally highlighted by the order of singularity of the input signal. A combined use of a multi-scale geometric transform in order to synthesize visual contours with a wavelet transform to synthesize visual textures will allow to reconstruct visual units both using shape and texture information. Sparse multi-resolution representation algorithms using basis functions having different characteristics will allow both extracting smooth regions, texture clustering, coding and in-painting.

Feature extraction and description technologies will be extended to regions of interest in space and time [24]. Scale-space descriptors will lead in a natural way to hierarchical indexing which eventually can be used to create visual ontologies [27]. Describing the structure and geometry of the information will allow us to improve the compactness of the representation and at the same time to handle content-based functionalities for edition and retrieving. Applying perceptual laws, visual groupings can be retrieved and help in bridging the gap between numerical objects and real ones.

This new representation paradigm will thus rely on the storage of normalized representations of shape, texture, or other visual primitives in dictionaries as shown in Figure 59. Images will be only described by the visual primitives and their geometric transform enabling to replace them in the image plane or the video sequence. Patches of surfaces or bounded wavelets will enable coding, rendering of visual content over geometrically defined piecewise supports, as well as editing in the form of manipulation, deletion or insertion of image regions.

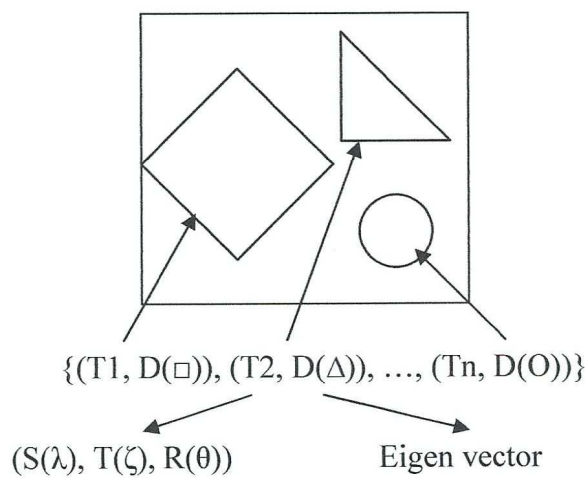

$\{(T1, D(\square)), (T2, D(\triangle)), \ldots, (Tn, D(O))\}$

$(S(\lambda), T(\zeta), R(\theta))$      Eigen vector

**Figure 47 : Self-descriptive visual coding syntax**

The breakthrough of the project is the introduction of the concept of visual dictionary on which visual content can be described and manipulated thus proposing a new path for narrowing the semantic gap in visual information based on a mid-level representation.

### 5.1.4. Progress beyond the state of the art

Some FP6 IST projects have already focused on image and video representation and retrieval systems and provided some interest results, as:

- VIBES (IST-2000-26001) that has proposed a content-based representation of video data, which emphasizes the geometric, photometric and dynamic components of video information;
- SCHEMA (IST-2001-32795), a Network of Excellence, that has worked on Content-based multimedia analysis, retrieval processes which understand the nature of the information in the database, semantic web technologies, and used MPEG-7 and MPEG-21 standards;
- DSSCV (IST-2001-35443) that has settled the scale-space techniques for analyzing the deep structure of images and applied them to medical imaging;
- PROFI (IST-2004-511572) that tries to retrieve figurative images from large databases by extracting and matching perceptually relevant shape features using perceptual grouping, geometrical pattern matching and shape feature indexing.

However the project has a more foundational nature and involves implementing a long-term research activity which requires gathering different skills in computer vision, video analysis, signal processing for compression that are present at the European level. The European dimension is also essential from an expertise point of view, but also to ensure that the outcome influences positively standardization bodies thus enabling European industry to get a competitive advantage on a world-wide market in the future. The targeted paradigm goes clearly far beyond current visual representation models and techniques which remain very close to the digitized signal captured by input sensors. By essence the research proposed is high risk, however it will capitalize on recent advances in the areas of content description, multi-scale and anisotropic signal representation and leverage developments made for other media as speech and audio, e.g. on source separation which have common scientific foundations. The concept and the project results are expected to pave the way of future research in the area of visual coding and indexing. Also, a long term high pay-off in a large number of application sectors, is expected, going from professional or user generated content production, to seamless access and delivery.

The project stands as the first step of a long-term vision in which the foundations of audiovisual information would be renewed. This vision is relying on the assumption that there exists a way to structure digital information such that the most meaningful part can be extracted and that the most efficient way to do it is to find a language that can describe how information is organized.

Besides that, textual information like words and sentences can now easily be managed since an algebraic framework, the Structural Query Language (SQL), has been found for efficiently retrieving data in large databases.

Then this vision is powered by the possibility to find out some visual languages making it possible to describe and code audiovisual content on which ail retrieval tools that are successfully used for managing textual databases should apply. The first step towards this vision relies on the definition of a visual alphabet that would help to manage lexicographically image and video content.

## 5.2. Detailed description of the work

The targeted paradigm requires foundational research in a set of disciplines such as image and video analysis including feature extraction, anisotropic wavelet and sparse representations, global and local descriptor extractions. The project gathers expertise from the different communities: analysis and computer vision, indexing, signal processing for compression.

The project will be structured into several packages, according to the enabling technologies coming from these different disciplines as well as to the targeted application fields as follows:

- image analysis (segmentation including region grouping, feature extraction);
- self-descriptive representation (dictionary building and management);
- compression (synthesis, residual coding, rate-distortion analysis);
- searching (query-by-example and semantic query);
- editing (compositing and post-processing);
- simplification, integration and evaluation.

*Image and sequence analysis*: Scale-space segmentation and feature extraction will be performed together. Concerning the time scale, optical flow and object tracking techniques will be compared. All studied techniques will be in line with scale-space theory in detecting stable information along several levels of space or/and time descriptions and in computing geometric invariant features. It will be extended to perceptual groupings in order to favor a compact description of visual content.

*Self-descriptive representation*: Techniques to construct, organize and access the dictionaries of visual primitives will be developed. It will enable region or grouping indexing in order to retrieve complex visual units in an image or a sequence. Hierarchical indexing schemes would enable to describe visual units in a scalable way. Hierarchical indexing is closely linked to classification techniques in statistical data analysis: visual dictionaries should then be built by using unsupervised learning techniques. As training sets are not necessarily made from the data to be encoded, dictionaries get a wider spectrum of use by enabling the coding of different image or sequence sets. Dictionaries need not to be sent with the encoded content and may provide a useful key for protecting the content when it is sent independently to a consumer. If a same dictionary is available for a full collection of images or sequences, it will represent the settlement of an ontology for describing the content held by this collection.

*Compression:* Visual units extracted by image and sequence analysis and registered in a dictionary using a set of descriptors must be reconstructed at rendering stage. A multi-resolution representation must be defined to model their shape and their texture in order to store and render the objects independently of the capture and display resolutions. In this context, developments will lead to the study of adaptive geometric transforms on piece-wise multi-resolution representations. Residual coding will be set up in order to provide an encoding close loop and be able to perform a rate distortion optimization on the encoding content. Sparse signal representations will actually be used as a common framework for texture classification, clustering and synthesis, for prediction and for coding. These techniques are indeed powerful tools not only for compression but also for texture analysis and synthesis, for prediction and for in-painting which can be regarded as error or loss concealment. There is an intriguing relation between sparse representation and clustering. The transformed versions of the signal (an image) lie on a low-dimension manifold in the high-dimensional space spanned by all pixel values. These representations together with adaptive subspace self-organizing maps turn out to be amenable for image texture description.

*Searching:* A query-by-example interrogation system will be developed in order to show the efficiency of using visual dictionaries for image and video retrieval. In complement, a semantic layer that uses this mid-level representation will be developed to be able to build a real-life application where content can be retrieved in a given applicative context. In this manner, we will show how a mid-level representation for visual content may narrow present semantic gap.

*Editing:* The objective will be to show the benefit of the representation for video editing. The editing task will rely on the use of a visual dictionary that will be used as a library of visual units and groups (taking into account the relationships between visual units). Using such an approach will need to revisit object grouping, in-painting and matting techniques that allow deleting, replacing or inserting visual units in an image or a video sequence. In-painting techniques will be based on sparse representations also used for texture clustering and coding. It is expected that image and video compositing and post-processing will be improved by relying on an object-based representation.

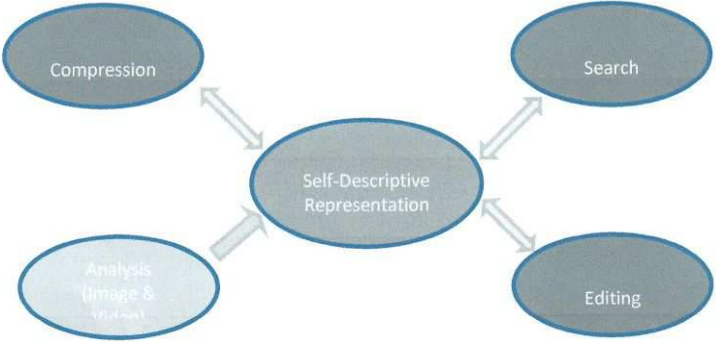

<div align="center">

**Figure 48 : Applications of self-descriptive visual coding**

</div>

*Simplification, integration and evaluation:* This particular work package will aim both at simplifying algorithms in order to provide either near real-time performance or human acceptable response time, and at integrating the results from the different technical work packages into a software platform. This platform will be used to demonstrate the benefit and to assess the performance of the approach for the three targeted application domains.

## 5.2.1. Image and sequence analysis

**Description:** Scale-space segmentation and feature extraction will be performed together. Concerning the time scale, optical flow and object tracking techniques will be compared. All studied techniques will be in line with scale-space theory in detecting stable information along several levels of space or/and time descriptions and in computing geometric invariant features. It will be extended to perceptual groupings in order to favor a compact description of visual content.

**Task 1.1: Scale-space segmentation and feature extraction:** A multi-scale approach will be used to analyze the geometric flow. Singular theory principles will be applied to insure that image decomposition would provide sufficiently regular patches in the transform domain to limit coefficient series expansion when signal has been in turn decomposed. Still image analysis is relying on differential geometry properties in order to identify smooth areas and will be based on a multi-resolution approach so as to find out a multi-level organization. Differential geometry features will be used to detect ridges and edges in order to prepare the use of geometric wavelets or meshing techniques to represent efficiently smooth areas.

**Task 1.2: Time scale analysis:** The spatial approach studied in the previous task will be extended to the temporal domain in order analyze the optical flow. Motion analysis compliant with affine displacements will be studied so as to take in account the visible motion of visual objects. This will allow following the evolution of steady structures in a wider range of geometric transforms than those currently used in video coding. Generalized motion estimation, optical flow estimation, lazy object tracking including occlusion detection will be studied.

**Task 1.3: Perceptual groupings:** The extraction of meaningful structures will be based on automatic scale detection and perceptual grouping. This will enable to simplify image and sequence description and to provide a mid-level representation for visual information. Groupings based on human visual perception can improve this structuring by providing a data organization that would be widely agreed without using some a-priori knowledge about a visual scene. The main perception laws that will be taken in account are symmetry, convexity and proximity of visual entities for allowing groupings.

**Deliverables and due dates:**

- T0 + 06 months: Spatial multi-resolution analysis report
- T0 + 12 months: Temporal multi-resolution analysis report
- T0 + 18 months: Perceptual groupings and feature generalization report

## 5.2.2. Self-descriptive representation

**Description:** Techniques to construct, organize and access the dictionaries of visual primitives will be developed. It will enable region or grouping indexing in order to retrieve complex visual units in an image or a sequence. Hierarchical indexing schemes would enable to describe visual units in a scalable way. Hierarchical indexing is closely linked to classification techniques in statistical data analysis: visual dictionaries should then be built by using unsupervised learning techniques.

**Task 2.1: Region and grouping indexing:** One of the major project's objectives is to design scalable dictionaries. A first approach will stay on the use of redundant dictionaries implied by Matching Pursuit techniques. To improve the search efficiency, a second one based on Multidimensional Binary Search Trees will be studied. It will afford a multi-resolution indexing scheme that would also provide scalable dictionaries. Applied to shape and photometric descriptors, it can be an efficient way to retrieve visual primitives into a dictionary and can be compared to vector quantization.

**Task 2.2: Visual primitives' dictionary building:** In this later case, supervised/unsupervised statistical learning techniques will be developed in order to provide means for automatically dictionary building by using training sets of annotated or not images and video sequences.

**Task 2.3: Content protection:** The use of geometric wavelets in combination with machine learning techniques will show that object dictionaries relies closely on their training sets but that also different dictionaries can be used to code a same picture or sequence. It will be shown that information confidential would preserve during transmission by sending separately the content with the dictionary on which it is coded.

**Deliverables and due dates:**

- T0 + 12 months: Region and grouping indexing report
- T0 + 18 months: Dictionary building techniques report
- T0 + 24 months: Content protection experimentation report

### 5.2.3. Compression

**Description:** Visual units extracted by image and sequence analysis and registered in a dictionary using a set of descriptors must be reconstructed at rendering stage. A multi-resolution representation must be defined to model their shape and their texture in order to store and render the objects independently of the capture and display resolutions. In this context, developments will lead to the study of adaptive geometric transforms on piece-wise multi-resolution representations. Residual coding will be set up in order to provide an encoding close loop and be able to perform a rate-distortion optimization on the encoding content.

**Task 3.1: Multi-resolution representation:** Multi-scale geometric data representations will be considered in the project. Their extension to video data representations is still an open field of investigation, and is one which we clearly intend to tackle. For instance, a possible approach for advanced scalable video representation would be to define three-dimensional time-space geometric flows and to construct geometric wavelet bases that are adapted to the space-time geometry of the video sequence.

**Task 3.2: Visual object rendering:** The use of geometric wavelets in combination with machine learning techniques will require descriptors that should not only identify the visual primitive but would also provide information related to the geometric/non-geometric data representation that synthesizes it. The link between shape and photometric descriptors with visual primitives will be clearly settled, so as to find a one-to-one mapping between descriptors and primitives inside a dictionary.

**Task 3.3: Rate-distortion optimization:** Coded representation of visual objects offers intrinsic scalability: in spatial domain by visual objects selection according to their size, in temporal one by frame rate sub-sampling and interpolation, in rate transmission by choosing more or less detailed dictionaries (uniform quantization through dictionary resolution selection). Code truncation in spatial, temporal and rate domain can be locally adapted inside an image and a sequence so as to minimize the distortion according to a given rate constraint. It will be studied how to make decision at an object level in order to reach an optimal result at an image or a sequence level. Coded representation will be handled entropy coded so as to enable comparison with currently available image and video compression algorithms. It will be then studied the optimal use of descriptors in relation to rate distortion, indexing efficiency and/or editing ease and bit-stream creation.

**Deliverables and due dates:**

- T0 + 18 months: Multi-resolution image and video representation report
- T0 + 24 months: Visual object rendering report
- T0 + 30 months: Entropy coding and rate-distortion report

### 5.2.4. Searching and editing

**Description:** A query-by-example interrogation system will be developed in order to show the efficiency of using visual dictionaries for image and video retrieval. The objective will be also to show the benefit of the representation for video editing. The editing task will rely on the use of a visual dictionary that will be used as a library of visual units and groups (taking into account the relationships between visual units).

**Task 4.1: Dictionary-based interrogation system:** Similarity-based image and video retrieval will be developed for content-based query (query-by-example) and multi-resolution browsing applications. Shape and photometric descriptors can be used for content-based image and video retrieval. Using hierarchical indexing schemes, content is straight registered sorted into a dictionary and hierarchical data organization produces a classification tree that holds in itself a topological distance useful for developing similarity-based queries. The consequences due to the topological organization of dictionaries will be studied in depth for content-based image and video retrieval techniques.

**Task 4.2: Editing visual content:** Image and video editing will be experimented by modifying the structure of content, but also by trying to handle shapes and responses at a control-point level on representation of visual primitives. It will afford personal dictionary editing and will allow to create visual libraries that enables editing capitalization and improves editing productivity. Using such an approach will need to revisit object grouping, in-painting and matting techniques that allow deleting, replacing or inserting visual units in an image or a video sequence. In-painting techniques will be based on sparse representations also used for texture clustering and coding.

**Deliverables and due dates:**

- T0 + 18 months: Similarity-based retrieval and multi-resolution browsing report

– T0 + 30 months: Image and video content-based editing report

### 5.2.5. Simplification, integration and evaluation

**Description:** The main objective of this work-package is to demonstrate the improvement (in terms of scalability, functionalities and quality) provided in image and video coding with the techniques defined in the previous work-packages under the framework of a content-based image/video editor in coded domain. It will include string editing operator and dynamically entropy coding/decoding for enabling comparison with present image and video coding standards.

**Task 5.1: Image/video authoring toolbox:** The main meaningful results from previous work-packages are gathered and adapted in such a way that it provides a toolset easy to use and enables to build a demonstrator. A significant effort will be afforded on the simplification of algorithms developed in the previous work-packages. It will potentially include the following functions:

– image/sequence content management;
– visual shape and signal response dictionaries handling;
– scalable dictionary/image/sequence import/export;
– content-based retrieval;
– multi-scale and similarity-based browsing;
– image/sequence studio composer;
– dictionary statistical inference;
– entropy coding data packing.

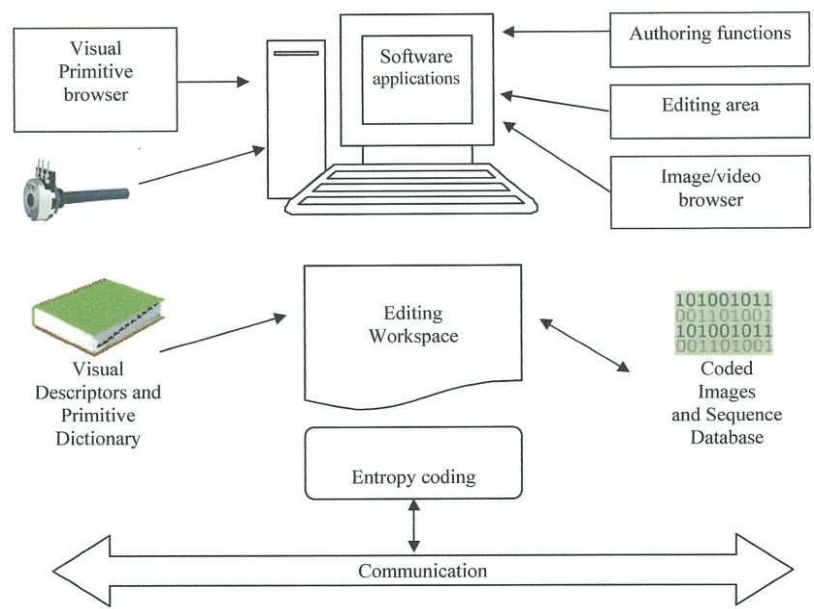

**Figure 49 : Man-machine authoring interface and toolbox integration**

**Task 5.2: Man-machine authoring interface and toolbox integration:** Using the toolbox developed in the previous task, several friendly-to-use software applications are foreseen to test, enhance and demonstrate the potential and the functionalities brought by the project. These applications concern mainly coding, editing, retrieving and visualizing. Though these applications share the same information representation and processing, they differ in the fact that:

- they are not destined for the same users;
- they do not require the same level of interaction;
- they make more or less abstraction;
- they use different methods and parameterizations.

*Coding application*: used by a technician, it will allow to tune any parameter of the coding process and to provide him or her with a direct visual feedback control over some samples before launching the coding process over the whole images or video.

*Editing application:* dedicated to the computer graphist, it will provide authoring functions such as tools to manipulate dictionaries and dictionary elements as well as tools to modify the bit stream to change the content or the visual representation of the image or video.

*Content-base retrieval application:* dedicated to the end user, it will offer an intuitive way to browse images and videos and to query by sample.

*Visualizing application:* again for the end-user, a "player" will allow to render the images or the videos with the best quality according to a small set of parameters. It could be link to the retrieval application since it is dedicated to the same end-user and since viewing images and playing video are the direct consequences of retrieving information.

**Task 5.3: Evaluation of available coding/editing/retrieval functionalities:** Coding/editing/retrieval functionalities are demonstrated and compared to already known techniques. Objective and subjective measures over coded image and video quality are performed taking into account the rate/distortion and a representative set of images and videos.

**Deliverables and due dates:**

- T0 + 24 months: Self-descriptive toolbox development and algorithm simplification report
- T0 + 30 months: Integrated hardware and software demonstrator
- T0 + 36 months: Demonstration evaluation report

## *5.3. Expected results and potential impact*

SDVC is trying to lay the foundations of a solution by proposing an innovative representation for describing visual information allowing defining visual ontologies and managing visual content in a similar manner as textual content. Using such a modeling scheme, SDVC will provide an efficient tool for fast searching mixed (visual and textual) content and would define a similar way for taking in account audio content in the future.

In this scope, the SDVC targets innovative research activities and will advance the state-of-the-art by:

- proposing a new mid-level information description enabling to build more efficient retrieving systems for image and video data;
- taking into account perceptual laws for narrowing the gap between digital objects and perspective views of real world ones;
- managing visual objects independently to their location in the viewing plane and to a given set of geometric transforms;
- proposing an indexing scheme from which visual ontologies can be inferred and allowing to define a kind of visual alphabet for simply connected shapes and some visual dictionary for perceptual groupings;
- and consequently providing a means for narrowing the current semantic gap in visual information.

The project has a strong exploratory perspective and paves the way for future investigations about information content representation and self-descriptive information coding technologies.

The project will also contribute to reinforce the national research by performing leading edge research both foundational and application-oriented in the search engines and computer-vision domains and by bringing together several key players from different scientific and technical disciplines.

## 5.4. Addendum: Generalized images

The research works especially led in the field of synthetic generation have allowed to define more precisely what is an image by listing all the operations that can apply on this kind of information.

When an image is handled, the components that must be part of a generalized model are the following ones

- the support space : 2D, 3D (for some sensors, only accessible in projective geometry), 4D (temporal motion : image sequence) ;
- the functional space : binary, mono-chromatic, multispectral (color, other wave lengths) ;
- description resolution : macroscopic (geometry) microscopic (texture) ;
- environment: illumination conditions, reflection properties (shadow, reflectance analysis).

Texture and reflectance are for instance deeply studied in synthetic image generation, at the opposite very coarsely in pattern recognition, although they may present a huge interest in industrial inspection (surface roughness).

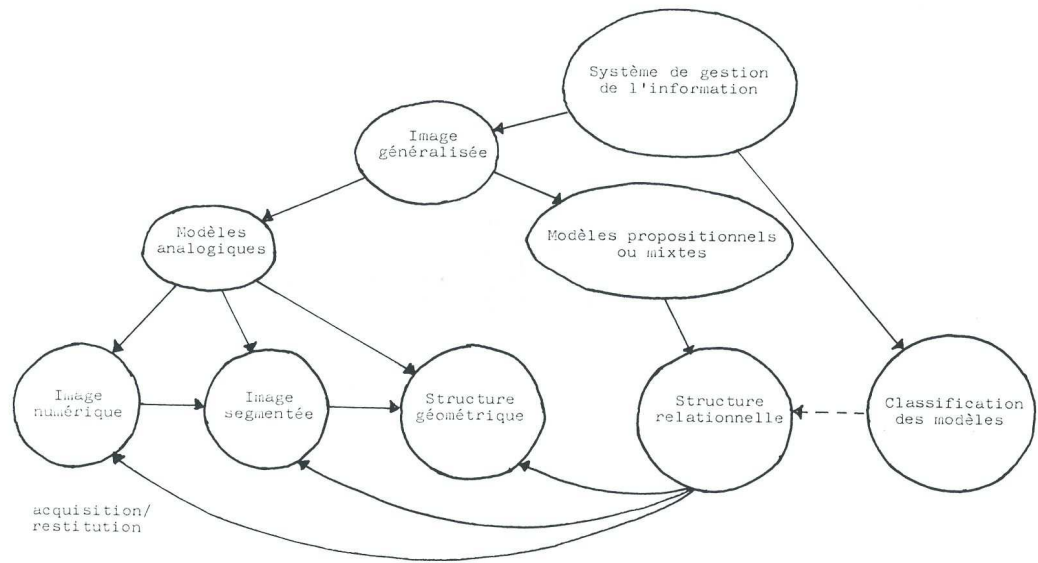

**Figure 50 : Generalized image model**

The taking in account and the handling of the different components of an image by an information processing system lead to draw the figure above, as a model for a generalized image (or intrinsic image for D.H. BALLARD, C.M. BROWN – Computer Vision - Prentice Hall - 1985):

- the digital image corresponds to the description level of a luminescence table;
- the segmented image enables to define homogenous objects (support segmentation into components sharing common properties in the functional space);
- the geometric structure : decomposition of the segmented supports over a base of geometric primitives, description level corresponding to the image graphical information;
- the relational structure : grouping of sub-sets belonging to the original partition of the image by discovering the spatial or similar relations with some already known organization models.

The two intermediate models enable to analyze and recognize printed texts. Several research works are showing that the hand-made writing needs to imply the third one.

To these four levels, it can be added one more taking in account the classification of different representation models for describing an image. It is corresponding to the organized ordering of digital information enabling to access to heterogeneous information by showing an example (for instance showing a drawing for retrieving an image).

Giving that an image may be described on different modeling levels (of complexity), these ones may convey information pieces that a direct analysis might found contradictory. The direct resolving algorithms may fail because of the existence of these different description levels.

Let us for instance focus on the case of recognition of continuous speaking (J. MARIANI-La Reconnaissance de la Parole in "L'Intelligence Artificielle" Numéro spécial La Recherche - Octobre 1985).

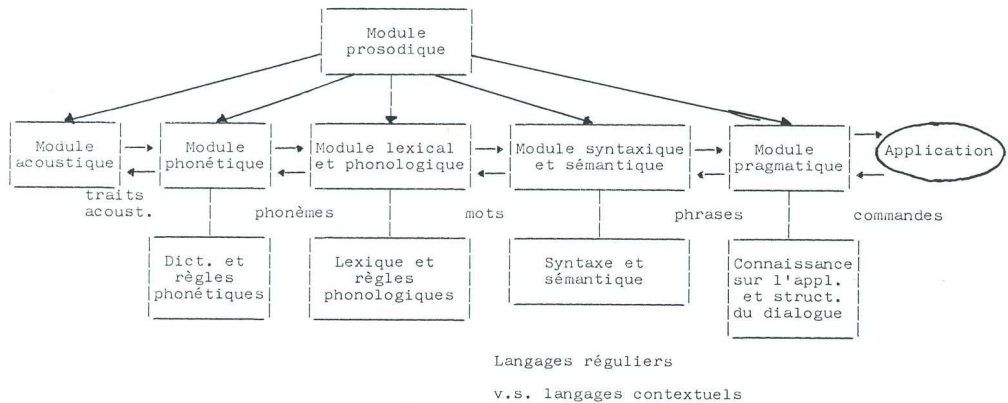

**Figure 51 : Interpretation model in speech recognition**

Several decomposition levels are ruling this one:

- − acoustic level : level where the phonic signal is sampled and from which are extracted les phonic features ;
- − phonetic level : from which are extracted phonemes ;
- − lexical level : where words are composed ;
- − syntactic and semantic level : where sentences are composed ;
- − pragmatic level : level linked to the application, only true in a finite universe defined in advance, without it the previous steps cannot be validated (nowadays, the actual semantic level).

The modules converting the different representation formats are performing alone their task with some success part, but without providing a full solution for the problem.

The full problem solving (for instance, the correct analysis of a question) cannot be led until its end without a parallel progress of each of these modules towards a solution globally satisfying for the sum of all of them.

This approach has been widely adopted for solving complex problems in pattern recognition.

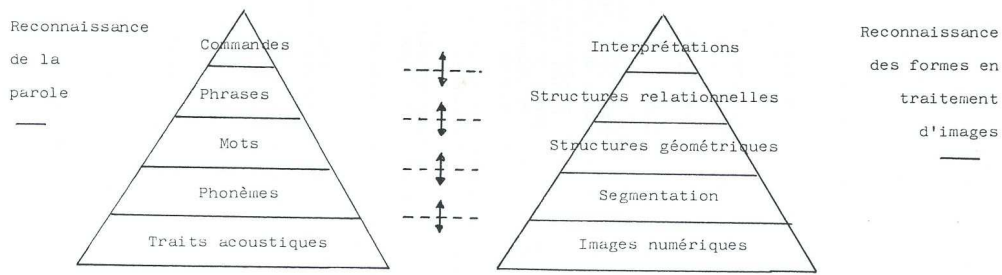

**Figure 52 : Information pyramidal nesting**

It consists in isolating the handling modules at different solving levels and in making them running with a cooperative manner in a network (here a linear one), a module only knowing its nearest neighbors from which it is getting data and to which it is sending its expertise.

The cooperating behavior will then converge towards the most plausible solution to the posed problem (according to the principle that an organized set is stronger than the sum of its isolated parts, in complex system theory).

The solving cooperative mechanisms are implemented over pyramidal information structures as shown in the above figure, where each level can communicate with its closest neighbors.

The decomposition of the continuous speech into acoustic features, phonemes, ..., commands may be reused in the case of generalized images into digital images, segmented images, ..., interpretations.

# Annex: Image analysis algorithms

# 1. Boundary-based image analysis

## 1.1. Image preprocessing

image :          table of number of rows by number of columns of pixels described by a grey level histogram :      table of number of grey levels of values counting the number of pixels in an image having for value the grey level used as index number of jumps :        number of pixel to jump after each processed pixel when computing an histogram in the aim to reduce its computing time (under-sampling) threshold :      grey level used for binarizing an image or retrieving the transitions between the background and the objects in a visual scene background :     grey level defining the value to assign pixels belonging to the background of a visual scene when binarizing an image or generating a frame before looking for transitions in the image

### 1.1.1   Image histogram computation

PROCEDURE image histogram computation
BEGIN
    number of pixels <- number of rows * number of columns
    /* erase the histogram of grey levels */
    FOR grey level = 0 UNTIL (number of grey levels -1)
    DO histogram (grey level) <- 0
    /* histogram computation taking in account only a subset of image pixels */
    FOR pixel number = 0 UNTIL (number of pixels -1) BY STEP OF number of jumps DO
        grey level <- image(pixel number)
        histogram(grey level) <- +1
    END
END

### 1.1.2 Image binarization

PROCEDURE image binarization

BEGIN

    number of pixels <- number of rows * number of columns

    /* image binarization */

    FOR pixel number = 0 UNTIL (number of pixels -1) DO

        IF (image(pixel number) > threshold)

        THEN image(pixel number) <- maximum number of grey levels/2

        ELSE image(pixel number) <- 0

    END

END

### 1.1.3 Generation of a background colored frame

PROCEDURE image frame

BEGIN

    /* processing of the first and the last rows */

    index1 <- 0

    index2 <- (number of rows – 1) * number of columns

    FOR column number = 1 UNTIL number of columns DO

        image(index1) <- background

        image(index2) <- background

        index1 <- +1

        index2 <- +1

    END

    /* processing of the first and the last columns */

    index1 <- 0

    index2 <- number of columns - 1

    FOR row number = 1 UNTIL number of rows DO

        image(index1) <- background

        image(index2) <- background

        index1 <- + number of columns

        index2 <- + number of columns

    END

END

## 1.2.   Border following

root of cycle :   table of maximum number of cycles  length which elements are holding the
                number of the first transition of a cycle

root of row trans. :   table of number  of rows length which elements are holding the number
                of the first transition in a pixel row

root of col. trans. :   table of number of columns length which elements are holding the
                number of the first transition in a pixel column

succ. of row trans. :   table of maximum number of transitions length which elements are
                holding the number of the following transition of the current one in a pixel row

succ. of col. trans. :   table of maximum number of transitions length which elements are
                holding the number of the following transition of the current one in a pixel
                column

succ. of transition :   table made from the union of the two previous tables

way of row trans. :   table associated to the row transitions saving if a transition is rising
                (background-to-object) or descending(object–to-background)

way of col. trans. :   table associated to the column transitions saving if a transition is rising
                (background-to-object) or descending(object-to-background)

way of transition :   table made from the union of the two previous tables

abs. of row trans. :   table associated to the row transitions saving the abscissas of the row
                transitions

abs. of col. trans. :   table associated to the column transitions saving the abscissas of the
                column transitions

abs. of transition :   table made from the union of the two previous tables

ord. of row trans. :   table associated to the row transitions saving the ordinates of the row
                transitions

ord. of col. trans.:   table associated to the column transitions saving the ordinates of the
                column transitions

ord. of transition :   table made from the union of the two previous tables

succ. of cycle trans.:   table associated to all the transitions indicating the number of the
                following transition when traversing a cycle

cycle num. of trans.:   table associated to all the transitions indicating the cycle number to
                which is belonging the current transition

### 1.2.1  Looking for row transitions

PROCEDURE stacking row transitions

BEGIN

    transition number <- 0

    row number <- 0

    /* scanning image rows */

    WHILE (row number < (number of rows – 1)) DO

        /* scanning columns in a row */

        FOR column number = 0 UNTIL (number of columns – 1) DO

            IF (((image(row number, column number) – threshold) *

            (image(row number, column number + 1) – threshold)) < 0) THEN DO

                /* processing a row transition */

                IF (background = DARK)

                THEN way of transition <- image(row number, column number + 1)

                – image(row number, column number )

                ELSE way of transition <- image(row number, column number)

                – image(row number, column number + 1)

                transition number <- +1

                way of row trans.(transition number) <- way of transition

                ord. of row trans.(transition number) <- row number

                IF (way of transition > 0)

                THEN abs. of row trans.(transition number) <- column number

                ELSE abs. of row trans.(transition number) <- column number + 1

        END

        row number <- + 1

    END

    number of row transitions <- transition number

END

PROCEDURE generation of row transition links

BEGIN

    IF (number of row transitions > 0) THEN DO

        /* initialization of list pointers */

        cycle kind <- NULL

        FOR row number = 0 UNTIL (number of rows – 1) DO

            root of row trans.(row number) <- NULL

        END

        abs. of row trans.(NULL) <- 0

        ord. of row trans.(NULL) <- 0

        succ. of row trans.(NULL) <- NULL

        succ. of cycle trans.(NULL) <- NULL

        FOR transition number = 1 UNTIL number of row transitions DO

            succ. of row trans.(transition number) <- NULL

            succ. of cycle trans.(transition number) <- NULL

            object indicator(transition number) <- NULL

        END

        /* generation of transition upwards links */

        FOR transition number = 1 UNTIL number of row transitions DO

            row number <- ord. of row trans.(transition number))

            IF (root of row trans.(row number) = NULL)

            THEN root of row trans.(row number) <- transition number

            ELSE succ. of row trans.(predecessor) <- transition number

            predecessor <- transition number

        END

    END

END

## 1.2.2 Looking for column transitions

PROCEDURE stacking column transitions

BEGIN

    transition number <- 0

    column number <- 0

    /* scanning image columns */

    WHILE (column number < (number of columns – 1)) DO

        /* scanning rows in a column */

        FOR row number = 0 UNTIL (number of rows – 1) DO

            IF (((image(row number, column number) – threshold) *

            (image(row number + 1, column number) – threshold)) < 0) THEN DO

                /* processing a column transition */

                IF (background = DARK)

                THEN way of transition <- image(row number + 1, column number)

                – image(row number, column number )

                ELSE way of transition <- image(row number, column number)

                – image(row number + 1, column number)

                transition number <- +1

                way of col. trans.(transition number) <- way of transition

                abs. of col. trans.(transition number) <- column number

                IF (way of transition > 0)

                THEN ord. of col. trans.(transition number) <- row number

                ELSE ord. of col. trans.(transition number) <- row number + 1

        END

        column number <- + 1

    END

    number of col. transitions <- transition number

END

PROCEDURE generation of column transition links

BEGIN

    IF (number of col. transitions > 0) THEN DO

        /* initialization of list pointers */

        cycle kind <- NULL

        FOR column number = 0 UNTIL (number of columns – 1) DO

            root of col. trans.(column number) <- NULL

        END

        abs. of col. trans.(NULL) <- 0

        ord. of col. trans.(NULL) <- 0

        succ. of col. trans.(NULL) <- NULL

        succ. of cycle trans.(NULL) <- NULL

        FOR transition number = 1 UNTIL number of col. transitions DO

            succ. of col. trans.(transition number) <- NULL

            succ. of cycle trans.(transition number) <- NULL

            object indicator(transition number) <- NULL

        END

        /* generation of transition upwards links */

        FOR transition number = 1 UNTIL number of row transitions DO

            column number <- abs. of col. trans.(transition number))

            IF (root of col. trans.(column number) = NULL)

            THEN root of col. trans.(column number) <- transition number

            ELSE succ. of col. trans.(predecessor) <- transition number

            predecessor <- transition number

        END

    END

END

### 1.2.3 Contour generation from transition lists

PROCEDURE generation of object contours

BEGIN

    IF ((number of row transitions > 0) AND (number of col. transitions > 0)) THEN DO

        /* initialization of structures */

        FOR cycle number = 0 UNTIL maximum number of cycles DO

            root of cycle(cycle number) <- NULL

            object(cycle number) <- FALSE

        END

        /* looking for cycle beginnings in row transitions */

        cycle number <- 0

        transition number <- 0

        WHILE ((cycle number < maximum number of cycles)

        AND (transition number < number of row transitions)) DO

            transition number <- +1

          IF (succ of cycle trans.(transition number) = NULL) THEN DO

              /* a cycle beginning has been found, it is then analyzed */

              cycle number <- +1

              root of cycle(cycle number) <- transition number

              CALL cycle generation(transition number, cycle number)

              object(cycle number) <- TRUE

        END

    END

    number of cycles <- cycle number

END

PROCEDURE cycle generation (transition number, cycle number)

BEGIN

   cycle end <- FALSE

   row scrutinizing <- TRUE

   pointer <- transition number

   /* following the object border */

   WHILE ((NOT cycle end) AND (succ. of cycle trans.(pointer) = NULL)) DO

      transition number <- pointer

      cycle end <- TRUE

      /* central point examination */

      IF (row scrutinizing) THEN DO

        /* looking for a neighbor on the column of same abscissa */

        abscissa <- abs. of transition(transition number)

        ordinate <- ord. of transition(transition number)

        visiting way <- (way of transition(transition number))

        CALL scrutinizing column transitions

      END

      ELSE DO

        /* looking for a neighbor on the row of same ordinate */

        abscissa <- abs. of transition(transition number)

        ordinate <- ord. of transition(transition number)

        visiting way <- (- way of transition(transition number))

        CALL scrutinizing row transitions

      END

      /* examination of 4-connected neighbors */

      IF (cycle end) THEN DO

        IF (row scrutinizing) THEN DO

            IF (way of transition(transition number) >= 0) THEN DO

                /* looking for a neighbor on the above row */

                abscissa <- abs. of transition(transition number)

```
                ordinate <- ord. of transition(transition number) - 1

                visiting way <- POSITIVE

                CALL scrutinizing row transitions

            END

            ELSE DO

                /* looking for a neighbor on the underneath row */

                abscissa <- abs. of transition(transition number)

                ordinate <- ord. of transition(transition number) + 1

                visiting way <- NEGATIVE

                CALL scrutinizing row transitions

            END

        END

        ELSE DO

            IF (way of transition(transition number) >= 0) THEN DO

                /* looking for a neighbor on the right column */

                abscissa <- abs. of transition(transition number) + 1

                ordinate <- ord. of transition(transition number)

                visiting way <- POSITIVE

                CALL scrutinizing column transitions

            END

            ELSE DO

                /* looking for a neighbor on the left column */

                abscissa <- abs. of transition(transition number) - 1

                ordinate <- ord. of transition(transition number)

                visiting way <- NEGATIVE
```

```
        CALL scrutinizing column transitions

    END

  END

END

/* examination of 8-connected neighbors */

IF (cycle end) THEN DO

  IF (row scrutinizing) THEN DO

    IF (way of transition(transition number) >= 0) THEN DO

        /* looking for a neighbor upwards on left side */

        abscissa <- abs. of transition(transition number) - 1

        ordinate <- ord. of transition(transition number) - 1

        visiting way <- NEGATIVE

        CALL scrutinizing column transitions

    END

    ELSE DO

        /* looking for a neighbor downwards on right side */

        abscissa <- abs. of transition(transition number) + 1

        ordinate <- ord. of transition(transition number) + 1

        visiting way <- POSITIVE

        CALL scrutinizing column transitions

    END

  END

  ELSE DO

    IF (way of transition(transition number) >= 0) THEN DO

        /* looking for a neighbor upwards on right side */

        abscissa <- abs. of transition(transition number) + 1
```

```
            ordinate <- ord. of transition(transition number) - 1

            visiting way <- POSITIVE

            CALL scrutinizing row transitions

        END

        ELSE DO

            /* looking for a neighbor downwards on left side */

            abscissa <- abs. of transition(transition number) - 1

            ordinate <- ord. of transition(transition number) + 1

            visiting way<- NEGATIVE

            CALL scrutinizing row transitions

        END

      END

     END

    END

   END
```

PROCEDURE scrutinizing column transitions

BEGIN

    /* looking for a neighbor on a column */

    pointer <- root of col. trans. (abscissa)

    IF (way of col. trans. (pointer) <> visiting way)

    THEN pointer <- succ. of col. trans.(pointer)

    WHILE ((ord. of col. trans. (pointer) < ordinate) AND (pointer <> NULL)) DO

        pointer <- succ. of col. trans. (pointer)

        pointer <- succ. of col. trans. (pointer)

    END

    IF ((ord. of col. trans. (pointer) = ordinate) AND (pointer <> NULL)) THEN DO

        cycle end <- FALSE

        succ of cycle trans.(transition number) <- pointer

        cycle num. of trans.(transition number) <- cycle number

    END

END

PROCEDURE scrutinizing row transitions

BEGIN

    /* looking for a neighbor on a row */

    pointer <- root of row trans. (abscissa)

    IF (way of row trans.(pointer) <> visiting way)

    THEN pointer <- succ. of row trans.(pointer)

    WHILE ((abs. of row trans.(pointer) < abscissa) AND (pointer <> NULL)) DO

        pointer <- succ. of row trans.(pointer)

        pointer <- succ. of row trans.(pointer)

    END

    IF ((abs. of row trans.(pointer) = abscissa) AND (pointer <> NULL)) THEN DO

        cycle end <- FALSE

        succ of cycle trans.(transition number) <- pointer

        cycle num. of trans.(transition number) <- cycle number

    END

END

## *1.3.    Computation of object characteristics*

object :          table associated to the cycles indicating if a cycle is closed or not (contour of an object) and in which other cycle it may be included perimeter :       table associated to the cycles storing the perimeter lengths of closed cycles surface area :    table associated to the cycles storing the surface areas of closed cycles abs. of gravity center : table associated to the cycles storing the abscissas of the gravity center of validated cycles ord. of gravity center : table associated to the cycles storing the ordinates of the gravity center of validated cycles

### 1.3.1   Computation of the object perimeters

PROCEDURE computation of the object perimeters

BEGIN

    FOR cycle number = 1 UNTIL number of cycles DO

       object perimeter(cycle number) <- 0

       IF (object(cycle number) <> FALSE) THEN DO

          predecessor <- root of cycle(cycle number)

          successor <- succ of cycle trans.(predecessor)

          DO

             object perimeter(cycle number) <- +1

             predecessor <- successor

             successor <- succ of cycle trans.(predecessor)

         UNTIL THAT (predecessor = root of cycle(cycle number))

      END

    END

END

### 1.3.2 Computation of the mass, the gravity center and the inclusion relation of objects between themselves

PROCEDURE computation of the first object moments

BEGIN

  IF (number of cycles > 0) THEN DO

     /* initialization of the pre-computed power series */

     integer sum(0) <- 0

     FOR abscissa = 1 UNTIL number of columns DO

       integer sum(abscissa) <- integer sum(abscissa – 1) + abscissa

     END

     /* initialization of object attributes */

     FOR cycle number = 1 UNTIL number of cycles DO

       IF (object(cycle number)) THEN DO

         object surface area(cycle number) <- 0

         abs. of gravity center(cycle number) <- 0

         ord. of gravity center(cycle number) <- 0

       END

     END

     /* computation of object attributes */

     FOR row number = 0 UNTIL (number of rows – 1) DO

      predecessor <- root of row trans.(row number)

      cycle number <- cycle number(predecessor)

      WHILE (predecessor <> NULL) DO

        IF ((object(cycle number) AND (way of transition(rac.de cycle(cycle number)) = way of transition(predecessor))) THEN DO

          /* looking for the segment end belonging to the object contour */

```
successor <- succ. of row trans.(predecessor)

WHILE ((cycle number((successor) <> cycle number)

AND (successor <> NULL)) DO

    /* marking encountered objects included in the object being analyzed */

    IF (object(cycle number(successor)) <> FALSE)

    THEN object(cycle number(successor)) <- cycle number

    successor <- succ. of row trans.(successor)

    successor <- succ. of row trans.(successor)

END

IF (successor <> NULL) THEN DO

    /* sum up object attributes */

    abscissa <- abs. of row trans.(predecessor)

    length <- abs. of row trans.(successor)

    - abs. of row trans.(predecessor) + 1

    object surface area(cycle number) <- +length

    contribution <- length * abscissa + sum(length – 1)

    abs. of gravity center(cycle number) <- +contribution

    contribution <- row number * length

    ord. of gravity center(cycle number) <- +contribution

END

END

predecessor <- succ. of row trans.(predecessor)

cycle number <- cycle number(predecessor)

END

END
```

```
/* invalidation of zero-valued masses */
FOR cycle number = 1 UNTIL number of cycles DO
  IF (object(cycle number)) THEN DO
      IF (object surface area(cycle number) = 0)
      THEN object(cycle number)) <- FALSE
    END
  END
 END
END
```

## 1.4.   Object convex hulls

convex trans. number :          number of row transitions which convexity is plausible

way of convex trans. :          table associated to the convex transitions indicating if the
                                transition is rising (background-to-object) or descending (object-to-background)

abs. of convex trans. :  table associated to the convex transitions storing the abscissas of the row
                                transitions

ord. of convex trans. :  table associated to the convex transitions storing the ordinates of the row
                                transitions

convexity of  transition:          table associated to the convex transitions indicating if the
                                transition is truly convex or not

### 1.4.1   Looking for lower and higher row transitions

PROCEDURE search for lower and higher row transitions

BEGIN

    transition number <- 0

    /* scrutinizing row transitions */

    FOR row number = 1 UNTIL number of rows DO

        pointer <- root of row trans. (row number)

        minimum <- TRUE

        WHILE (pointer <> NULL) DO

          IF (object(cycle number(pointer))) THEN DO

              IF (minimum) THEN DO

                  /* lower transition to be stored */

                  transition number <- +1

                  way of convex trans.(transition number) <- way of transition(transition number)

                  ord. of convex trans.(transition number) <- ord. of transition(transition number)

                  abs. of convex trans.(transition number) <- abs. of transition(transition number)

                  transition number <- +1

minimum <- FALSE

END

ELSE DO

/* higher transition to be stored */

way of convex trans.(transition number) <- way of transition(transition number)

ord. of convex trans.(transition number) <- ord. of transition.(transition number)

abs. of convex trans.(transition number) <- abs. of transition.(transition number)

END

END

pointer <- succ. of row trans.(pointer)

END

END

number of convex trans. <- transition number

END

## 1.4.2  Identification of object convex vertices

PROCEDURE search for convex vertices

BEGIN

/* initialization of convexity marks */

FOR transition number = 1 UNTIL number of convex trans. DO

convexity of transition(transition number) <- TRUE

END

/* minima analysis */

FOR transition number = 3 UNTIL (number of convex trans. – 3) BY STEP OF 2 DO

FOR index1 = 1 UNTIL (transition number – 2) BY STEP OF 2 DO

IF (convexity of transition(transition number)) THEN DO

slope1 <-

(abs. of convex trans.(transition number) – abs. of convex trans.(index1))

/ (ord. of convex trans.(transition number) – ord. of convex trans.(index1))

FOR index2 = (transition number + 2) UNTIL (number of convex trans. – 1) BY STEP OF 2 DO

    slope2 <-

    (abs. of convex trans.(transition number) – abs. of convex trans.(index2))

    / (ord. of convex trans.(index2)– ord. of convex trans.(transition number))

    difference of slopes <- slope1 – slope2

    IF (difference of slopes < 0) THEN DO

        convexity of transition(transition number) <- FALSE

        index1 <- transition number

        index2 <- number of convex trans.

    END

  END

END

END

END

/* maxima analysis */

FOR transition numbers = 4 UNTIL (number of convex trans. – 2) BY STEP OF 2 DO

  FOR index1 = 2 UNTIL (number of transitions– 2) BY STEP OF 2 DO

   IF (convexity of transition(transition number)) THEN DO

    slope1 <-

    (abs. of convex trans.(transition number) – abs. of convex trans.(index1))

    / (ord. of convex trans.(transition number) – ord. of convex trans.(index1))

    FOR index2 = (transition number + 2) UNTIL number of convex trans. BY STEP OF 2 DO

      slope2 <-

(abs. of convex trans.(transition number) – abs. of convex trans.(index2))

/ (ord. of convex trans.( index2) – ord. of convex trans.( transition number)))

difference of slopes <- slope1 – slope2

IF (difference of slopes > 0) THEN DO

    convexity of transition(transition number) <- FALSE

    index1 <- transition number

    index2 <- number of convex trans.

    END

   END

  END

  END

 END

END

# 2. Region-based image analysis

## *2.1.    Median transformations*

### 2.1.1    Median filtering of a pseudo-color image

PROCEDURE median filtering of a pseudo-color image

BEGIN

    FOR each image point DO

        erase color counters

        FOR each neighbor point DO

           IF (color(neighbor) <> NOT ASSIGNED)

               THEN increment count(color(neighbor))

        END

        search for the color of higher count

        IF (count(maximum color) > count(color (point)))

           THEN color(point) <- maximum color

    END

END

## 2.1.2  Extension of a pseudo-color image

PROCEDURE extension of a pseudo-color image

BEGIN

    number of modified points <- expansion of a pseudo-color image

    WHILE (number of modified points > 0)

        number of modified points <- expansion of a pseudo-color image

END

FUNCTION expansion of a pseudo-color image

BEGIN

    number of modified points <- 0

    FOR each image point DO

        IF (color(point) = NOT ASSIGNED) THEN DO

          erase color counters

          FOR each neighbor point DO

            IF (color(neighbor) <> NOT ASSIGNED)

               THEN increment count(color(neighbor))

          END

          search for the color of higher count

          color(point) <- maximum color

          IF (color(point) <> NOT ASSIGNED)

            THEN number of modified points <- +1

        END

    END

    RETURN (number of modified points)

  END

## 2.2. Image segmentation

### 2.2.1 Multicolor image segmentation

PROCEDURE pseudo-color image segmentation

BEGIN

    FOR each point (X,Y) from the image DO

      /* determination of the current point color */

      IF (pseudo-color image (X,Y) = pseudo-color image (X-1,Y)

         THEN image of blobs (X,Y) <- image of blobs (X-1,Y)

         ELSE IF (pseudo-color image (X,Y) = pseudo-color image (X,Y-1))

            THEN image of blobs (X,Y) <- image of blobs (X,Y-1)

            ELSE image of blobs (X,Y) <- new blob

      /* merge of connected blobs */

      IF (pseudo-color image (X,Y) = pseudo-color image (X-1,Y) = pseudo-color image (X,Y-1))

         THEN make equivalent blobs (image of blobs (X-1,Y),  image of blobs (X-1,Y))

    END

END

Notice : It is corresponding to a 4-connectivity segmentation ; the 8-connectivity version can be obtained by adding the complementary following test for the determination of the current point color:

IF (pseudo-color image (X,Y) = pseudo-color image (X-1,Y-1))

   THEN image of blobs (X,Y) <- image of blobs (X-1,Y-1)

   ELSE...

## 2.2.2 Complements to segmentation

FUNCTION new blob

BEGIN

    number of blobs <- + 1

    equivalent blob (number of blobs) <- number of blobs

    RETURN (number of blobs)

END

PROCEDURE make equivalent blobs (first blob, second blob)

BEGIN

    blob number <- first blob

    WHILE (equivalent blob (blob number) <> blob number))

       DO blob number <- equivalent blob (blob number)

    equivalent blob (second blob) <- blob number

END

### 2.2.3 Segmentation post-processing

PROCEDURE update of the equivalent blob table

BEGIN

    FOR blob number = 1 UNTIL number of blobs

        DO make equivalent blobs (blob number, blob number)

END

PROCEDURE re-coloring the blob image

BEGIN

    FOR each point (X,Y) of the blob image

        DO blob image (X,Y) <- equivalent blob (blob image (X,Y))

END

## 2.3.  Binary image operations

### 2.3.1  Splitting of a pseudo-color image into binary images

PROCEDURE splitting of a pseudo-color image into binary images

BEGIN

    FOR each pseudo-color image point DO

        FOR each pseudo-color DO

            IF (pseudo-color image = pseudo-color)

            THEN binary image (pseudo-color) <- BLACK

            ELSE binary image (pseudo-color) <- WHITE

        END

    END

END

### 2.3.2 Erosion of a binary image

PROCEDURE erosion of a binary image

BEGIN

    FOR each image point DO

        IF (color(point) = BLACK) THEN DO

            erase color counters

            FOR each point neighbor

                DO increment count(color(neighbor))

            IF (count(WHITE) > count (BLACK))

                THEN color(point) <- WHITE

        END

    END

END

### 2.3.3 Dilatation of a binary image

PROCEDURE dilatation of a binary image

BEGIN

    FOR each image point DO

        IF (color(point) = WHITE) THEN DO

            erase color counters

            FOR each point neighbor

                DO increment count(color(neighbor))

            IF (count(BLACK) > count(WHITE))

                THEN color(point) <- BLACK

        END

    END

END

### 2.3.4 Thinning of a binary image

PROCEDURE thinning of a binary image

BEGIN

 /* delete non median points, west- and south-connected to background */

 FOR each image point DO

    IF (color(point) = BLACK) THEN

      IF ((color(west neighbor) = WHITE) OR (color(south neighbor) = WHITE))

     THEN

        /* non median point deletion */

        IF ((NOT (color(west neighbor) = color(east neighbor) = WHITE)))

        AND (NOT (color(north neighbor) = color(south neighbor) = WHITE))

         THEN color (point) <- WHITE

 END

 /* delete non median points, east- and north-connected to background */

 FOR each image point DO

    IF (color(point) = BLACK) THEN

   IF ((color(east neighbor) = WHITE) OR (color(north neighbor) = WHITE))

    THEN

        /* non median point deletion */

        IF ((NOT (color(west neighbor) = color(east neighbor) = WHITE)))

        AND (NOT (color(north neighbor) = color(south neighbor) = WHITE))

         THEN color (point) <- WHITE

  END

END

## 2.4. Image composition

### 2.4.1 Composition of the thinned blob image

PROCEDURE composition of the thinned blob image

BEGIN

    FOR each point of the image of blobs DO

        assignment <- FALSE

        FOR each pseudo-color

            DO IF (binary image (pseudo-color) = BLACK)

                THEN assignment <- TRUE

        IF (NOT assignment) THEN blob (point) <- NOT ASSIGNED

    END

END

### 2.4.2 Computation of the pseudo-color image of blobs of a given kind

PROCEDURE computation of the pseudo-color image of blobs of a given kind

BEGIN

    FOR each point of the image of blobs DO

        IF (table of dimensions (blob) <> selected dimension)

            THEN pseudo-color image <- NOT ASSIGNED

    END

END

### 2.4.3 Computation of the pseudo-color image of blobs of all kinds except one

PROCEDURE computation of the pseudo-color image of blobs of all kinds except one

BEGIN

    FOR each point of the image of blobs DO

        IF (table of dimensions (blob) = selected dimension)

            THEN pseudo-color image <- NOT ASSIGNED

    END

END

### 2.4.4 Union of blob images of the three types: points, lines and polygons

PROCEDURE union of blob images of the three types: points, lines and polygons

BEGIN

    FOR each point of the image of blobs DO

        IF (image of points (point) <> NOT ASSIGNED)

            THEN image of blobs (point) <- image of points (point)

            ELSE IF (image of lines (point) <> NOT ASSIGNED)

                THEN image of blobs (point) <- image of lines (point)

                ELSE image of blobs (point) <- image of polygons (point)

    END

END

## 2.5.  Attribute calculus

### 2.5.1  Computation of the blob intrinsic dimensions

PROCEDURE computation of the blob intrinsic dimensions

BEGIN

    FOR each point of the image of blobs DO

        IF (blob (point) <> NOT ASSIGNED) THEN DO

          /* evaluation of the point dimension */

          dimension <- 2

          IF ((blob (west neighbor) <> blob (point))) AND (blob (east neighbor) <> blob (point))

            THEN dimension <- - 1

          IF ((blob (north neighbor) <> blob (point))) AND (blob (south neighbor) <> blob (point))

            THEN dimension <- - 1

          /* update of the table of blob dimensions */

          IF (table of dimensions (blob) < dimension))

            THEN table of dimensions (blob) <- dimension

    END

END

### 2.5.2 Computation of the blob bounding boxes

PROCEDURE computation of the blob bounding boxes

BEGIN

    FOR each point (X, Y) of the image of blobs DO

        IF (blob (point) <> NOT ASSIGNED) THEN DO

            IF ( Xmin (blob (point)) > X) THEN Xmin (blob (point)) <- X

            IF ( Xmax (blob (point)) < X) THEN Xmax (blob (point)) <- X

            IF ( Ymin (blob (point)) > Y) THEN Ymin (blob (point)) <- Y

            IF ( Ymax (blob (point)) < Y) THEN Ymax (blob (point)) <- Y

        END

    END

END

### 2.5.3 Computation of the blob perimeters

PROCEDURE computation of the blob perimeters

BEGIN

    FOR each point of the image of blobs DO

    IF (blob (point) <> NOT ASSIGNED) THEN DO

        /* determination if the point is ont the blob boundary */

        boundary <- FALSE

        FOR each 8-connected neighbor of the point DO

            IF ((blob (neighbor) <> NOT ASSIGNED)) AND (blob (neighbor) <> blob (point))

                THEN boundary <- TRUE

        /* sum up to the blob perimeter */

        IF (boundary) THEN table of perimeters (blob (point)) <-+1

        END

    END

END

PROCEDURE computation of the blob moments

BEGIN

    FOR each point (X, Y) of the image of blobs DO

        IF (blob (point) <> NOT ASSIGNED) THEN DO

| | |
|---|---|
| m (blob (point)) | <-- + 1 |
| mx (blob (point)) | <-- + X |
| my (blob (point)) | <-- + Y |
| mxx (blob (point)) | <-- + $X^2$ |
| mxy (blob (point)) | <-- + XY |
| myy (blob (point)) | <-- + $Y^2$ |
| mxxx (blob (point)) | <-- + $X^3$ |
| mxxy (blob (point)) | <-- + $X^2Y$ |
| mxyy (blob (point)) | <-- + $XY^2$ |
| myyy (blob (point)) | <-- + $Y^3$ |

      END

    END

END

Notice : it is necessary to complete this computation with moment centering (after gravity center computation) and with moment normalization (after inertia axes and rotation angle computation).

## 2.6.  Computation of the inclusion relation

### 2.6.1  Computation of the inclusion relation between blobs

PROCEDURE computation of the inclusion relation between blobs

BEGIN

    FOR each image row DO

        previous blob <- NOT ASSIGNED

        stack initialization

        FOR each row point DO

            blob <- image (point)

            IF (blob <> previous blob) THEN DO

                IF (stack (stack top - 1) = blob) THEN DO

                    unstack

                    IF (father blob (previous blob) <> blob))

                        THEN assign (previous blob, blob)

                END

                ELSE stack up (blob)

                previous blob <- blob

            END

        END

    END

END

### 2.6.2 Complement to the computation of the inclusion relation

PROCEDURE assign (previous blob, blob)

BEGIN

    IF (father blob (previous blob) = NOT ASSIGNED)

        THEN father blob (previous blob) <- blob

        ELSE IF (father blob (father blob (previous blob)) = blob)

           THEN father blob (previous blob) <- blob

END

Notice : the second test takes in account the case of adjacent blobs but not necessarily included and serves to correct a first wrong assignment by re-assigning the same father.

## 2.7. Blob vectorization

precision :    accepted maximum error

abs :    contour point abscissa

ord :    contour point ordinate

indicator :    contour point indicator specifying if it is a vector vertex

origin :    first point of a contour point list linked to a tree node

end :    last point of a contour point list linked to a tree node

lftson :    left son node in a vector tree

rgtson :    right son node in a vector tree

lftmax :    the most left point in a vector list

rgtmax :    the most right point in a vector list

lftwdth :    width of the bounding box of the left son

rgtwdth :    width of the bounding box of the right son

### 2.7.1  Computation of the blob boundaries

PROCEDURE computation of the blob boundaries

BEGIN

    FOR each image point of blobs DO

        IF (blob (point) <> NOT ASSIGNED) THEN DO

            /* determination if the point is on the boundary */

            boundary <- FALSE

            FOR each 4-connected neighbor of the point DO

                IF ((blob (neighbor) <> NOT ASSIGNED)) AND (blob (neighbor) <> blob (point)))

                    THEN boundary <- TRUE

            /* erase interior points of the blob */

            IF (NOT boundary) THEN blob (point) <- NOT ASSIGNED

        END

    END

END

## 2.7.2 Contour digitalization

PROCEDURE contour digitalization (contour number)

BEGIN

/* looking for the first point of the contour */

X <- Ymin (blob number)

Y <- Xmin (blob number)

WHILE (blob (point(X,Y)) <> blob number) DO X <- + 1

/* border following and contour point digitalization */

number of points <- 0

store the point (X,Y)

WHILE (there is a 4-connected point with the same blob number)

    DO store the point (X,Y)

END

PROCEDURE store the point (X,Y)

BEGIN

number of points <- +1

abscissa (number of points) <- X

ordinate (number of points) <- Y

indicator (number of points) <- NOT ASSIGNED

blob (point(X,Y)) <- NOT ASSIGNED

END

### 2.7.3 Vectorization of a list of points

PROCEDURE vectorization of a list of points (precision)

BEGIN

    initialization of the vector tree

    node <- 1

    /* building the vector tree */

    WHILE (((lftwdth (node) + rgtwdth (node)) > precision) OR  (node <> 1)) DO

        IF ((lftwdth (node) + rgtwdth (node)) > precision) THEN DO

            IF ((lftson (node)=0) AND (rgtson (node)=0)) THEN DO

                /* node division and descent on left side */

                divide a vector in two halves (node)

                compute the vector widths (lftson (node))

                node <- lftson (node)

            END

            ELSE DO

                IF ((lftwdth (rgtson (node)) + rgtwdth (rgtson (node))) > precision) THEN DO

                    /* descent on right side */

                    compute the vector widths (rgtson(node))

                    node <- rgtson (node)

                END

                ELSE DO

                  /* ascent by right side, width update */

                  node <- father (node)

                  lftwdth (node) <- MAX(lftwdth (lftson(node)) + rgtwdth (lftson (node),

                    lftwdth (rgtson (nœud) + rgtwdth (rgtson (node)))

                  rgtwdth (node) <- 0

                END

            END

```
            END

        ELSE node <- father (node)

    END

END

PROCEDURE initialization of the vector tree

BEGIN

    nbnode <- 1

    father (nbnode) <- 0

    lftson (nbnode) <- 0

    rgtson (nbnode) <- 0

    origin (nbnode) <- 1

    end (nbnode) <- number of points

    indicator (origin (nbnode)) <- VERTEX

    indicator (end (nbnode)) <- VERTEX

    lftwdth (nbnode) <- precision

    rgtwdth (nbnode) <- precision

END
```

PROCEDURE divide a vector in two halves (node)

BEGIN

    nbnode <- + 2

    father (nbnode – 1) <- current node

    father (nbnode) <- current node

    lftson (nœud) <- nbnode – 1

    rgtson (nœud) <- nbnode

    lftson (nbnode – 1) <- 0

    rgtson (nbnode – 1) <- 0

    lftson (nbnode) <- 0

    rgtson (nbnode) <- 0

    lftwdth (nbnode – 1) <- precision

    rgtwdth (nbnode – 1) <- precision

    lftwdth (nbnode) <- precision

    rgtwdth (nbnode) <- precision

    origin (nbnode – 1) <- origin (node)

    end (nbnode) <- end (node)

    /* selection of the most distant vector point as a dividing point */

    IF (lftwdth(node) > rgtwdth (node)) THEN DO

        end (nbnode - 1) <- lftmax (node)

        origin (nbnode) <- lftmax (node)

        indicator (lftmax(node) <- VERTEX

    END

    ELSE DO

        end (nbnode - 1) <- rgtmax (node)

        origin (nbnode) <- rgtmax (node)

        indicator (rgtmax(node)) <- VERTEX

    END

END

PROCEDURE compute the vector widths (node)

BEGIN

   /* initialization of computing parameters */

  left width <- 0

  right width <- 0

  left maximum <- origin (node)

  right maximum <- origin (node)

  compute the normalized coefficients of the straight line (a, b, c)

  /* computation of the distance of each point to the vector */

  point number <- origin (node)

  WHILE (point number < end (node)) DO

     d <- a * abs (point number) + b * ord (point number) + c

    IF (d < 0) THEN

      /* point on the left side of the vector */

      IF (d < left width) THEN DO

        left width <- - d

        left maximum <- point number

      END

    ELSE

      /* point on the left side of the vector */

      IF (d > right width) THEN DO

        right width <- d

        right maximum <- point number

      END

    /* point move */

    point number <- + 1

  END

  lftwdth (node) <- left width

  rgtwdth (node) <-  right width

lftmax (node) <- left maximum

rgtmax (node) <- right maximum

END

PROCEDURE compute the normalized coefficients of the straight line (a, b, c)

BEGIN

a <- ord (end (nbnode)) – ord (origin (nbnode))

b <- abs (origin (nbnode)) – abs (end (nbnode))

c <- abs (end (nbnode)) * ord (origin (nbnode)) - abs (origin (nbnode)) * ord (end ( nbnode))

d <- SQRT ( a * a + b * b )

a <- a/d

b <- b/d

c <- c/d

END

### 2.7.4 Digitalization of a vector

PROCEDURE digitalization of a vector (node)

BEGIN

   /* initialization */

   u <- |abs (end (node)) – abs (origin (node))|

   v <- |ord (end (node)) – ord (origin (node))|

   IF (abs (end (node)) > abs (origin (node))) THEN Xinc <- +1 ELSE Xinc <- -1

   IF (ord (end (node)) > ord (origin (node))) THEN Yinc <- +1 ELSE Yinc <- -1

   X <- abs (origin (node)) ; Y <- ord (origin (node))

   store the point (X, Y)

   /* select the kind of move */

   IF (u > v) THEN DO

      sum <- u/2

      /* a rather horizontal move */

      FOR step number = 1 UNTIL u DO

         X <- X + Xinc

         sum <- sum + v

         IF (sum >= u) THEN DO

            sum <- sum - u

            Y <- Y + Yinc

         END

      store the point (X, Y)

      END

   ELSE DO

      sum <- v/2

      /* a rather vertical move */

      FOR step number = 1 UNTIL u DO

         Y <- Y + Yinc

```
            sum <- sum + u

            IF (sum >= v) THEN DO

                sum <- sum - v

                X <- X + Xinc

            END

        store the point (X, Y)

        END

    END

END
```